\documentclass[nohyperref]{article}

\usepackage{microtype}
\usepackage{graphicx}
\usepackage{subfig}
\usepackage{booktabs} 
\usepackage{dblfloatfix}

\usepackage{hyperref}

\usepackage[accepted]{icml2022}

\usepackage{amsmath}
\usepackage{amssymb}
\usepackage{mathtools}
\usepackage{amsthm}

\usepackage[capitalize,noabbrev]{cleveref}

\theoremstyle{plain}

\theoremstyle{definition}

\theoremstyle{remark}

\usepackage[textsize=tiny]{todonotes}

\icmltitlerunning{Large Scale Radio Frequency Signal Classification}

\begin{document}

\twocolumn[
\icmltitle{Large Scale Radio Frequency Signal Classification}

\icmlsetsymbol{equal}{*}

\begin{icmlauthorlist}
\icmlauthor{Luke Boegner}{equal,comp1}
\icmlauthor{Manbir Gulati}{equal,comp2}
\icmlauthor{Garrett Vanhoy}{equal,comp1}
\icmlauthor{Phillip Vallance}{comp3}
\vskip 0.05in
\icmlauthor{Bradley Comar}{comp3}
\icmlauthor{Silvija Kokalj-Filipovic}{comp1}
\icmlauthor{Craig Lennon}{comp3}
\icmlauthor{Robert D. Miller}{comp1}
\end{icmlauthorlist}

\icmlaffiliation{comp1}{Peraton Labs, Silver Spring, MD}
\icmlaffiliation{comp2}{Applied Insight, Fulton, MD}
\icmlaffiliation{comp3}{Laboratory for Telecommunication Sciences, College Park, MD}

\icmlcorrespondingauthor{Luke Boegner}{luke.boegner@peratonlabs.com}
\icmlcorrespondingauthor{Phillip Vallance}{pvallance@ltsnet.net}

\icmlkeywords{Machine Learning, Deep Learning, Radio Frequency, Signal, Classification, Dataset, Transformer, XCiT, EfficientNet, TorchSig, Sig53, RFML}

\vskip 0.3in
]



\printAffiliationsAndNotice{\icmlEqualContribution} 

\begin{abstract}
    Existing datasets used to train deep learning models for narrowband radio frequency (RF) signal classification lack enough diversity in signal types and channel impairments to sufficiently assess model performance in the real world.
    We introduce the Sig53 dataset consisting of 5 million synthetically-generated samples from 53 different signal classes and expertly chosen impairments. 
    We also introduce TorchSig, a signals processing machine learning toolkit that can be used to generate this dataset.
    TorchSig incorporates data handling principles that are common to the vision domain, and it is meant to serve as an open-source foundation for future signals machine learning research.
    Initial experiments using the Sig53 dataset are conducted using state of the art (SoTA) convolutional neural networks (ConvNets) and Transformers.
    These experiments reveal Transformers outperform ConvNets without the need for additional regularization or a ConvNet teacher, which is contrary to results from the vision domain.
    Additional experiments demonstrate that TorchSig's domain-specific data augmentations facilitate model training, which ultimately benefits model performance.
    Finally, TorchSig supports on-the-fly synthetic data creation at training time, thus enabling massive scale training sessions with virtually unlimited datasets.
\end{abstract}

\section{Introduction}

\begin{figure}[!t]
    \vskip 0.1in
    \begin{center}
    \centerline{\includegraphics[width=0.48\textwidth]{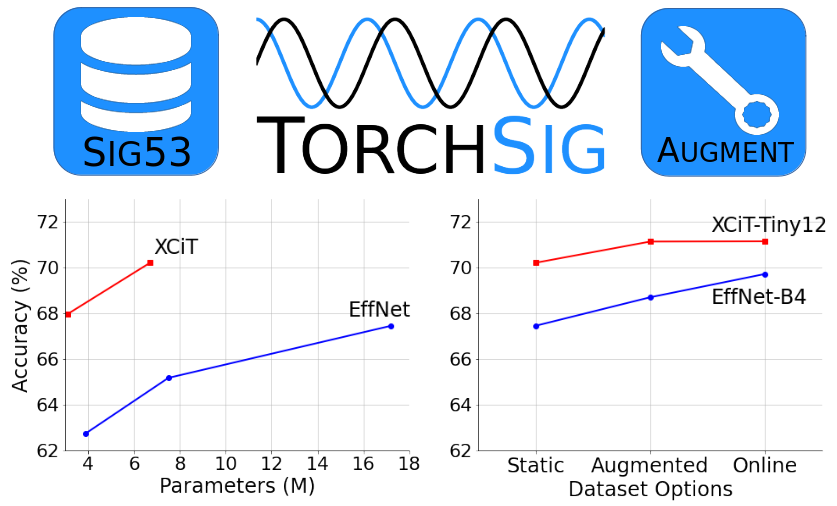}}
    \caption{
        In this work, we introduce the Sig53 modulated signals dataset and TorchSig, 
        an open-source signals processing machine learning toolkit.
        We evaluate EfficientNet and XCiT performance against the Sig53 dataset, 
        and we demonstrate the benefits of TorchSig's domain-tailored data augmentations 
        and online data generation capabilities.
        }
    \label{fig:paper-overview}
    \end{center}
    \vskip -0.3in
\end{figure}

The RadioML dataset originally presented in \cite{o2016radio} and later revised in \cite{o2018radio} has been one of the radio frequency machine learning (RFML) community's most heavily used datasets. It has been used extensively in the development of deep learning models for narrowband RF signal classification.
These works, along with other early efforts, collectively represent the first applications of modern deep learning to the field of RF.
These datasets made significant headway in a field where it was common to focus on only a few signals in the same family with relatively simple channel impairments (e.g. Gaussian noise).
However, progress in SoTA deep learning models necessitates more complicated RF datasets that better approximate the nature of captured signals.

Along with RadioML, there have been other datasets that have been created for different purposes.
\cite{wong2021rfml} summarizes other RFML datasets used for various purposes. 
However, many of these prior works lack the realism of real-world impairments and lack reproducibility by not sharing the datasets openly.
Additional works have shown benefits in data generation and augmentation techniques, but there does not yet exist a robust community tool for implementing and experimenting with these findings \cite{miller2019policy}.

Among the existing work on these RFML datasets, very few researchers have attempted to use Transformer architectures \cite{vaswani2017attention}.
These networks were originally created in the natural language processing (NLP) domain but then were adapted for the computer vision domain where they challenged the existing SoTA \cite{dosovitskiy2020vistrans}. 
In this work, simple adaptations are made to the Transformer architecture to evaluate their performance in the RF domain, particularly against our newly introduced Sig53 dataset.

This paper is laid out as follows: 
\cref{sec:radioml} demonstrates limitations in the seminal RFML dataset, RadioML. 
\cref{sec:dataset} introduces the Sig53 dataset and the TorchSig RFML software toolkit.
\cref{sec:experiments} shares preliminary experimentation and classification performance of ConvNets and Transformers on the Sig53 dataset.
The paper then concludes with \cref{sec:conclusion}.

\section{Existing Work: RadioML}
\label{sec:radioml}

\begin{figure*}[ht]
    \centering
    \begin{tabular}{lcr}
        \subfloat[Training vs Validation Loss]{\label{fig:rml-train-val-loss}
        \includegraphics[width=.32\textwidth]{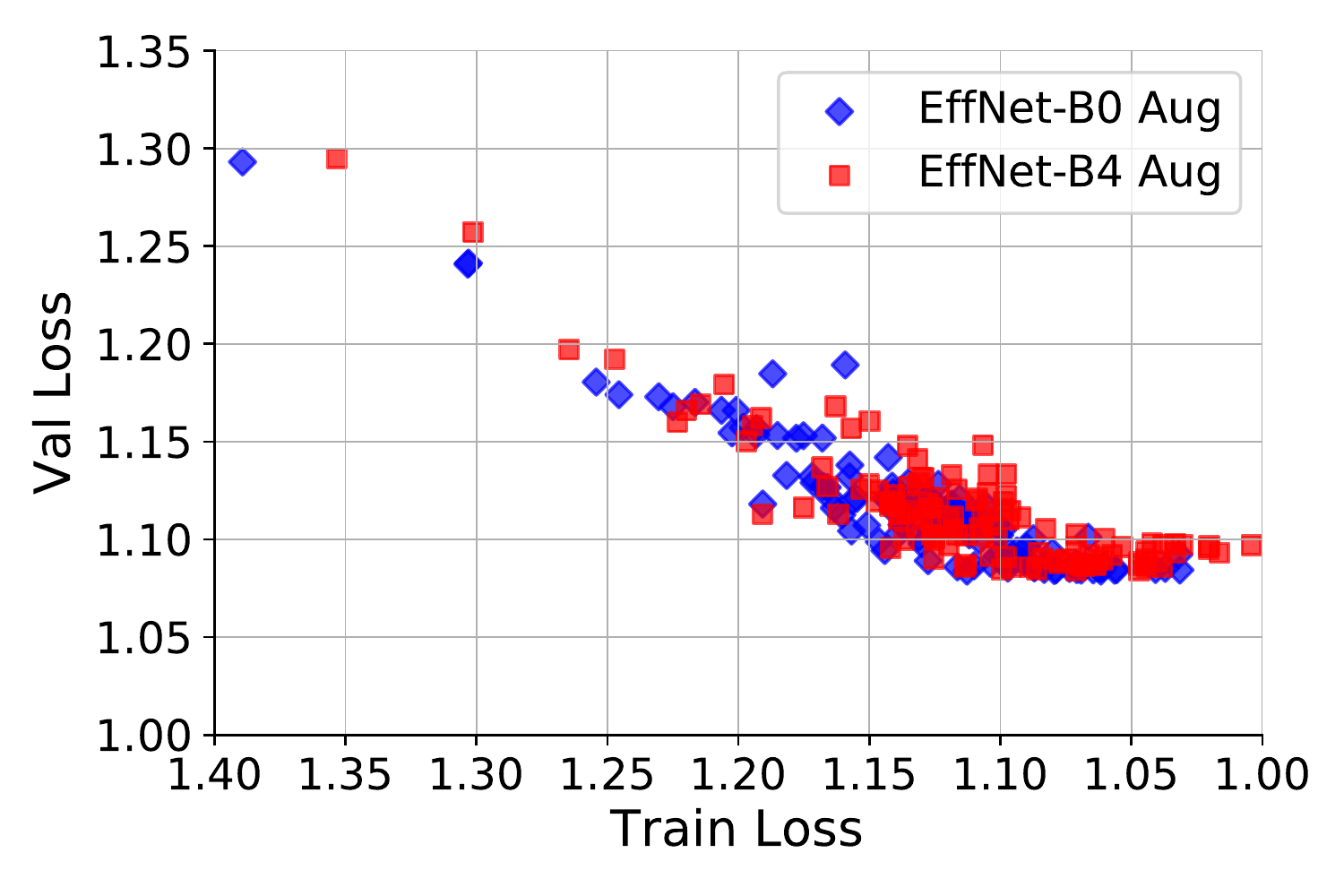}} &
        \subfloat[Accuracy vs Number of Parameters]{\label{fig:rml-acc-params}
        \includegraphics[width=.32\textwidth]{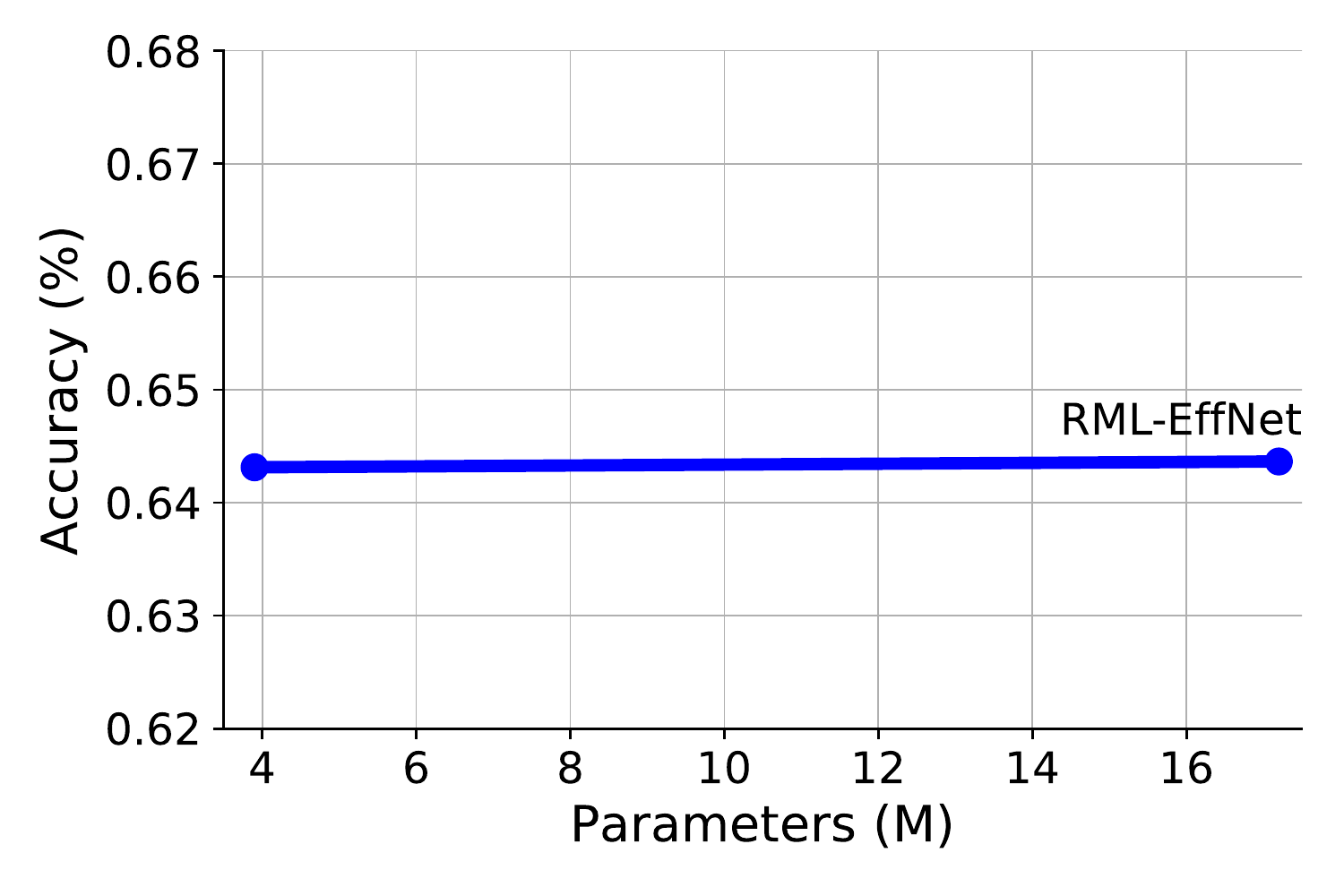}} &
        \subfloat[Accuracy vs SNR]{\label{fig:rml-acc-snr}
        \includegraphics[width=.32\textwidth]{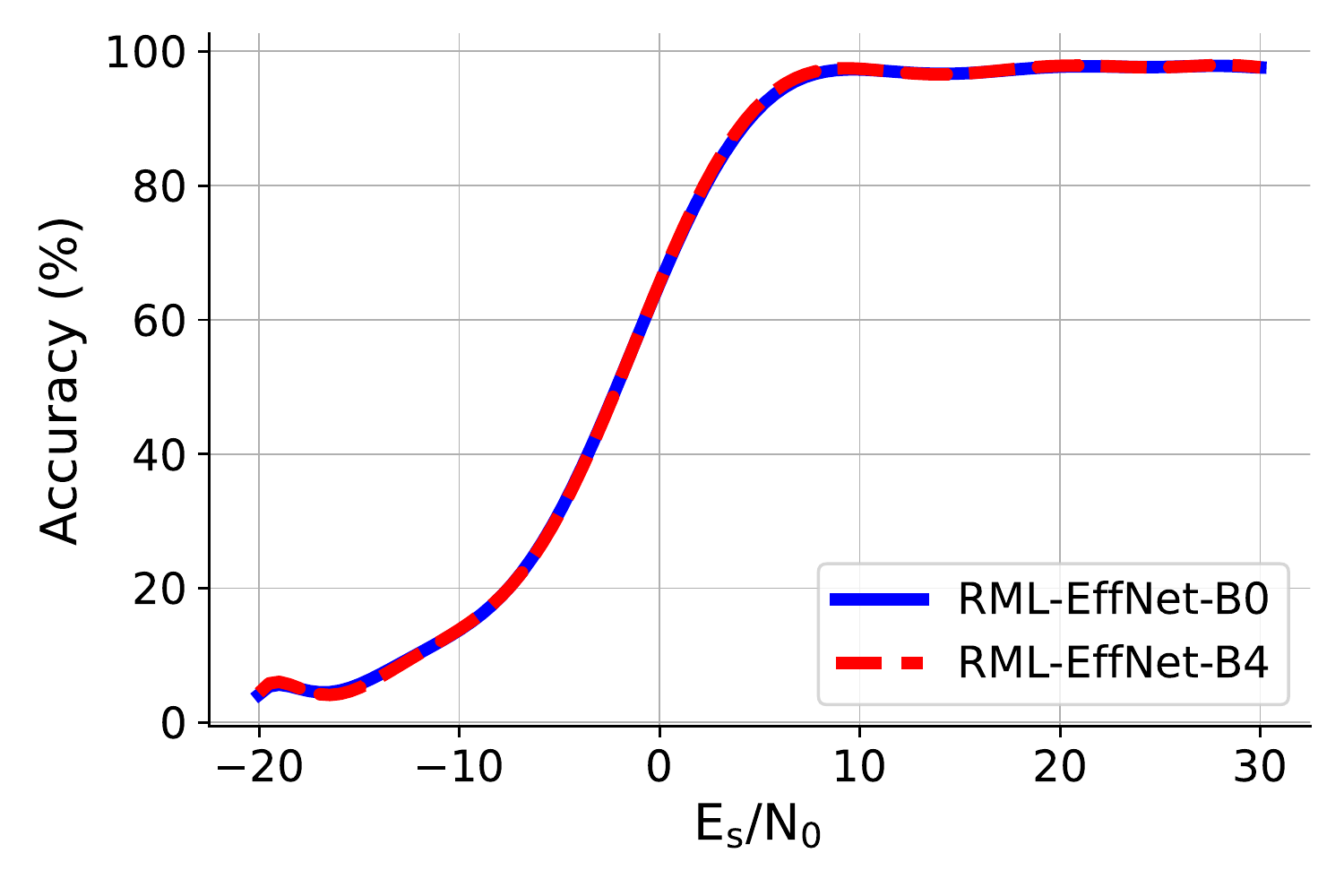}} \\
    \end{tabular}
    \caption{
        RadioML lacks performance scaling with model scaling.  
        \subref{fig:rml-train-val-loss} indicates that both the B0 and B4 EfficientNets (with data augmentations) experience overfitting as the training loss continues to decrease 
        while the validation loss ceases to do so.  
        \subref{fig:rml-acc-params} indicates that the RadioML dataset fails to provide training data of sufficient quality to improve the performance of the larger EfficientNet model over the smaller one.  
        \subref{fig:rml-acc-snr} shows that this lack of accuracy distinction occurs over all the various SNRs in the dataset.
        }
    \label{fig:rml-results}
\end{figure*}

The RadioML datasets, along with the use of ConvNets tailored to these datasets, are recognized as the foundational work in the application of deep neural networks to the RF domain.
The initial release in 2016 consisted of 11 signal classes, where each class example is comprised of 128 complex-valued samples.
A 5-layer ConvNet was included as the architecture for signal classification.
In 2018, a revised RadioML dataset was released, increasing the class set to 24 and lengthening the class examples to 1024 complex-valued samples.
More advanced neural network architectures were explored using this dataset.
In \cite{o2018radio}, the authors modified Residual Networks (ResNets) \cite{he2015deep} for the RadioML dataset and reported the models' performance across signal classes and SNR levels.
These datasets and networks have been well cited since their release and were a catalyst for machine learning research in the RF domain.
As machine learning research evolves, it is imperative that the underlying benchmark keeps pace.
While the RadioML datasets vitalized RFML research, we postulate several key limitations are holding back progress in this field.

We investigated qualitative limitations with RadioML. 
First, from the ML research perspective, the RadioML datasets consist of a single set, leaving researchers to decide how to split their training and validation sets.
While this detail can easily be handled by researchers, comparative performance analysis requires a static, consistent set of data examples.
Additionally, many publications follow the authors' performance metrics by creating accuracy versus SNR curves.
These metrics are justified in understanding how models perform as SNR levels change. 
However, the multi-dimensional metric makes performance comparisons more complicated.
Thus, we suggest adopting the single accuracy metric as used in the vision domain.
Finally, while the RadioML dataset serves well as a static dataset, 
we suggest that the community would benefit from a dataset that also has its generation tools available through open-source code.

From an RF domain perspective, there are limitations with RadioML in areas such as signal diversity, impairment applications, and sample selection.
The authors partially address these limitations in their improved 2018 dataset through an increase in signal classes and applied impairments.
However, we postulate recent ML improvements and qualitatively observed gaps from synthetic to real operations necessitate an additional step in complexity.

In addition to the qualitative limitations highlighted above, we conducted experiments to demonstrate quantitative limitations in network architecture research using the RadioML 2018 dataset.
Within the vision domain, EfficientNets demonstrate model scaling capabilities, where neural architectures can be systematically enlarged or reduced for streamlining the performance versus speed tradeoff \cite{tan2019efficientnet}.
We demonstrated that performance scaling of EfficientNets trained on RadioML's 2018 dataset were problematic due to the limited complexity of the signal samples.
As an initial experiment, the smallest network, EfficientNet-B0, was trained on the RadioML dataset, using an 80-20 random train-val split on the 2.5 million data samples.
Despite the seemingly large dataset size, EfficientNet-B0 immediately overfits to the training data.
To avoid overfitting, we apply data augmentations (detailed in \cref{sec:dataset}) during training.
We conducted training sessions using identical schedules for both EfficientNet-B0 and EfficientNet-B4 using augmentations with the hope of seeing a scaled performance boost for the larger network.
Unfortunately, despite the increased network size, both networks achieved comparable performance on the RadioML 2018 dataset. 
Despite the use of data augmentations, both networks eventually overfit to the training data (\cref{fig:rml-results}).
The inability to train EfficientNet-B0 without augmentations, the lack of performance scaling with increasing model size from EfficientNet-B0 to EfficientNet-B4, 
and the eventual overfitting with data augmentations all suggest that modern neural architectures' learning capacities are outpacing the data complexity observed in the RadioML 2018 dataset.

\begin{figure*}[ht]
    \centering
    \begin{tabular}{lcr}
        \subfloat[4 ASK]{\label{fig:dataset_ask}
        \includegraphics[width=.32\textwidth]{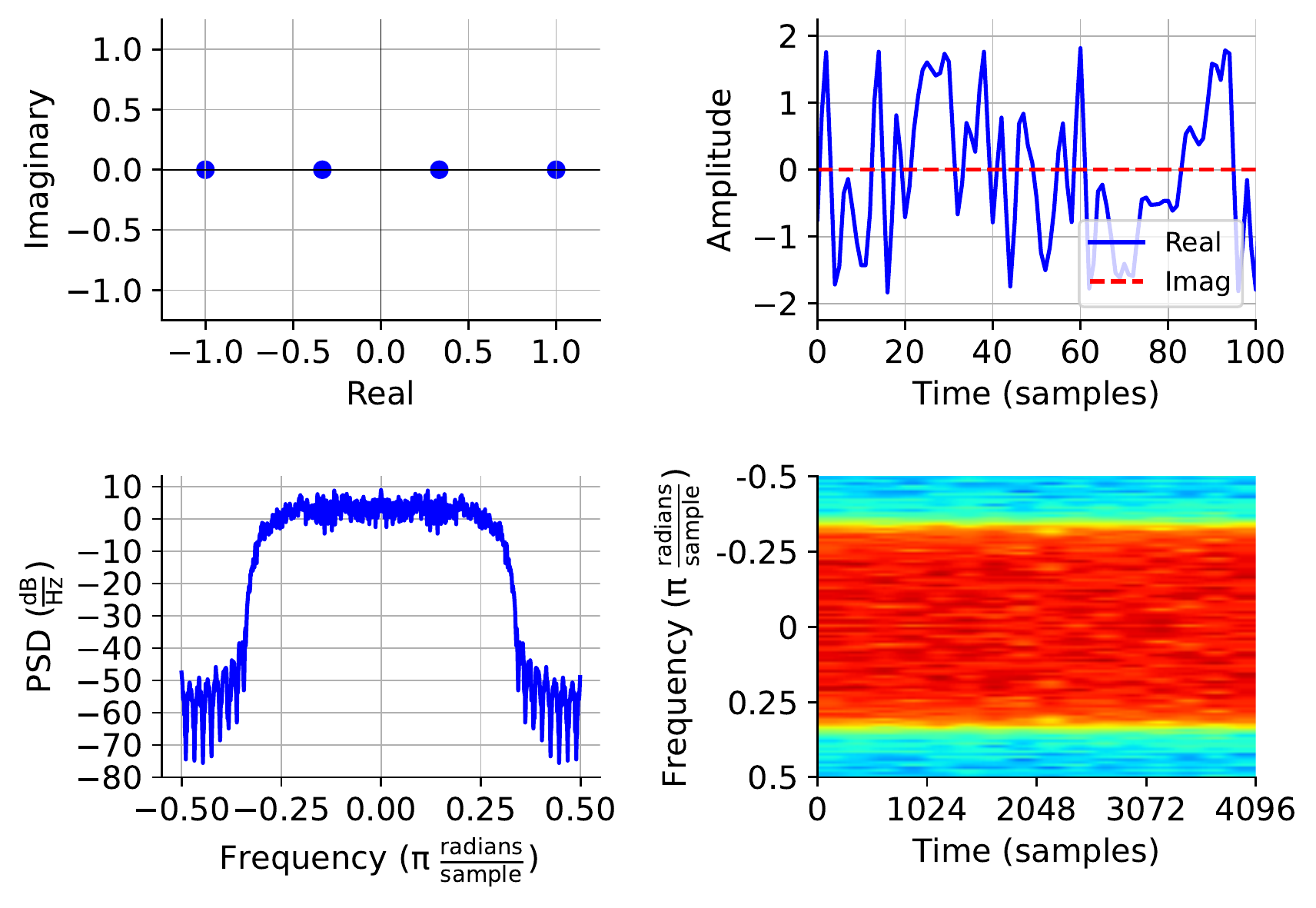}} &
        \subfloat[8 PAM]{\label{fig:dataset_pam}
        \includegraphics[width=.32\textwidth]{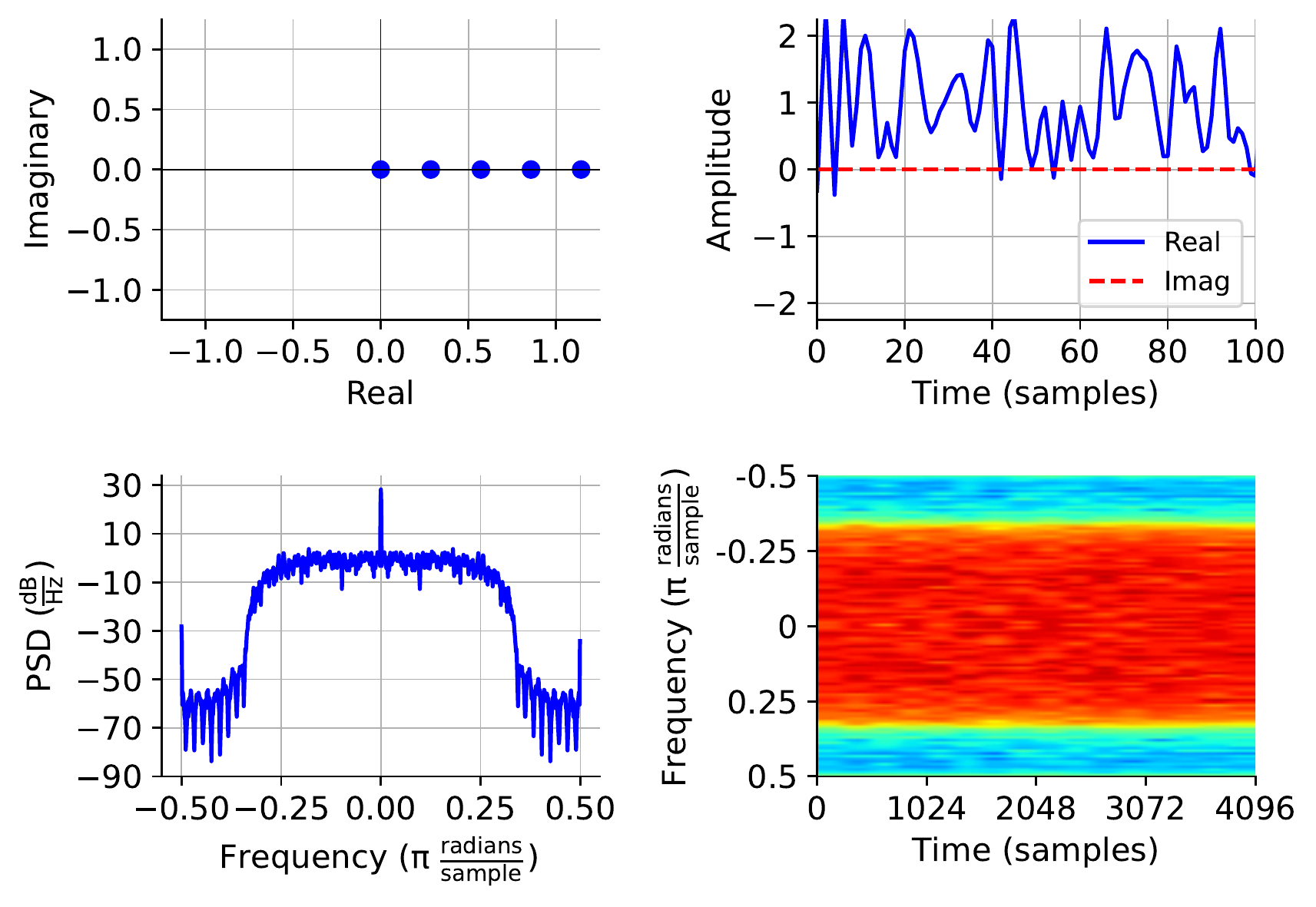}} &
        \subfloat[16 PSK]{\label{fig:dataset_psk}\includegraphics[width=.32\textwidth]{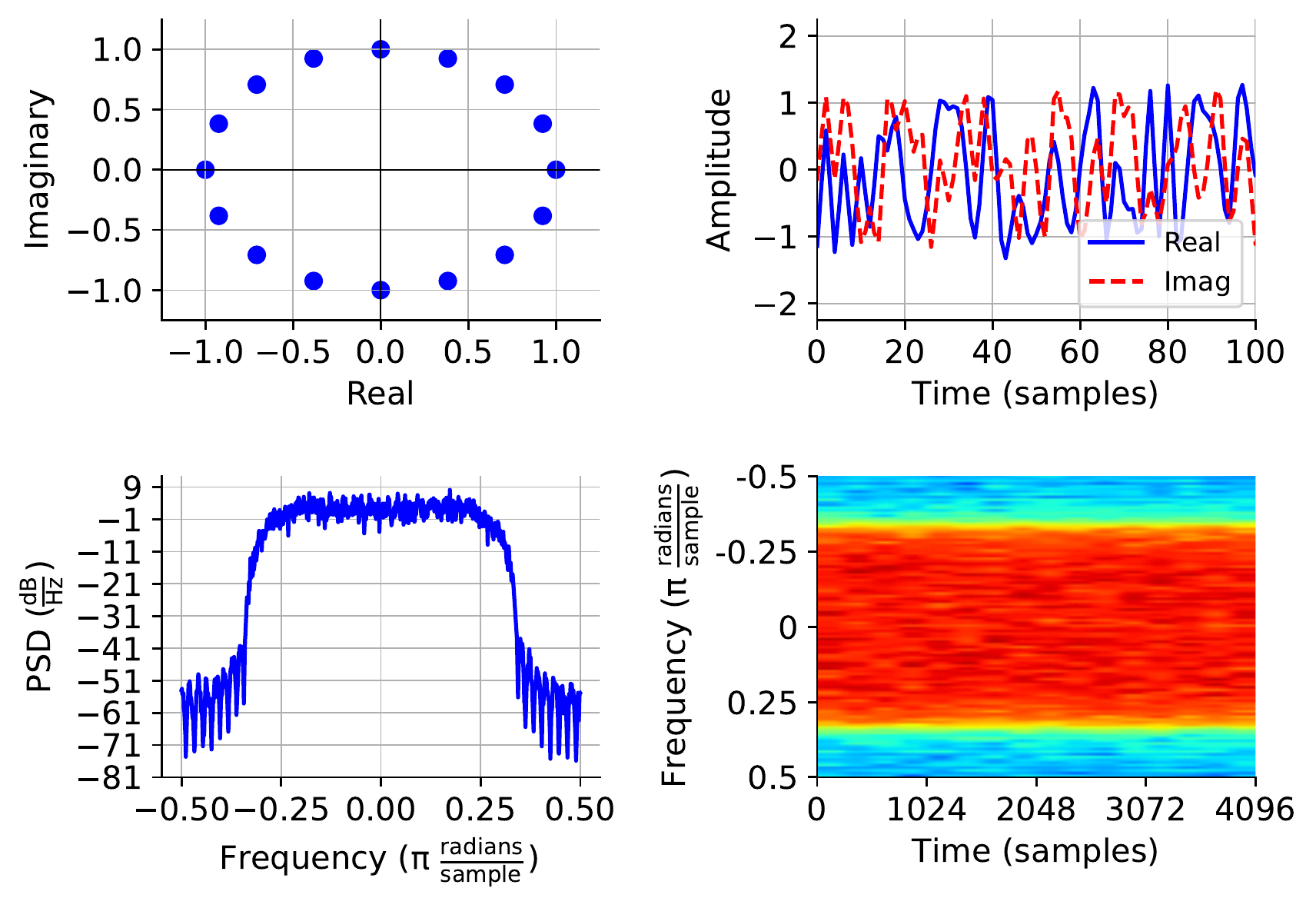}} \\
        \subfloat[32 QAM-Cross]{\label{fig:dataset_qam}\includegraphics[width=.32\textwidth]{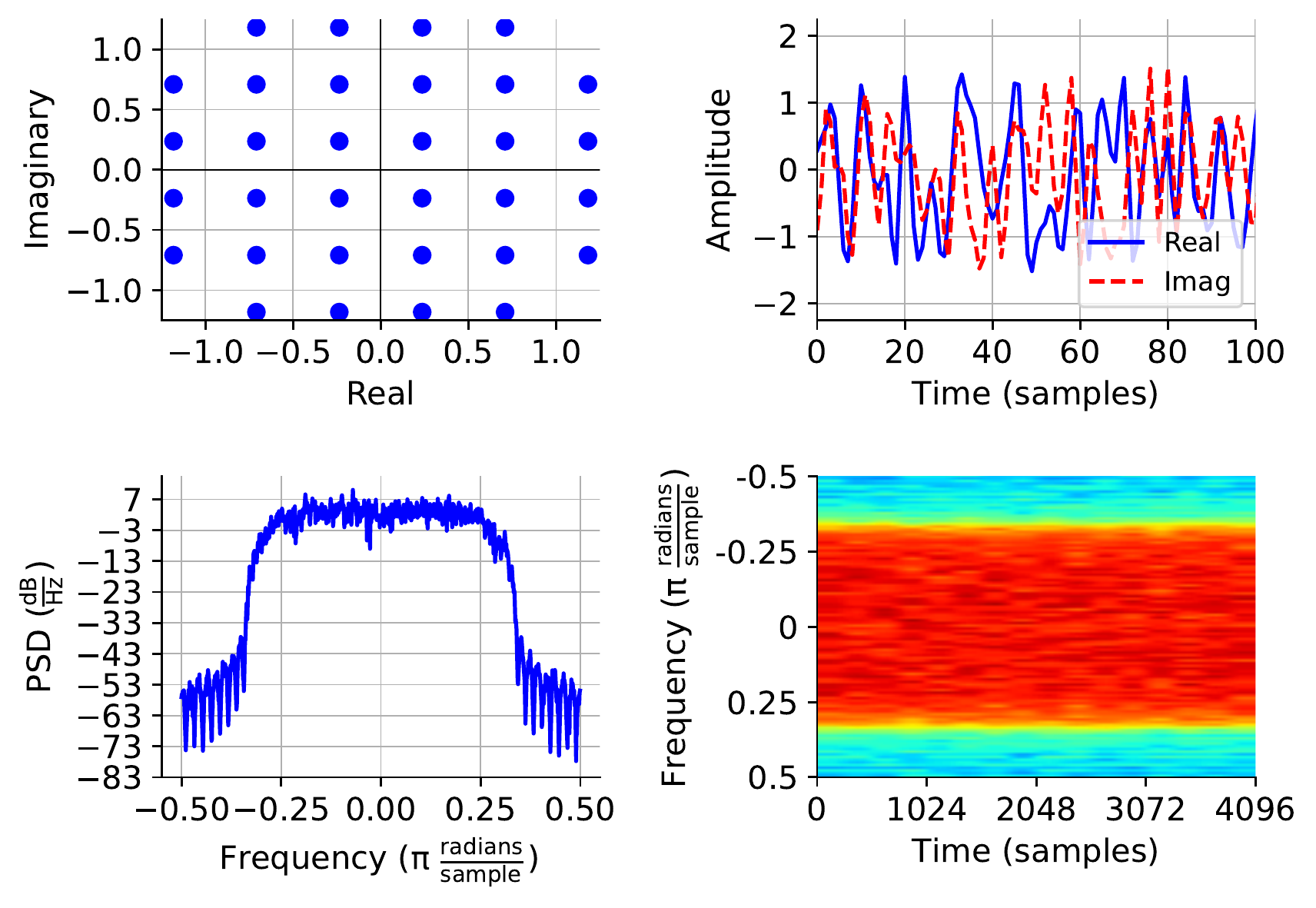}} &
        \subfloat[2 FSK]{\label{fig:dataset_fsk}
        \includegraphics[width=.32\textwidth]{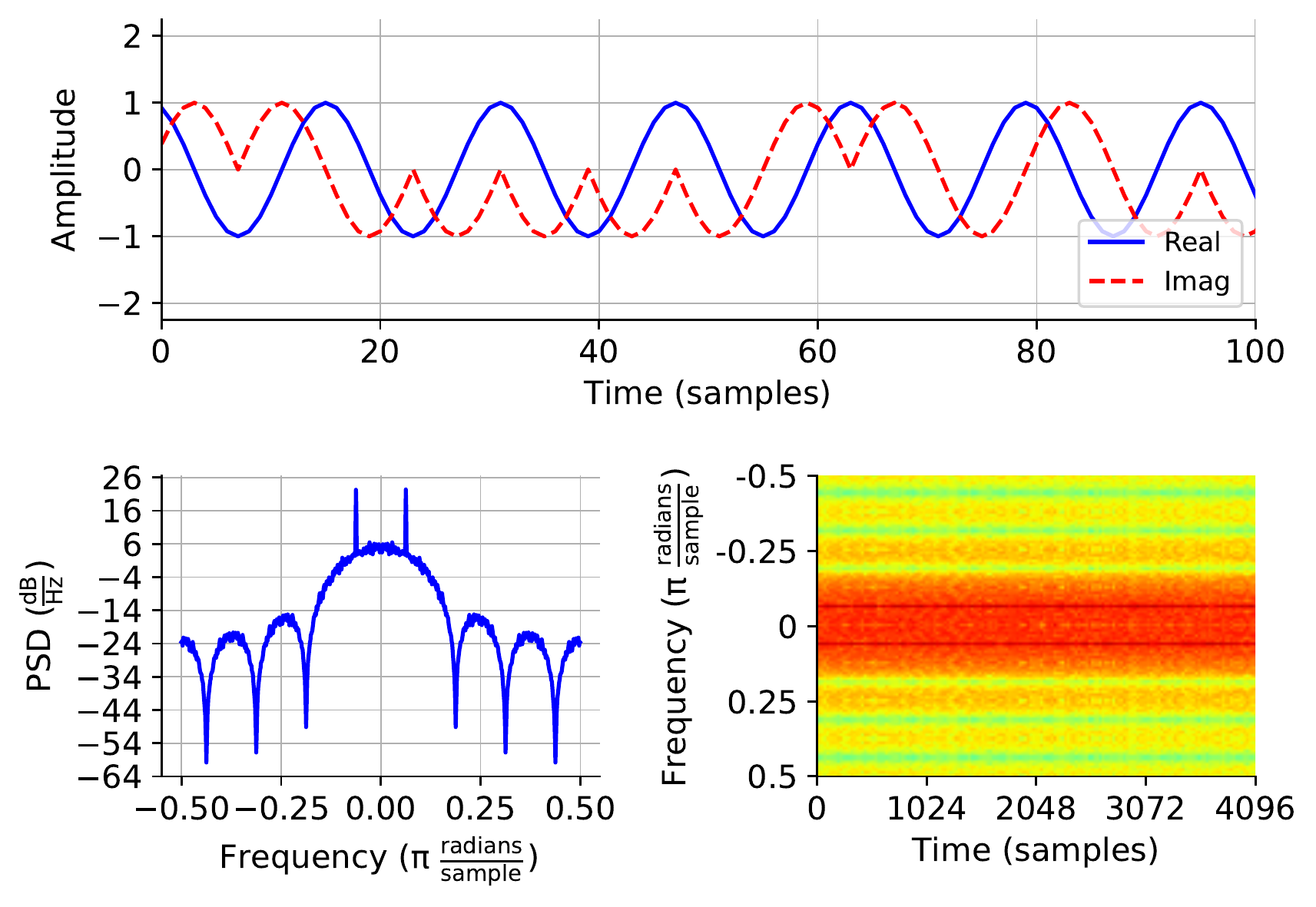}} &
        \subfloat[OFDM-256]{\label{fig:dataset_ofdm}\includegraphics[width=.32\textwidth]{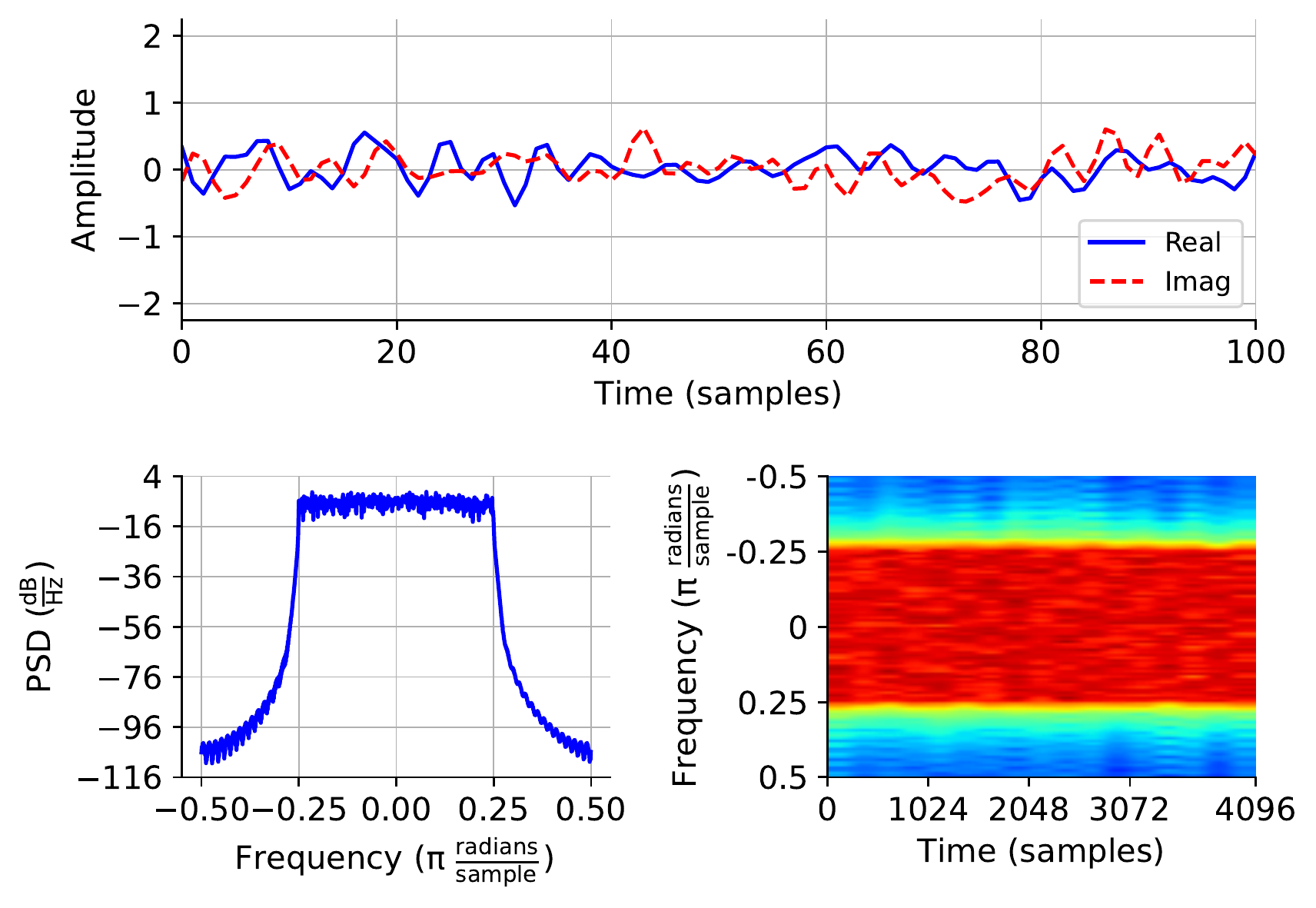}}
    \end{tabular}
    \caption{Visualizations of clean signals are shown with 1 example for each modulation family.  Signals are shown in both the time and frequency domains, including a spectrogram which shows frequency patterns over time.  Constellation diagrams are also given for relevant signals.}
\end{figure*}

Similar to the migration from MNIST \cite{deng2012mnist} to ImageNet \cite{deng2009imagenet} in the vision domain, a next generation RF signals dataset needs to supersede the foundational RadioML datasets.

\section{Sig53 Dataset and TorchSig}
\label{sec:dataset}

This work presents the next evolution of RFML signal classification datasets, termed the Sig53 dataset.
The goal of creating such a dataset is to facilitate the advancement of SoTA techniques in the RFML domain.
Given the diversity of wireless environments and signal types that exist, it is impossible to include every signal and environment in a single dataset.
However, the goal is to create a dataset that provides significant challenges inherent to the problem at hand.
When a model performs well on a challenging dataset, then ideally, it should be able to successfully transfer-learn to a dataset with new signals or environments.
These new signals and environments may be obtained through signal captures or from different synthetic models.
This concept mirrors how computer vision models can be pre-trained on ImageNet, and those models can then leverage transfer learning for specific applications.

We incorporate lessons learned from the RadioML dataset and build the Sig53 dataset to provide the following improvements:

\textbf{Signal Diversity:} We increase the number of signal classes from 24 to 53 and include more families of signal modulations.
The increase in signal diversity challenges machine learning models to differentiate additional signals from multiple families as well as 
signals that are more closely related from within the same family.  

\textbf{Impairment Diversity:} In RadioML's 2016 dataset, impairments are seeded the same across signal modulations types;
thus, the same impairments are applied across all signals.
In the Sig53 dataset, impairments are randomized according to specified distributions across all signal modulation types in Sig53.  
This reduces the risk of a model learning that there is a common channel across signal types 
and compensating for it internally to the model instead of learning discerning features of the signals themselves.

\textbf{Sample Selection:} RadioML's datasets include signals with SNRs down to -20dB so performance analysis can clearly show accuracy collapse across all signals.
The tradeoff is that the examples with very low SNR are often still included during model training procedures, causing the model to receive examples with unlearnable signals.
In Sig53, we carefully select signal types and impairments so that a machine learning model can make use of a large majority of training examples to learn more about a signal type. 

The Sig53 dataset consists of over 6 million examples split into 4 distinct sub-datasets:
\begin{itemize}
    \setlength\itemsep{0em}
    \item Clean Training (1M examples),
    \item Clean Validation (106k examples),
    \item Impaired Training (5.3M examples), and
    \item Impaired Validation (106k examples).
\end{itemize}
The Sig53 sub-datasets intentionally split training and validation examples, such that the validation datasets can be used collaboratively and competitively across the RFML community.
This is similar to the computer vision community's use of the ImageNet dataset (and corresponding challenge) as a benchmark.
The Sig53 dataset also provides training and validation datasets in both clean and impaired formats. 
Impaired examples are passed through realistic data impairments as defined below, while the clean examples enable researchers to apply customized impairments tailored to their use case.

The default example size across all sub-datasets is 4096 complex-valued samples, representing I (in-phase or real) and Q (quadrature or imaginary) samples. 
This is larger than many existing RFML datasets. 
The selected example size ensures sufficient symbol diversity in the higher order modulation schemes.
It is also short enough to hold reasonable assumptions on constant, time-invariant RF channels for certain baud rates. 
Additionally, the 4096 IQ sample size is short enough to enable consensus for longer duration real-world signals through multiple ML inferences.

In addition to the static Sig53 dataset, we also present TorchSig, a general-purpose PyTorch-based RFML software toolkit \cite{paszke2017automatic}. 
TorchSig is the first complex-valued signal processing ML toolkit and can be used to generate the Sig53 dataset as well as apply numerous complex-valued signal domain-tailored impairments, augmentations, and transformations.
Most importantly, TorchSig allows the community to more rapidly develop, iterate, and share future RFML research. 
Details on how TorchSig generates the Sig53 dataset and additional relevant features are described below.

\subsection{Clean Dataset}
\label{sec:clean-dataset}
Clean datasets are synthetically generated using randomized data that is fed into 1 of 6 families of modulations: amplitude-shift keying (ASK), pulse-amplitude modulation (PAM), phase-shift keying (PSK), quadrature amplitude modulation (QAM), frequency-shift keying (FSK), or orthogonal frequency-division multiplexing (OFDM). 
Each of these families of modulations contains multiple specific classes that are used to generate the 53 unique modulation variations. 
See Appendix A.1 for the full class list.

The ASK, PAM, PSK, and QAM families are all constellation-based waveforms, 
where data is mapped to symbols that are plotted along real and imaginary axes.
IQ symbols are then pulse-shaped using a root-raised cosine (RRC) filter with an alpha value of 0.35.
ASK signals' symbols are evenly distributed from -1 to +1 along the real axis (\cref{fig:dataset_ask}).
PAM signals' symbols are evenly distributed from 0 to +1 along the real axis (\cref{fig:dataset_pam}).
PSK signals' symbols are evenly distributed around the complex unit circle (\cref{fig:dataset_psk}).
QAM signals' symbols are evenly distributed within a 2-dimensional box with a lower left corner at -1-1j\footnote{$j = \sqrt{-1}$} and an upper right corner at 1+1j (\cref{fig:dataset_qam}).

The FSK family of signals modulates the random data sequence to distinct frequency values. 
Within the FSK family, there are pure FSK classes that simply implement frequency-shift key modulation.
There are also MSK classes that implement minimum-shift keying modulation through changing the frequency modulation index from 1.0 to 0.5.
Additionally, there are GFSK classes that implement a Gaussian filter on top of the base FSK signals with a bandwidth-time (BT) product of 0.35.
Finally, there are GMSK classes that represent MSK signals using a Gaussian pulse-shaping filter. 
When the Gaussian filter is not present, the pure FSK and MSK signals have large sidelobes. 
Because of these large sidelobes, the pure FSK and MSK signals are sampled at a higher rate of 8 IQ samples per symbol, while the GFSK and GMSK signals are sampled at 2 IQ samples per symbol. 
While this breaks from the standard used by other families of signals, the main goal of the clean dataset is to provide signals with minimal impairments.  This enables users to verify performance and/or apply impairments of their choosing. 
More details on our design choices for standardization across impaired signals can be found in \cref{sec:impaired-dataset}.\newline

The OFDM family of signals also differs because it is the only multi-carrier waveform. 
The subcarrier modulations are selected from a subset of constellation-based modulations to include: BPSK, QPSK, 16QAM, 64QAM, 256QAM, or 1024QAM. 
For half of the OFDM examples, the modulation is randomly selected and applied to all of the OFDM subcarriers.  
For the other half of the OFDM examples, the modulation is randomly selected for each of the subcarriers independently. 
The classes within the OFDM family of signals are set by the number of subcarriers. The number of sub-carriers, $N$, is drawn from a set selected to mirror real-world specifications for common WiFi and LTE signals. 
Additionally, within OFDM, cyclic prefixes are also used to assist in reducing intersymbol interference. 
Within the Sig53 dataset, OFDM signals are generated with a cyclic prefix duration of either an eighth or a quarter of the full symbol duration. 
A DC subcarrier is also present for half of the OFDM signals and absent for the remaining ones.
OFDM's symbol boundaries present an amplitude-phase discontinuity. 
So, for added realism, half of the OFDM signals are passed through a low pass filter, emulating the transmission filter of a real-world system.
The remaining half employs a windowing operation to suppress sidelobes.
Generally, a demodulator for OFDM does not use the standard 2 samples per symbol found in single carrier modulations.
To ensure that the OFDM signals are in distribution with the other signal types, they are generated to occupy half of the frequency bandwidth (\cref{fig:dataset_ofdm}).  
This closely approximates the 2 samples per symbol found in the other signal types.
Each of the above effects are applied randomly as well as independently with respect to each other.  
However, the networks must overcome these added differences to perform well because OFDM classes are distinguished only by the number of subcarriers.

\subsection{Impaired Dataset}
\label{sec:impaired-dataset}
Impaired datasets use the same signals described in the previous section and then impose synthetic impairments emulating real-world system impairments or environmental effects.
Key impairments applied to all examples are additive white Gaussian noise (AWGN) in order to achieve target SNR levels and the randomization of pulse shaping methods during signal generation.
Details on these impairments are described below.

\subsubsection{SNR Definitions}
The SNR metric within the Sig53 dataset is defined as the energy per symbol to noise power spectral density ($E_s / N_0$). 
\cite{wong2021rfml} discusses the distinction between defining SNR within RFML datasets using $E_s / N_0$ versus using the energy per bit to noise power spectral density ($E_b / N_0$).
The $E_b / N_0$ measurement is popular among modem designers who want to understand the price, in power, that must be paid for transmitting each bit of information across a given channel.
However, we assert all signals datasets should use the $E_s / N_0$ power definition, which describes the SNR of the symbols, irrespective of the number of bits each symbol contains.
This assertion is due to the necessity of balanced power levels when training neural networks.
In datasets defining SNR with $E_b / N_0$, there is an inherent imbalance in the higher order modulations mapping them to higher powered data examples.
This imbalance in SNR levels ultimately enables the model to learn undesirable shortcuts.
Within the $E_s / N_0$-defined SNR levels, the impaired Sig53 dataset uniformly distributes SNR levels within the range of -2dB to 30dB.
For more details on the differences in power definitions, see \cref{sec:appendix_dataset}.

\subsubsection{Random Pulse Shaping}
Randomized pulse shaping is performed on all classes except OFDM signals, causing otherwise identical signals to occupy slightly more or less bandwidth.
For the constellation-based signals using the RRC pulse-shaping filter, the alpha value is changed from 0.35 (for clean signals) to a uniformly random value in the range from 0.15 to 0.60.
This covers a slightly larger range than what is typically expected in real signals. 
The FSK random pulse shaping is implemented differently between the GFSK/GMSK sub-families of signals and the remaining FSK and MSK signals. 
Within the GFSK and GMSK signals, the bandwidth-time (BT) product (set to 0.35 for clean signals) is set to a uniformly random value in the range from 0.1 to 0.5. 

The non-Gaussian filtered FSK and MSK signals that previously contained no pulse shaping are now passed through a new low pass filter with a randomized passband and complementary downsampling. 
This practice emulates how a wideband spectral sensing pre-processing system may detect and extract these signals that contain very large sidelobes.
Ultimately, this operation modifies the FSK and MSK signals to be in distribution with regards to coarse sampling and occupied bandwidth.
For PSD visualizations of all the random pulse shaping techniques, see \cref{sec:appendix_dataset}.

\subsubsection{Impairments}
While we apply the 2 impairments above to every signal, we apply 6 additional impairments with certain likelihoods.
These likelihoods are intentionally below 100\% in order to introduce greater variability in the combined effects.
Each of these 6 impairments comes with their own uniformly randomized parameters. 
These impairments are: phase shift, time shift, frequency shift, Rayleigh fading, IQ imbalance, and resampling. 
The details of their effects are below, and \cref{sec:appendix_dataset} contains visualizations of their effects.

\textbf{Phase Shift:} We apply the phase shift impairment with a 90\% probability.  
These phase shifts range from -$\pi$ to $\pi$ radians when applied.

\textbf{Time Shift:} We apply the time shift impairment with a 90\% probability, and these time shifts range from -32 to +32 IQ samples when applied.  
Note that any resulting empty regions are filled with zeros.

\textbf{Frequency Shift:} We apply the frequency shift impairment with a 70\% probability.  
These frequency shifts range from -16\% to 16\% of the sampling rate when applied.

\textbf{Rayleigh Fading Channel:} We apply a Rayleigh fading channel model with a 50\% probability, 
with a randomized number of taps between 2 and 20 that follow a tapered power delay profile. 
The Rayleigh fading channel is modeled as a finite impulse response (FIR) filter with Gaussian distributed taps. 
The length of the filter determines the delay spread of the channel, and it is inversely proportional to the coherence bandwidth.

\textbf{IQ Imbalance:} We apply IQ imbalance with a probability of 90\%, with 3 randomized parameters: amplitude imbalance, phase imbalance, and DC offset. 
We apply amplitude imbalance over the range of -3 to +3 dB.  
We apply phase imbalance over the range of (-$\pi$/180) to ($\pi$/180) radians.  
We apply DC offset over the range of -0.1 to +0.1 dB.

\textbf{Random Resample:} The data is also passed through a randomized resampler with a 50\% probability. 
The resampling factor ranges from 0.75 to 1.5. 
Since resampling causes the total number of IQ samples to change, the data is either truncated or zero-padded back to the desired 4096 IQ samples.

While we believe our Sig53 dataset design decisions are optimal for the goal of our research, we understand other research in closely-related but slightly different tasks may require variations.
For these tasks, we invite researchers to use the open-source TorchSig RFML software toolkit to generate a modified Sig53 dataset.

\section{Experiments}
\label{sec:experiments}

In this section, we outline results from adapting SoTA vision classification networks to the 1D signals domain against an appropriately impaired Sig53 dataset.
We also demonstrate the benefits of our newly introduced open-source TorchSig toolkit's data augmentations and online data generation.
Finally, we analyze the performance of the networks' accuracies across SNR values and signal classes.

\subsection{Network Architectures}
\textbf{EfficientNets:} EfficientNets are SoTA ConvNets that are most notable for their parameter and FLOPS efficiency compared to other networks.
They consist of 8 networks, B0-B7, created using compound scaling of the base network B0, which is found by neural architecture search.
For this research effort, B0, B2, and B4 networks are selected for evaluation.
For additional EfficientNet architecture background, see \cref{sec:effnet_background}.

\textbf{XCiT Transformers:} Cross-Covariance Image Transformers (XCiT) \cite{el2021xcit} are modified transformer variants in which attention is performed over the channel dimension instead of the token dimension. 
This transformer is chosen for its strong performance and linear scaling in terms of sequence length. 
A single convolutional layer is used as the network frontend.
In this layer, we downsample by a factor of 2 and project to the correct channel count.
Similar to EfficientNets, XCiT transformers consist of multiple scales, ranging from Nano to Large.
In this research, Nano and Tiny12 networks are selected for evaluation.
For additional XCiT architecture background, see \cref{sec:xcit_background}.

\begin{table*}[h]
  \caption[fontsize=12pt]{Results for various networks using the Sig53 impaired dataset are shown here.}
  \label{tab:sig53-results}
  \vskip 0.1in
         \begin{center}
          \centering
          \begin{tabular}{l|l|l|l}
              \toprule[1.5pt]
           Model & Params. & FLOPS & Accuracy  \\ \midrule
            EfficientNet-B0 & 3.9M & 0.53B & 62.75\% \\
            EfficientNet-B2 & 7.5M & 1.0B & 65.18\% \\
            EfficientNet-B4 & 17.2M & 2.3B & 67.46\% \\ \midrule
            XCiT-Nano & 3.1M & 5.0B & 67.97\% \\
            XCiT-Tiny12 & 6.7M & 11.4B & 70.22\% \\
          \bottomrule[1.5pt]
          \end{tabular}
     \end{center}
  \vskip -0.1in
\end{table*}

\begin{figure*}[h]
  \centering
  \subfloat[Accuracy vs FLOPS]{\label{fig:sig53-acc-flops}\includegraphics[width=0.5\textwidth]{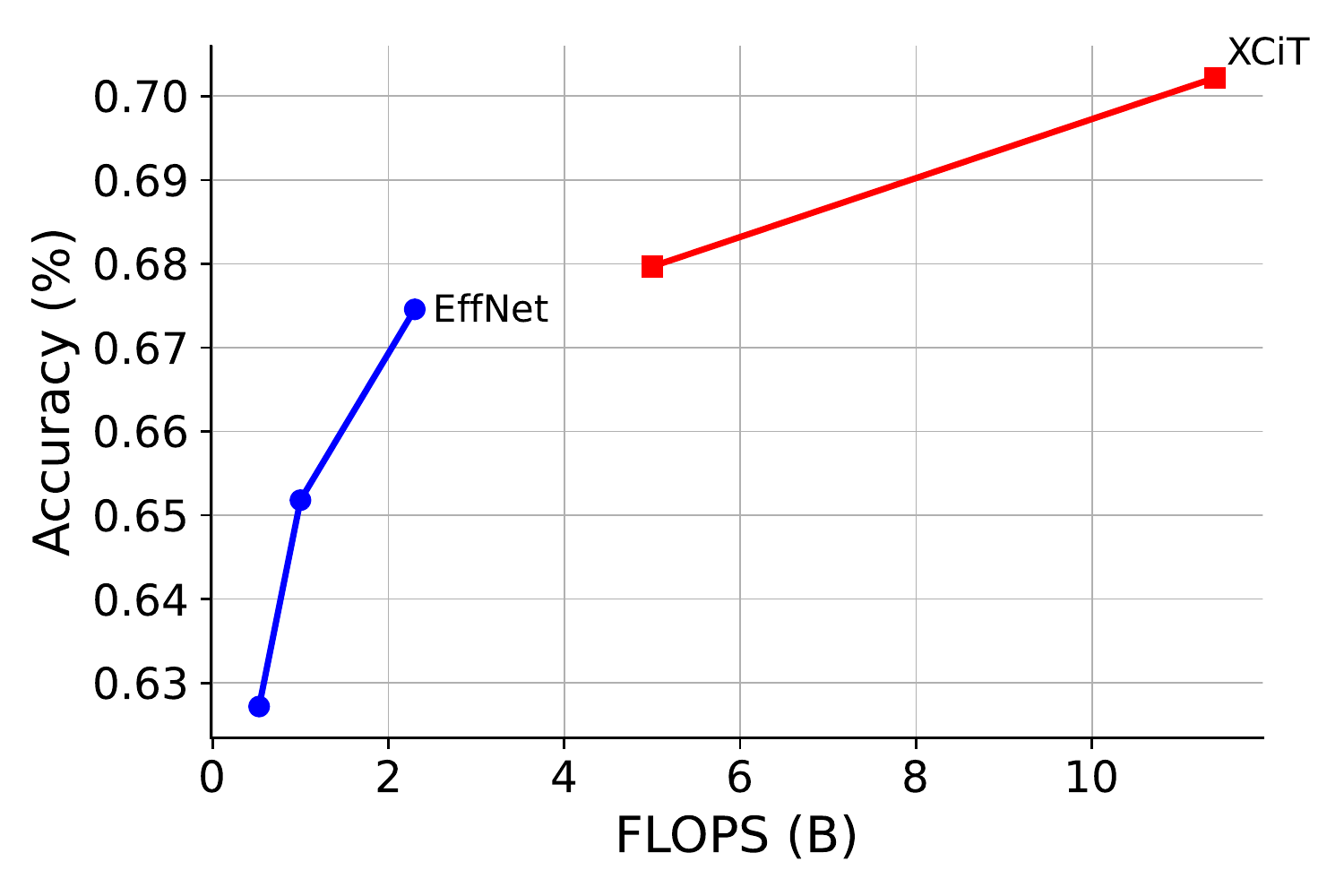}}
  \subfloat[Accuracy vs Number of Parameters]{\label{fig:sig53-acc-params}\includegraphics[width=0.5\textwidth]{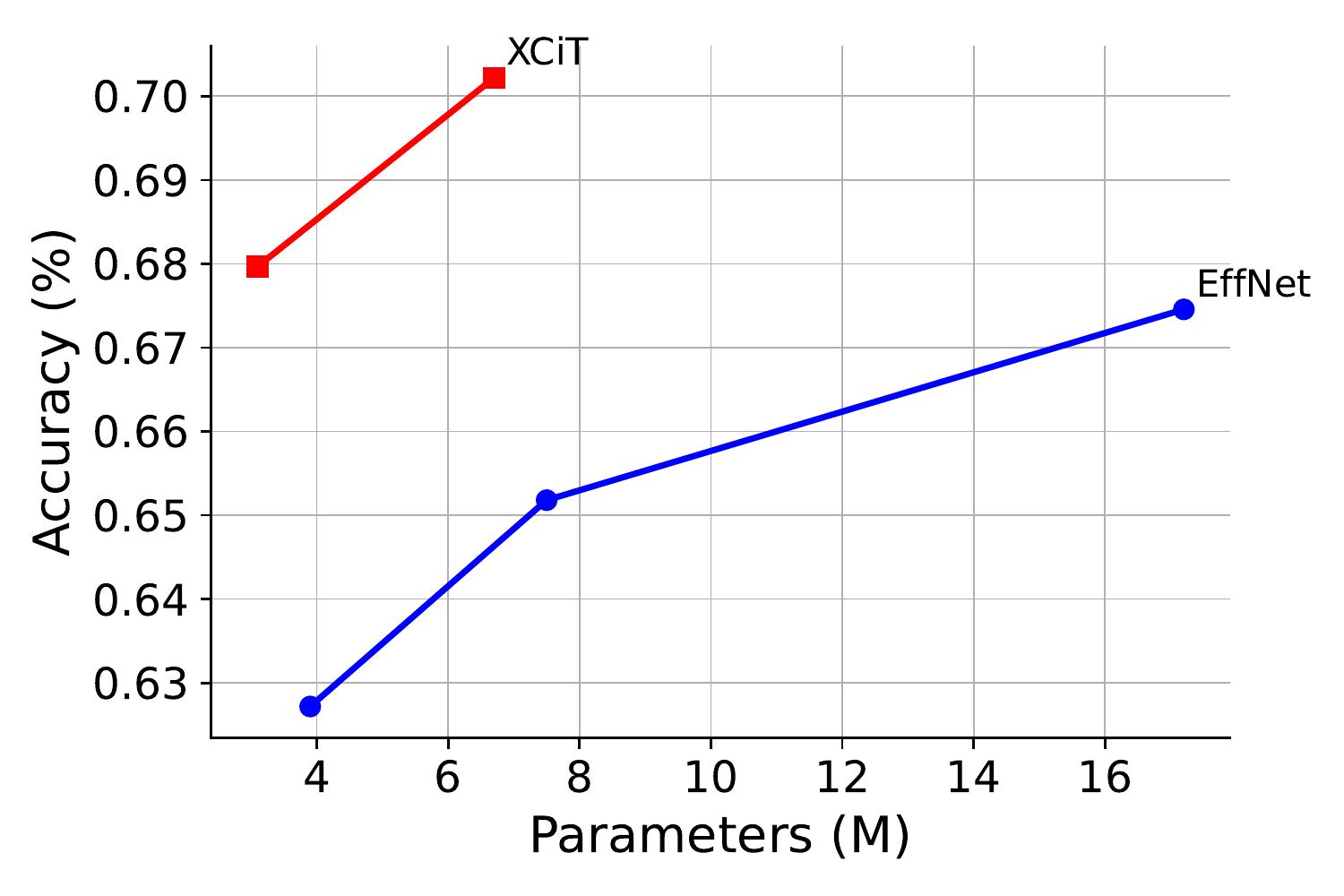}}
  \caption{
    \subref{fig:sig53-acc-flops} compares the EfficientNet and the XCiT models using the Sig53 impaired dataset.  
    Note that the XCiT models perform better but require more floating point operations.  
    \subref{fig:sig53-acc-params} shows that even with lower parameter counts, the XCiT networks perform better than EfficientNets.}
  \label{fig:sig53-results}
\end{figure*}

\subsection{Static Dataset Results}
All models are trained for 1 million steps with a batch size of 32 using the AdamW optimizer \cite{loshchilov2017decoupled} with a weight decay of 0.04.
For EfficientNets, the learning rate is set to 5e-4, and for XCiT networks the learning rate is set to 2.5e-4. 
Because our data is natively complex-valued, we convert them into real signals of 2 channels, using the full Sig53 input sequences of length 4096. 
EfficientNet inputs are downsampled by a factor of 32 internally while the inputs to the XCiT networks are downsampled by a factor of 2 before the body of the network.
Both of these networks were natively built for images, so 2D elements are converted to 1D for the temporal signals.
For the initial experiments, we do not use any data augmentations, training directly on the static Sig53 impaired training set.
Data augmentations, which are used later, are explained in the next section.
After the full training session is completed, the best model checkpoint is loaded and evaluated against the Sig53 impaired validation set, yielding the results seen in \cref{tab:sig53-results} and \cref{fig:sig53-results}.

Within both the EfficientNet and XCiT results, we see the desired model scaling effects where accuracy improves as the network increases in size.
This signifies that the Sig53 dataset is sufficiently challenging for this ML network research.
We can also see that XCiT significantly outperforms EfficientNets in terms of parameter efficiency.
An important point of note is that within the vision domain, to match the accuracies of EfficientNets, XCiT networks require both a heavily regularized training schedule and a convoluational teacher \cite{el2021xcit}.
Here, we do not augment at all and do not use a convolutional teacher, but we are able to outperform EfficientNets by a large margin.
We hypothesize this difference comes from the fact that chunking is not required here, as our signal consists of 4096 elements, compared to the tens/hundreds of thousands typical in the vision domain. 
This chunking introduces difficult dependencies between samples that are likely very difficult to recover. 
Specifically, the samples on the edges of chunk boundaries are locally close originally but become far in both time and channel space. 
ConvNets do not suffer from this due to their natural locality bias and from being able to examine the image directly in two dimensions.

\begin{figure}[!btp]
  \vskip 0.1in
  \begin{center}
  \centerline{\includegraphics[width=0.45\textwidth]{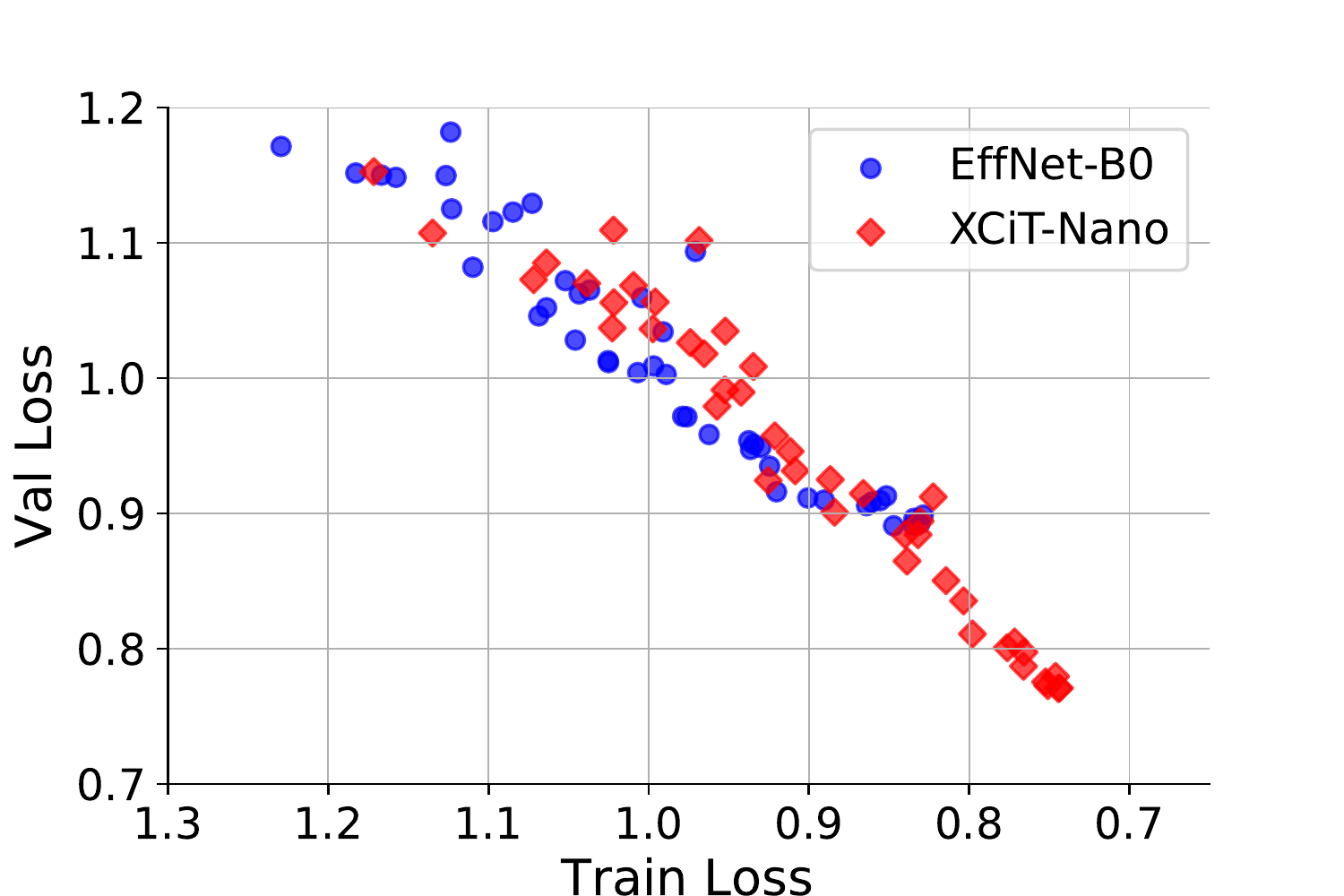}}
  \caption{
    A graph of the training loss vs the validation loss is shown for the EfficientNet-B0 and XCiT-Nano networks.  
    Plotted points on this graph generally travel from the top left to the bottom right as epoch counts increase.  
    Note that for the EfficientNet-B0 network, training and validation losses are not able to decrease to the levels of the XCiT-Nano network, 
    even though these networks are similar in size.
    }
  \label{fig:effnet-xcit-train-val}
  \end{center}
  \vskip -0.3in
\end{figure}

\begin{table*}[t]
  \caption[fontsize=12pt]{Effects of augmentations and the online dataset are shown here.}
  \label{tab:static-aug-online-results}
  \vskip 0.1in
         \begin{center}
          \centering
          \begin{tabular}{l|l|l|l|l}
              \toprule[1.5pt]
           Model & Dataset & Params. & FLOPS & Accuracy  \\ \midrule
            EfficientNet-B4 & Static & 17.2M & 2.3B & 67.46\% \\
            EfficientNet-B4 & Augmented & 17.2M & 2.3B & 67.78\% \\
            EfficientNet-B4 & Online & 17.2M & 2.3B & 69.73\% \\ \midrule
            XCiT-Tiny12 & Static & 6.7M & 11.4B & 70.22\% \\
            XCiT-Tiny12 & Augmented & 6.7M & 11.4B & 71.15\% \\
            \textbf{XCiT-Tiny12} & \textbf{Online} & \textbf{6.7M} & \textbf{11.4B} & \textbf{71.16\%} \\
          \bottomrule[1.5pt]
          \end{tabular}
     \end{center}
  \vskip -0.1in
\end{table*}

\begin{figure*}[h]
  \centering
  \begin{tabular}{lr}
      \subfloat[Accuracies for static, augmented, and online data]{\label{fig:static-aug-online}
      \includegraphics[width=.45\textwidth]{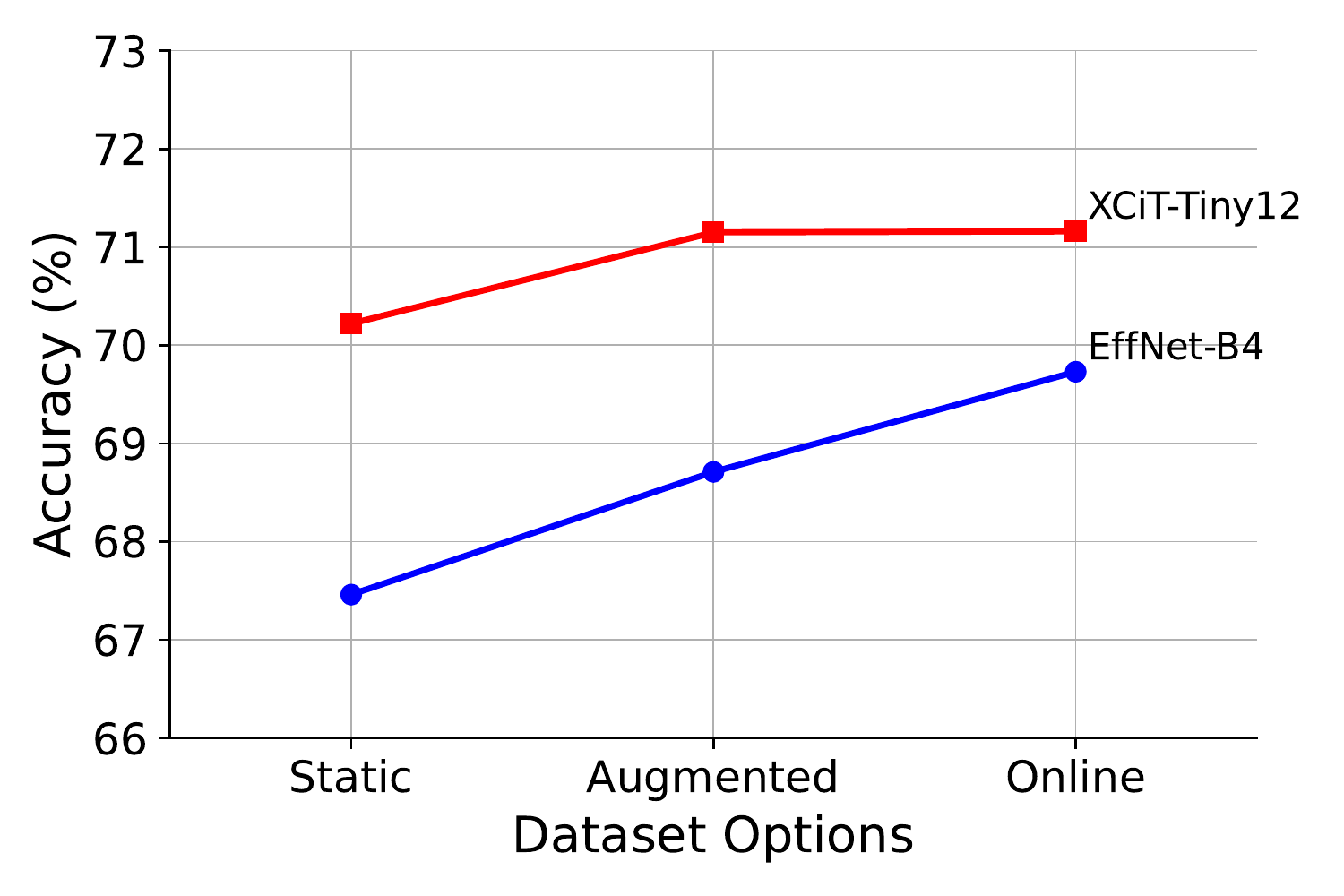}} &
      
      \subfloat[EfficientNet static, augmented, and online loss]{\label{fig:static-aug-online-loss}
      \includegraphics[width=.45\textwidth]{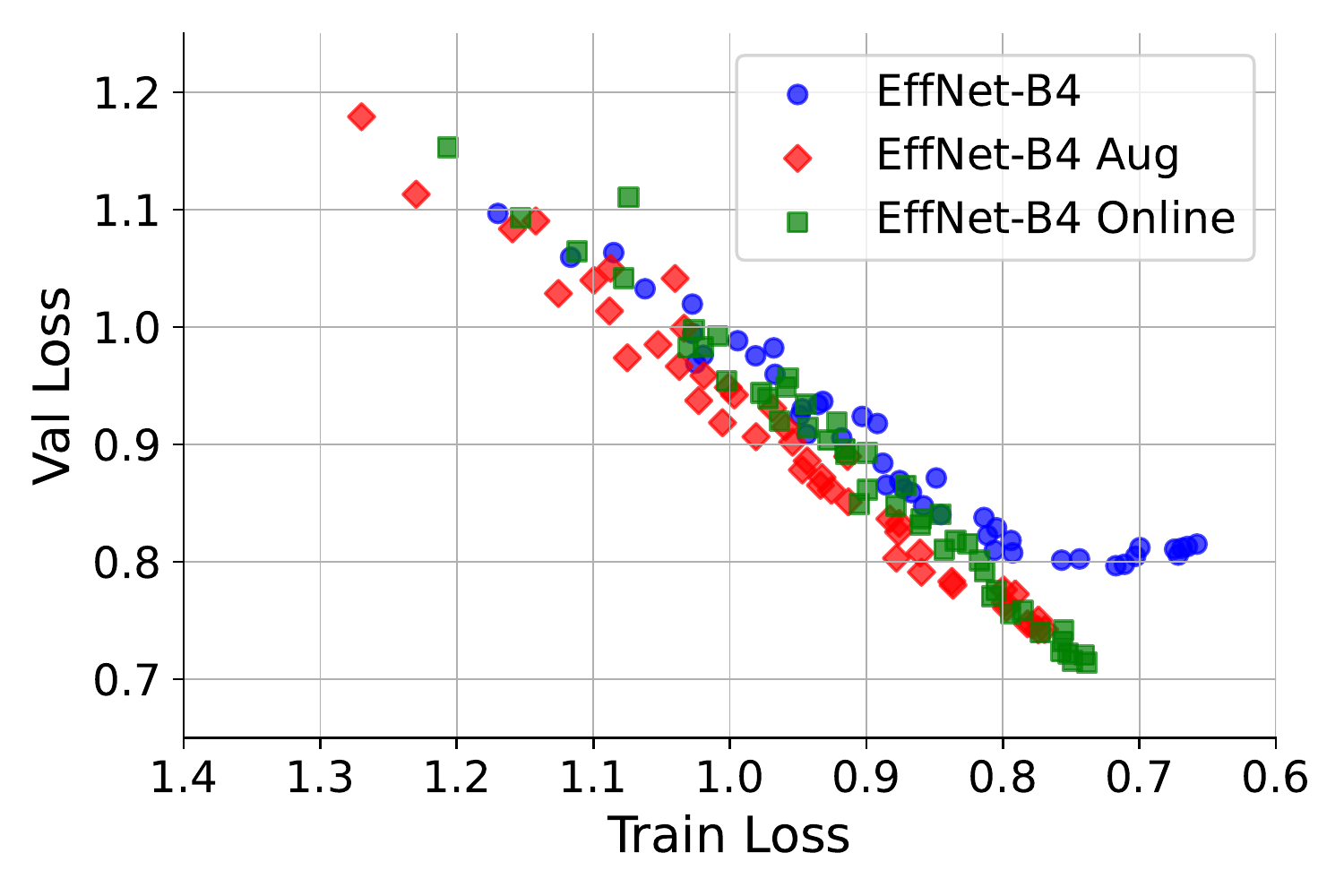}}
  \end{tabular}
  \caption{
    \subref{fig:static-aug-online} shows that training with augmented and online data has a positive effect on model performance.  
    With \subref{fig:static-aug-online-loss}, note that overfitting occurs when using static data as training loss continues to reduce without a corresponding reduction in the validation loss.  
    Using augmentations in training data can overcome this overfitting.  
    Using online data reduces the training and validation losses even further.
    }
\end{figure*}

The attention mechanism of the XCiT networks also seems to enable increased generalizability (lower validation loss) as well as learning capacity (lower training loss) when compared to EfficientNets.
In \cref{fig:effnet-xcit-train-val}, we see XCiT-Nano learning at a linear rate between the training and validation loss.
However, EfficientNet-B0, a network roughly the same size as XCiT-Nano, begins overfitting once the validation loss reaches 0.9.
As a result, it never achieves the same training or validation loss as the similarly sized XCiT-Nano network.

\subsection{Augmentations and Online Dataset}
\label{sec:AugOnline}
We next experiment with the effects of data augmentations.
At each epoch, new augmentations (listed below) are applied randomly to each example of the impaired training data examples.
Since the dataset is synthetic, we intentionally omit any augmentations used as part of the generating distribution of impairments.
Without this constraint, we would essentially be increasing the dataset size, changing the nature of the experiments.
Within TorchSig, we create a new RandAugment \cite{cubuk2020randaugment} for the signals domain, and within this RandAugment, we insert the following augmentations:
\begin{itemize}
  \setlength\itemsep{0em}
  \item Spectral inversion,
  \item Channel swap,
  \item Amplitude reversal,
  \item CutOut \cite{devries2017improved},
  \item Drop samples,
  \item Quantize,
  \item Magnitude rescale,
  \item PatchShuffle \cite{kang2017patchshuffle}, and
  \item Identity.
\end{itemize}
In addition to RandAugment, we also randomly apply a time reversal data augmentation prior to the RandAugment effects.
Time reversal also has the side effect of performing a spectral inversion, which we undo by default.
Finally, we normalize the data post-augmentations to match the normalization used in the static data experiments.
For more information on these augmentations, see Appendix \ref{sec:appendix_tools}.

\begin{figure*}[!ht]
  \centering
  \begin{tabular}{lr}
    \subfloat[Accuracy versus SNR]{\label{fig:acc-vs-snr}
    \includegraphics[width=.45\textwidth]{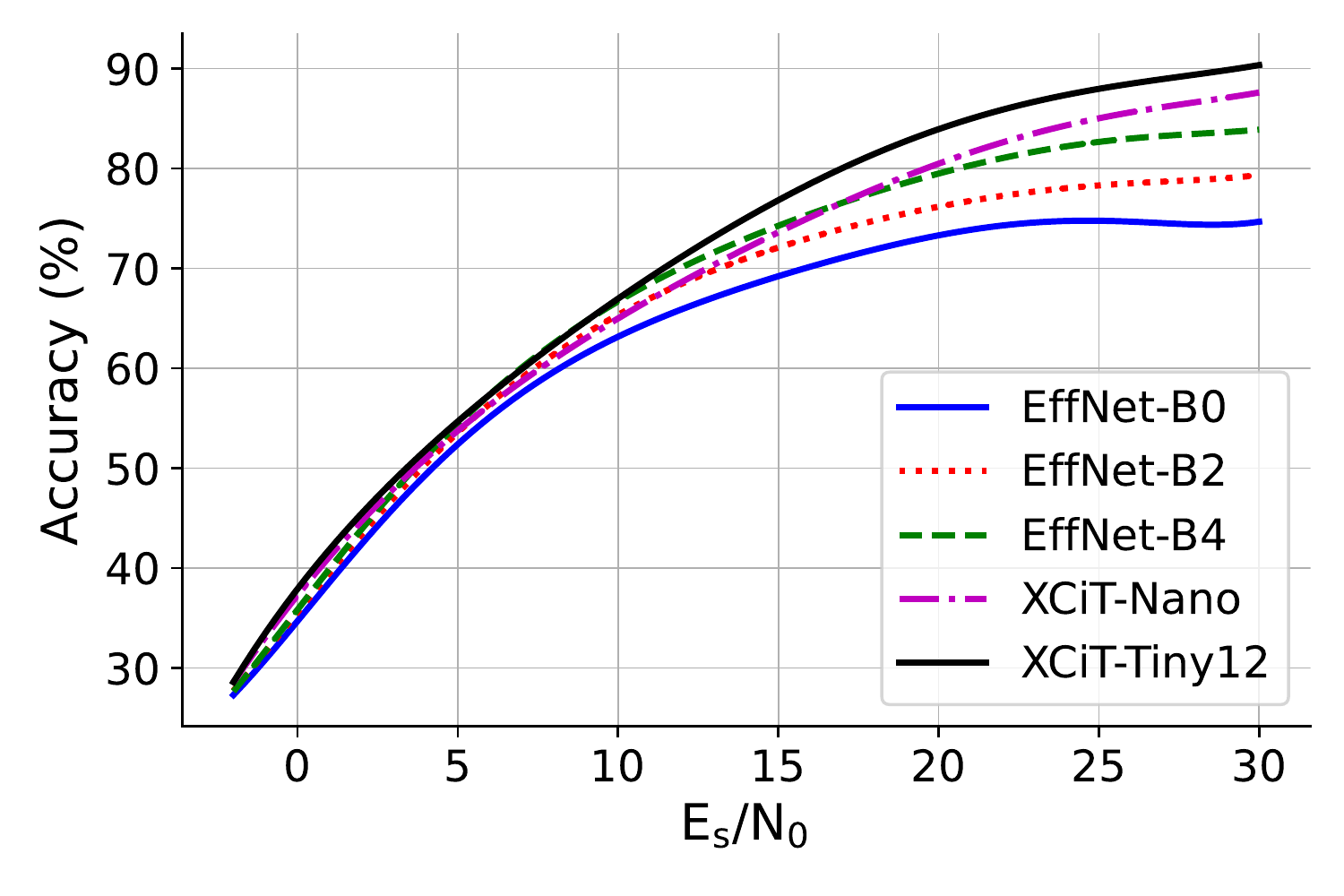}} &
    
    \subfloat[EfficientNet and XCiT modulation family accuracies]{\label{fig:mod-family-accs}
    \includegraphics[width=.45\textwidth]{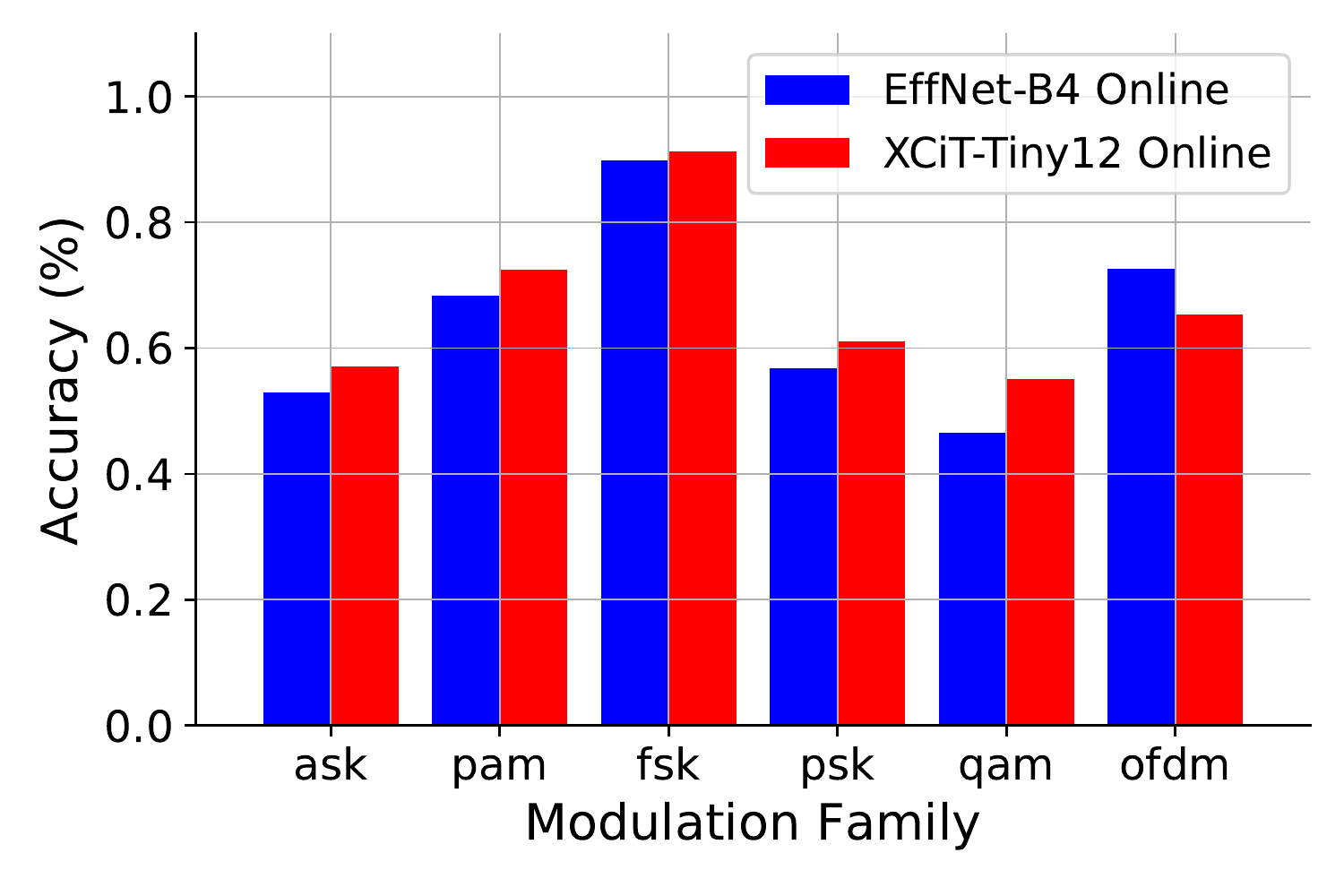}} \\

    \subfloat[EfficientNet-B4 modulation family confusion matrix]{\label{fig:effnet-mod-family-conf-matrix}
    \includegraphics[width=.45\textwidth]{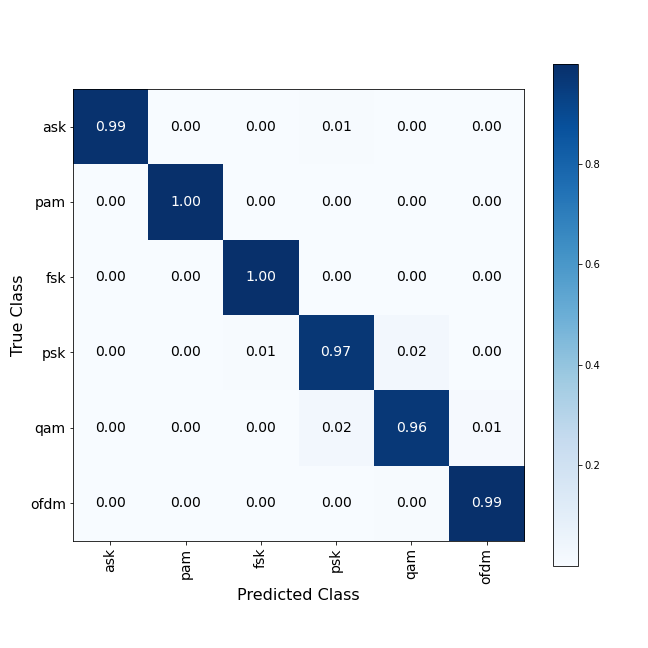}} &

    \subfloat[XCiT-Tiny12 modulation family confusion matrix]{\label{fig:xcit-mod-family-conf-matrix}
    \includegraphics[width=.45\textwidth]{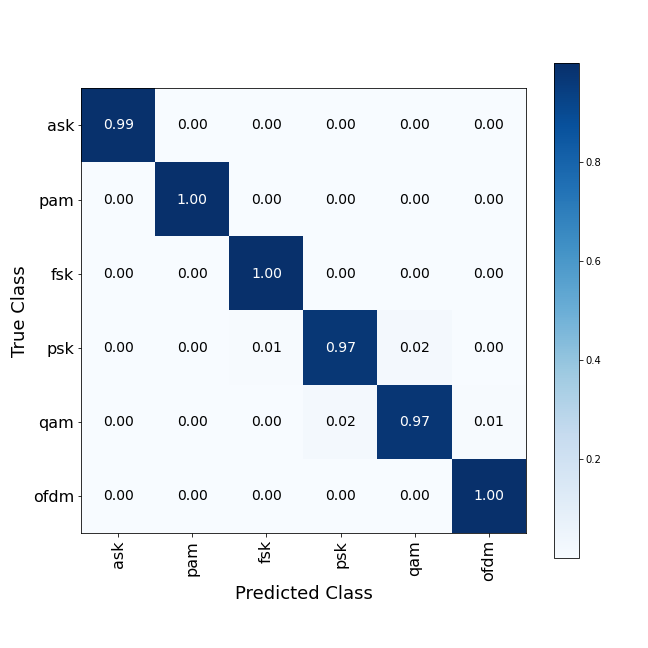}}

  \end{tabular}
  \caption{
    \subref{fig:acc-vs-snr} shows accuracies of the various models as a function of SNR.  
    \subref{fig:mod-family-accs} shows that the XCiT-Tiny12 model outperforms the EfficientNet-B4 model within all signal families except the OFDM family.  
    The confusion matrices in \subref{fig:effnet-mod-family-conf-matrix} and \subref{fig:xcit-mod-family-conf-matrix} show that performance is nearly perfect for these models when looking at modulation families.  
    This suggests that almost all errors occur within modulation families.
    }
\end{figure*}

In addition to experiments with data augmentations, we take advantage of the synthetic nature of the dataset to experiment with an online dataset.
The generation of the dataset is fast enough such that during training, the data can be generated on the fly, allowing the user to train with effectively new data at every step without the risk of overfitting.
Note that although new examples are created from new bit sequences, pulse shaping, and impairments in each step, the augmentations from the above list are not applied to the online dataset.  

We train an EfficientNet-B4 network and an XCiT-Tiny12 network with the data augmentations and then again with the online data and compare against the static Sig53 dataset.
During the augmented and online training sessions, a similar training schedule as was done in the static dataset cases is followed.
Thus, the online data models were trained with 32 million unique examples.
In \cref{fig:static-aug-online}, we show the incremental performance with each of these changes, demonstrating the benefits of the TorchSig toolkit's signal domain augmentations as well as the benefits provided from an online dataset.
In \cref{fig:static-aug-online-loss}, we show the overfitting with EfficientNet-B4's static data experiment is no longer present once augmentations are applied, and with the online data, the network achieves a lower validation loss than the other data methods, as expected.
Across all experiments conducted within this work, XCiT-Tiny12 with online data achieved the highest accuracy at 71.16\% (\cref{tab:static-aug-online-results}).
We claim this result is a competitive baseline for future work to use as a benchmark for comparative performance analysis.
We expect others will be able to improve on this result by scaling network size further, experimenting with additional architectural improvements, and conducting more thorough hyperparameter and training schedule studies.

\begin{figure*}[h]
  \vskip 0.1in
  \begin{center}
  \centerline{\includegraphics[width=0.825\textwidth]{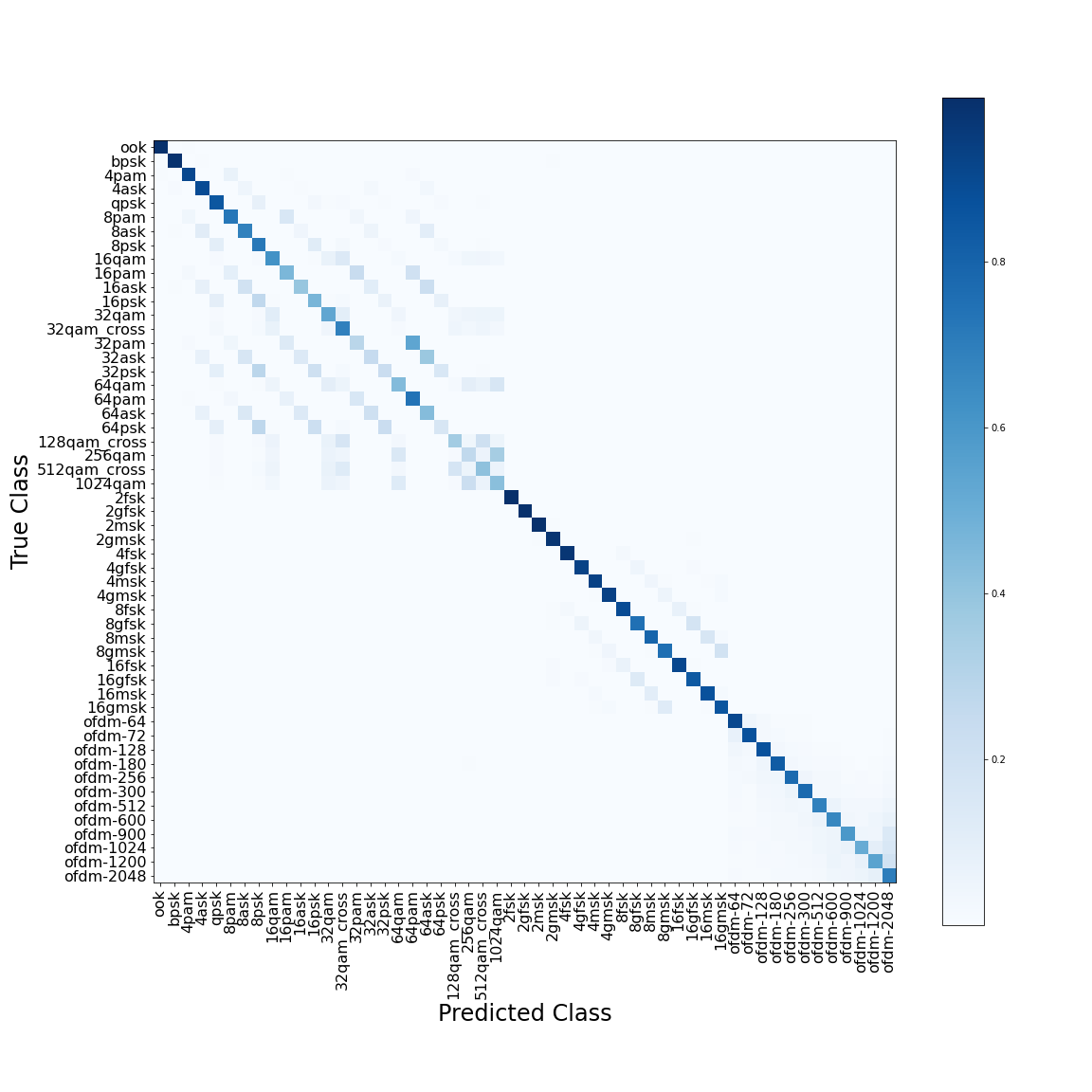}}
  \caption{
    The EfficientNet-B4 online data confusion matrix is shown with a total accuracy of 69.73\%. 
    Note that most of the confusion occurs with differing orders of modulations within the same family.
    Performance within the FSK family appears relatively strong.
    }
  \label{fig:effnetb4-conf-matrix}
  \end{center}
  \vskip -0.3in
\end{figure*}

\begin{figure*}[h]
  \vskip 0.1in
  \begin{center}
  \centerline{\includegraphics[width=0.825\textwidth]{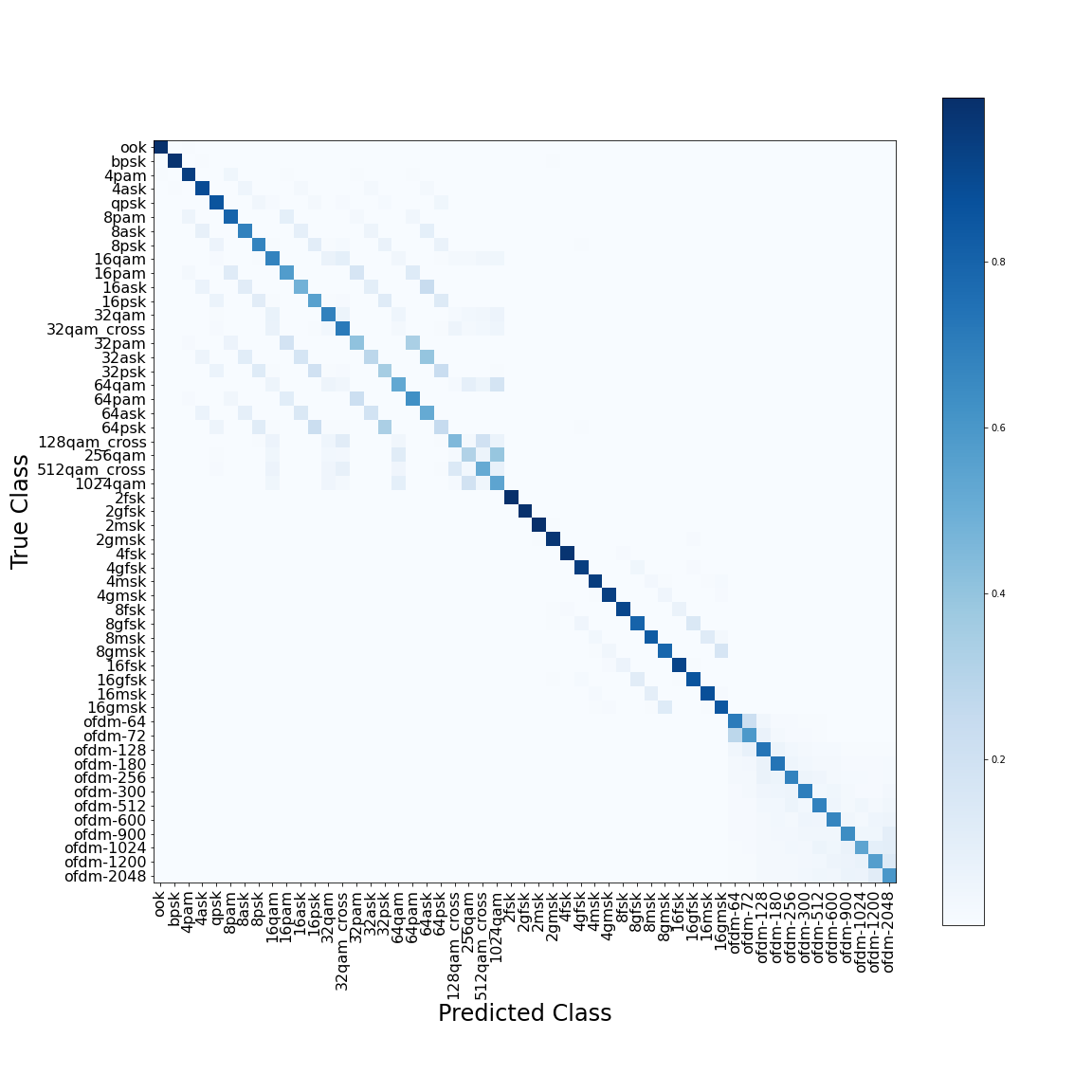}}
  \caption{
    The XCiT-Tiny12 online data confusion matrix is shown with a total accuracy of 71.16\%.  
    Note that, like the EfficientNet-B4's confusion matrix, errors tend to occur primarily within related modulation families' classes, and 
    performance of the FSK family appears relatively strong.  
    This corresponds with the bar graph in \cref{fig:mod-family-accs}.
    }
  \label{fig:xcit-tiny-conf-matrix}
  \end{center}
  \vskip -0.3in
\end{figure*}

\subsection{Performance Analysis}
While we believe a single performance metric is preferred for comparative analysis between networks, the accuracy versus SNR curves used in past RFML publications provide useful insights into how trained models perform.
In \cref{fig:acc-vs-snr}, we show the accuracy versus SNR performance of all of the models trained with the static Sig53 dataset.
As expected, network performance increases with SNR.  
Network performance also scales with the architecture size and complexity, especially at higher power levels.
This demonstrates that differentiating the signal classes within Sig53 is not only a difficult task at lower SNRs but also at the higher SNRs as well.
This difficulty makes Sig53 a good candidate for showing useful performance spreads across advanced network architectures.
Comparing the similarly sized EfficientNet-B0 and XCiT-Nano curves shows that a majority of the XCiT performance gain comes at the higher power levels.
\cref{fig:mod-family-accs} also shows XCiT-Tiny12's accuracies are higher than EfficientNet-B4's accuracies for all modulation families except for OFDM signals, with the largest improvement shown for QAM signals.
These results suggest the transformer's attention mechanism provides the most benefit from simultaneously viewing a higher quantity of symbols than the EfficientNet architecture due to its convolutional layers' locality bias.
Even with these errors, confusion matrices in \cref{fig:effnet-mod-family-conf-matrix} and \cref{fig:xcit-mod-family-conf-matrix} show near perfect performance, suggesting misclassifications almost always occur within modulation families.
The full 53 class confusion matrices for these 2 networks are shown in \cref{fig:effnetb4-conf-matrix} and  \cref{fig:xcit-tiny-conf-matrix}.

\section{Conclusion}
\label{sec:conclusion}

In this work, we demonstrate that limitations exist within RadioML, and we introduce the Sig53 dataset.
The Sig53 dataset is meant to serve as the RF domain's evolution over RadioML.
We share reproducible experimental results modifying SoTA scalable vision domain networks: EfficientNet and XCiT architectures. The use of XCiT is one of the earliest applications of transformer neural networks to the RF domain.
We also introduce TorchSig, a general-purpose complex-valued signal processing machine learning toolkit. Torchsig is capable of generating the Sig53 dataset, providing domain-specific data augmentations, and serving as an RFML framework for reproducible collaboration within the RFML community.
We demonstrate quantitative benefits of leveraging TorchSig's domain-tailored data augmentations. TorchSig's synthetic signal generation enables supervised machine learning tasks at large data scales.
Regardless of domain, very few labeled datasets exist at such a large scale, allowing our work to help democratize experimentation.
We have demonstrated sound transferability from networks in the vision domain.
We hope the accessibility of this dataset and toolkit allows for experimentation to improve results within the signals domain.
We invite others to build upon our research through open source RFML collaboration.

\section*{Software and Data}
The TorchSig toolkit is available at \url{https://github.com/torchdsp/torchsig} with additional documentation available at \url{https://torchsig.com}.
The Sig53 dataset can be generated using the TorchSig toolkit, or it can be downloaded through the TorchSig website.

\section*{Acknowledgements}
This research was funded by the Laboratory for Telecommunication Sciences.

\bibliographystyle{icml2022}
\bibliography{000_sig53}

\newpage
\appendix
\onecolumn
\section{Appendix}
\subsection{Dataset Appendix}
\label{sec:appendix_dataset}

\textbf{Full Class List.} The full class list, along with the modulation family of each class and the default class index when using Sig53, is shown in \cref{tab:class-list}

\begin{table}[H]
  \caption[fontsize=9pt]{Full Sig53 Class List}
  \label{tab:class-list}
  \centering
  \vskip 0.1in
  \begin{center}
  \begin{small}
  \begin{sc}
  \begin{tabular}{l|l|l}
    \toprule[1.5pt]
    Class Name & Modulation Family & Class Index  \\ 
    \midrule
    4ASK & ASK Family & 3 \\
    8ASK & ASK Family & 6 \\
    16ASK & ASK Family & 10  \\
    32ASK & ASK Family & 15  \\
    64ASK & ASK Family & 19  \\
    OOK (On-Off Keying) & PAM Family & 0  \\
    4PAM & PAM Family & 2 \\
    8PAM & PAM Family & 5 \\
    16PAM & PAM Family & 9 \\
    32PAM & PAM Family & 14 \\
    64PAM & PAM Family & 18 \\
    BPSK (Binary Phase-Shift Keying) & PSK Family & 1 \\
    QPSK (Quadrature Phase-Shift Keying) & PSK Family & 4 \\
    8PSK & PSK Family & 7 \\
    16PSK & PSK Family & 11 \\
    32PSK & PSK Family & 16 \\
    64PSK & PSK Family & 20 \\
    16QAM & QAM Family & 8 \\
    32QAM & QAM Family & 12 \\
    32QAM\_Cross & QAM Family & 13 \\
    64QAM & QAM Family & 17 \\
    128QAM\_Cross & QAM Family & 21 \\
    256AM & QAM Family & 22 \\
    512AM\_Cross & QAM Family & 23 \\
    1024AM & QAM Family & 24 \\
    2FSK & FSK Family & 25 \\
    2GFSK & FSK Family & 26 \\
    2MSK & FSK Family & 27 \\
    2GMSK & FSK Family & 28 \\
    4FSK & FSK Family & 29 \\
    4GFSK & FSK Family & 30 \\
    4MSK & FSK Family & 31 \\
    4GMSK & FSK Family & 32 \\
    8FSK & FSK Family & 33 \\
    8GFSK & FSK Family & 34 \\
    8MSK & FSK Family & 35 \\
    8GMSK & FSK Family & 36 \\
    16FSK & FSK Family & 37 \\
    16GFSK & FSK Family & 38 \\
    16MSK & FSK Family & 39 \\
    16GMSK & FSK Family & 40 \\
    OFDM-64 & OFDM Family & 41 \\
    OFDM-72 & OFDM Family & 42 \\
    OFDM-128 & OFDM Family & 43 \\
    OFDM-180 & OFDM Family & 44 \\
    OFDM-256 & OFDM Family & 45 \\
    OFDM-300 & OFDM Family & 46 \\
    OFDM-512 & OFDM Family & 47 \\
    OFDM-600 & OFDM Family & 48 \\
    OFDM-900 & OFDM Family & 49 \\
    OFDM-1024 & OFDM Family & 50 \\
    OFDM-1200 & OFDM Family & 51 \\
    OFDM-2048 & OFDM Family & 52 \\
    \bottomrule[1.5pt]
  \end{tabular}
  \end{sc}
  \end{small}
  \end{center}
\vskip -0.1in
\end{table}

\begin{figure}[!h]
\vskip 0.1in
\begin{center}
\centerline{\includegraphics[width=0.4\textwidth]{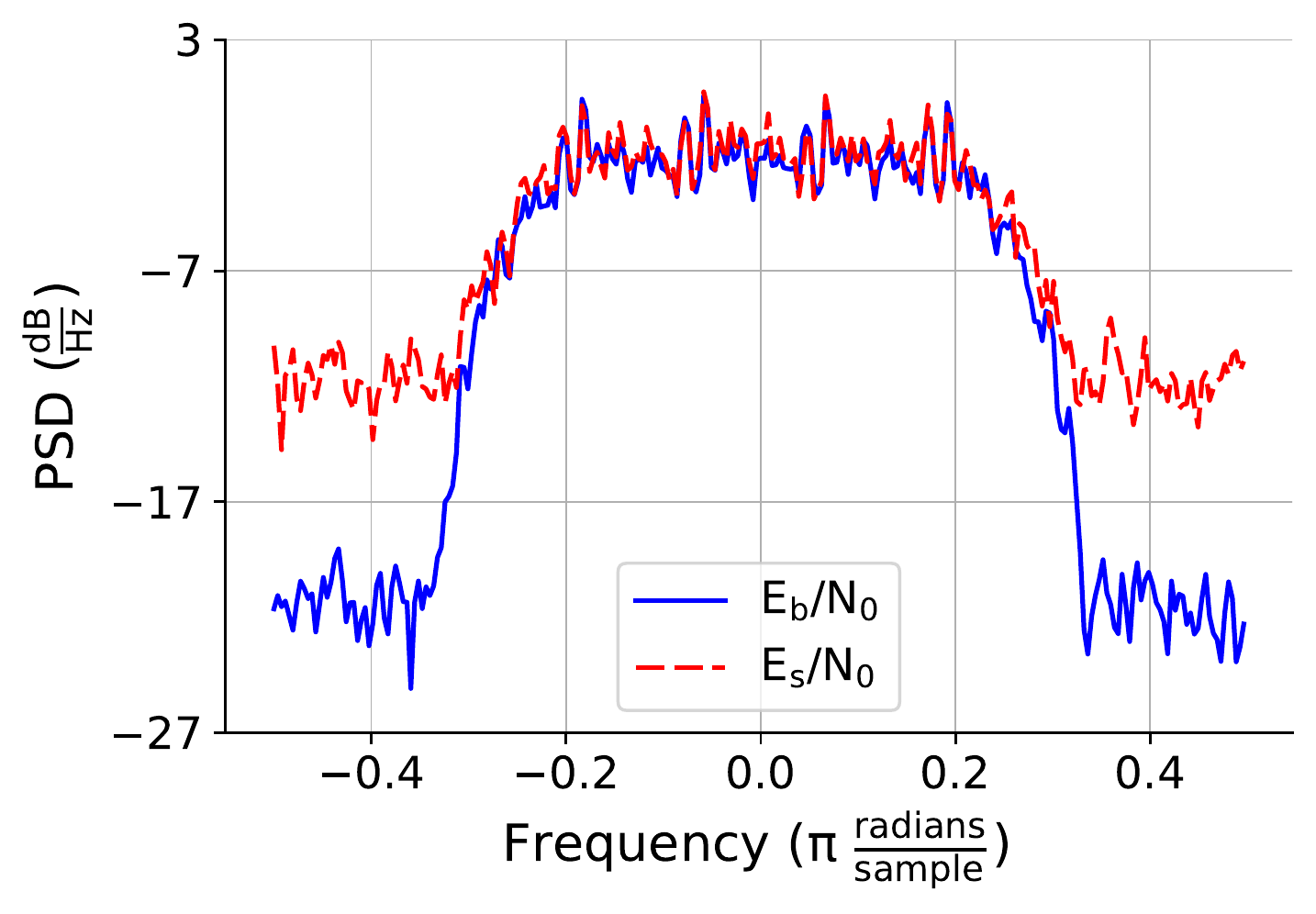}}
\caption{1024QAM $E_b / N_0$ Versus $E_s / N_0$ at 10dB. Note how the $E_b / N_0$-defined SNR signal appears much stronger than the $E_s / N_0$-defined SNR signal as seen by the lower noise floor.}
\label{fig:ebno_vs_esno}
\end{center}
\vskip -0.3in
\end{figure}

\textbf{SNR Definitions.} The difference in defining SNR levels with $E_b / N_0$ versus $E_s / N_0$ is best seen at the higher order modulations, for example, 1024QAM. 
\cref{fig:ebno_vs_esno} shows a PSD plot containing two 1024QAM signals, where one is using the 10dB $E_b / N_0$-defined SNR level and the other is using the same value of 10dB but with the $E_s / N_0$-defined SNR level.

\begin{figure*}[!h]
  \centering
  \centerline{\includegraphics[width=0.9\textwidth]{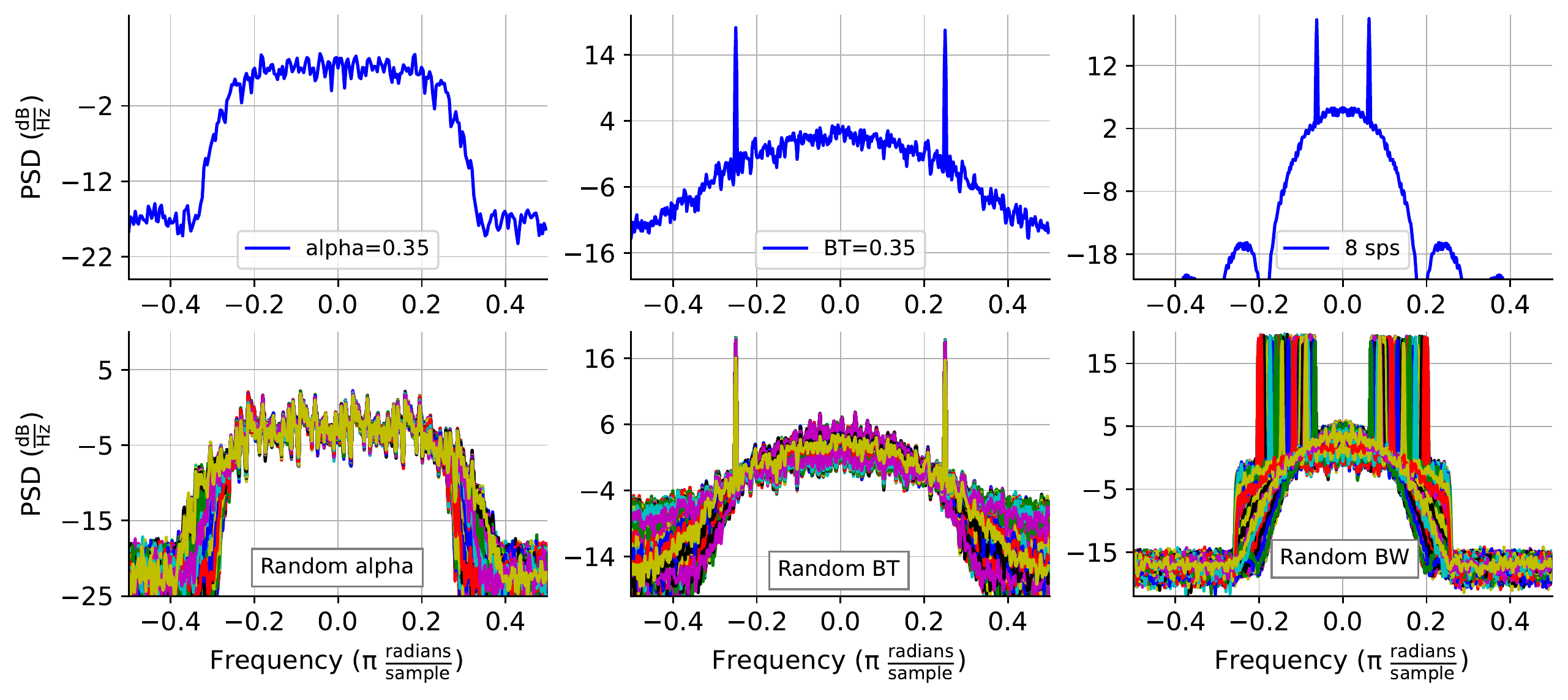}}
  \caption{
    The effects of the randomized pulse shaping can be seen by comparing the top 3 plots with the corresponding bottom 3 plots.
    On the top left, a BPSK signal has an RRC pulse shaping filter applied with an alpha value of 0.35 for the Sig53 clean dataset.
    On the bottom left, the same BPSK signal has a randomized RRC filter applied with alpha values in the range from 0.15 to 0.60 for the Sig53 impaired dataset.
    The top center plot shows a 2GFSK signal with a static BT value of 0.35 used in its Gaussian filter for the Sig53 clean dataset.
    The bottom center plot shows the same 2GFSK signal with the BT value randomized between 0.1 and 0.5 for the Sig53 impaired dataset.
    On the top right, a 2FSK signal is sampled at 8 IQ samples per symbol to avoid distortion through aliasing for the Sig53 clean dataset.
    On the bottom right, the same 2FSK signal is randomly low-pass filtered and resampled in the range 1.25 to 3.75 times the inverse of the samples per symbol for the Sig53 impaired dataset.
  }
  \label{fig:pulse_shaping}
\end{figure*}

\textbf{Random Pulse Shaping.} The impaired Sig53 dataset also implements a randomized pulse shaping process. 
\cref{fig:pulse_shaping} shows the effects of this randomized pulse shaping by first plotting the original, static pulse shapes of a BPSK signal using the RRC pulse shaping filter, a 2GFSK signal using a static Bandwidth-Time (BT) value in its Gaussian filter, and a 2FSK signal oversampled at 8 IQ samples per symbol. 
The second row of the figure then overlays randomized pulse shaping filters to show how the impaired dataset's pulse shapes differ from example to example. 
The RRC filter's alpha value is randomized uniformly between 0.15 and 0.60, 
the Gaussian filter's BT value is randomized uniformly between 0.1 and 0.5, and 
the FSK signals are randomly low-pass filtered and resampled in the range 0.15625 to 0.46875 of the total example's bandwidth.

\begin{figure*}[!h]
  \centering
  \subfloat[Original Data]{\includegraphics[width=0.35\textwidth]{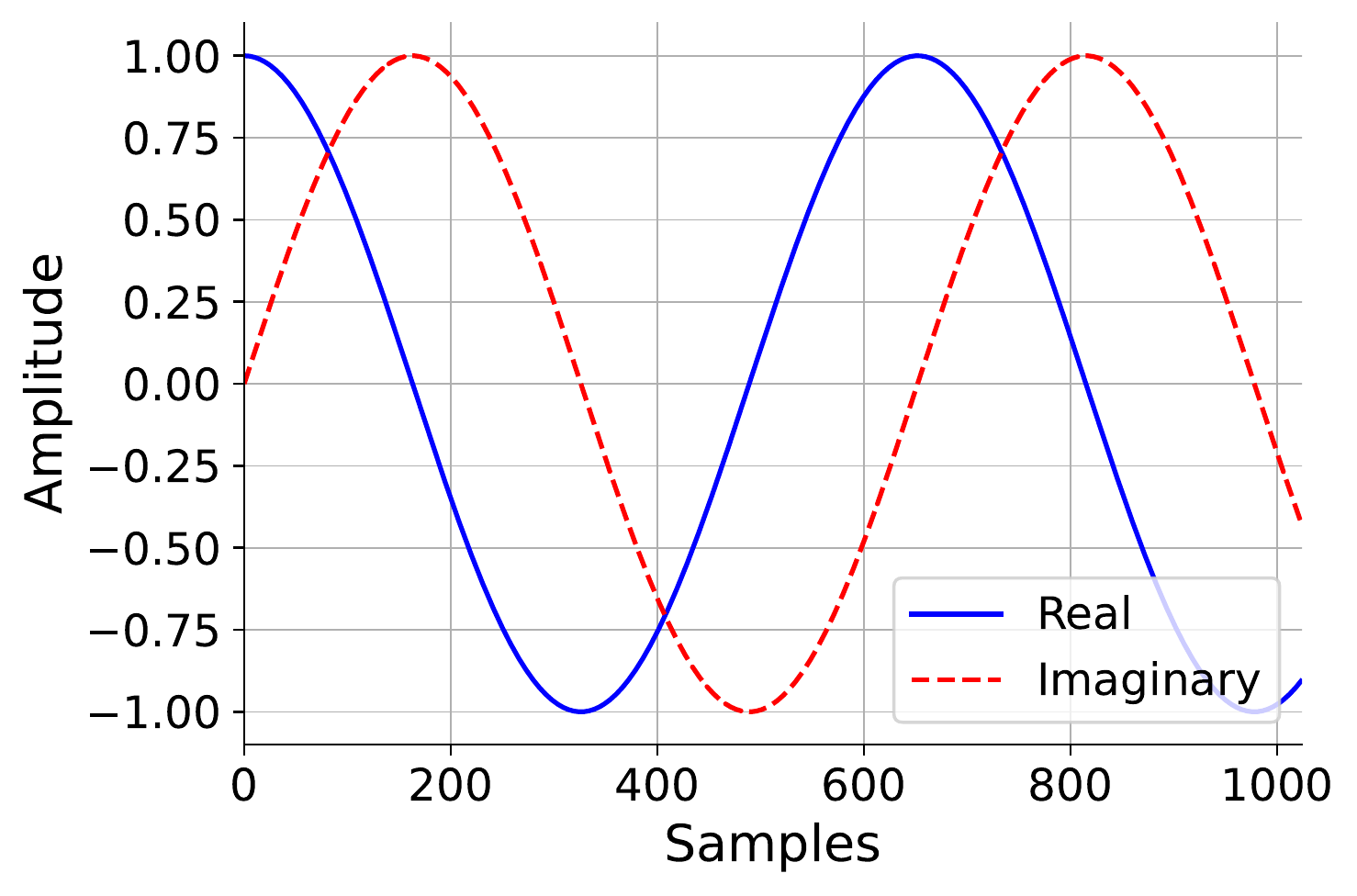}}
  \subfloat[Impaired Data]{\includegraphics[width=0.35\textwidth]{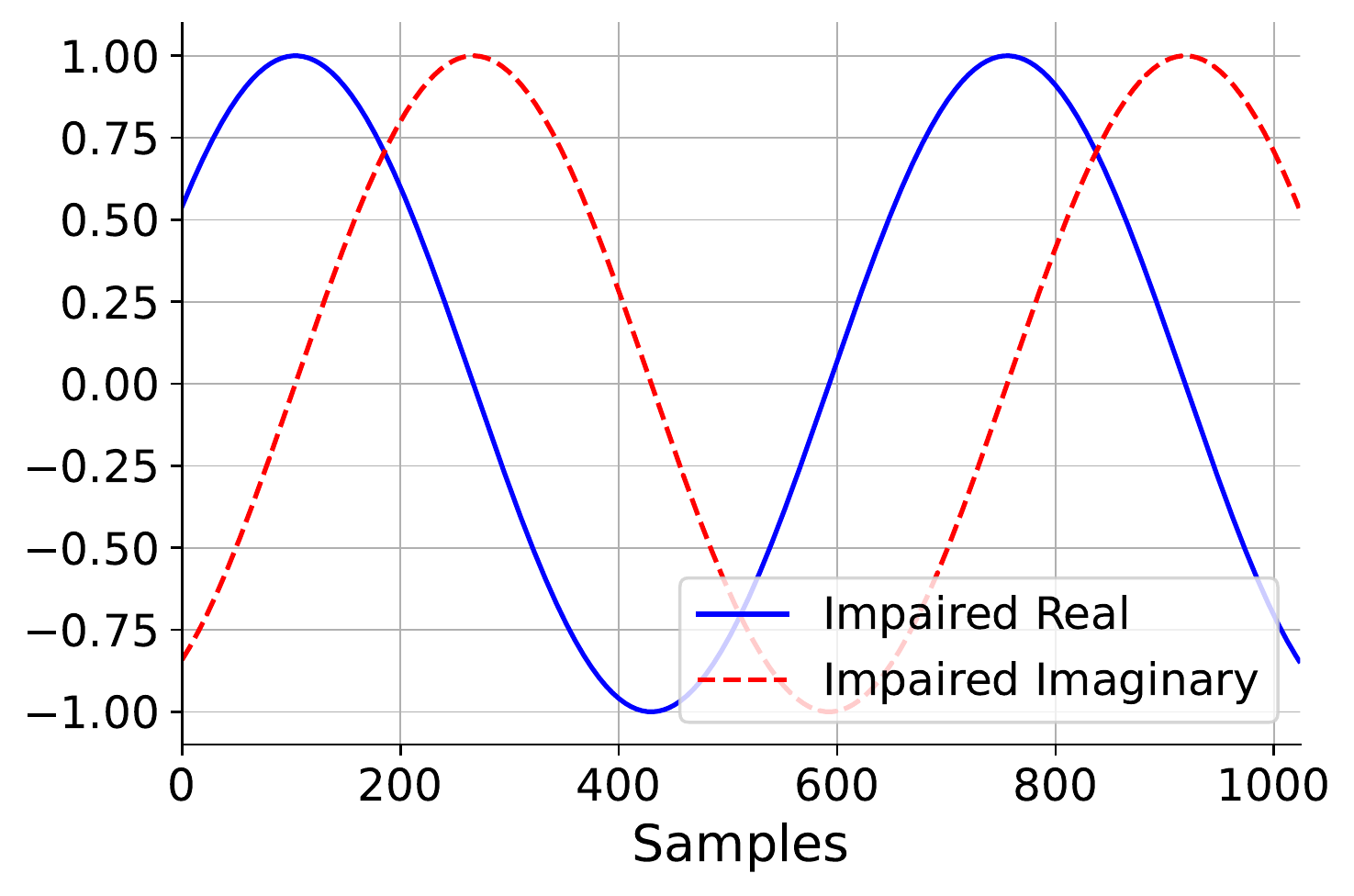}}
  \caption{Phase Shift Impairment}
  \label{fig:phase_shift}
\end{figure*}

\clearpage
\textbf{Phase Shift Impairment.} Phase shifts ($\phi$) are also applied to the impaired dataset during generation. 
An example of the phase shift impairment applied to complex sinusoids is displayed in \cref{fig:phase_shift}.

\[s'(t)=s(t)*e^{j*\phi} \]

\begin{figure*}[!h]
  \centering
  \subfloat[Original Data]{\includegraphics[width=0.35\textwidth]{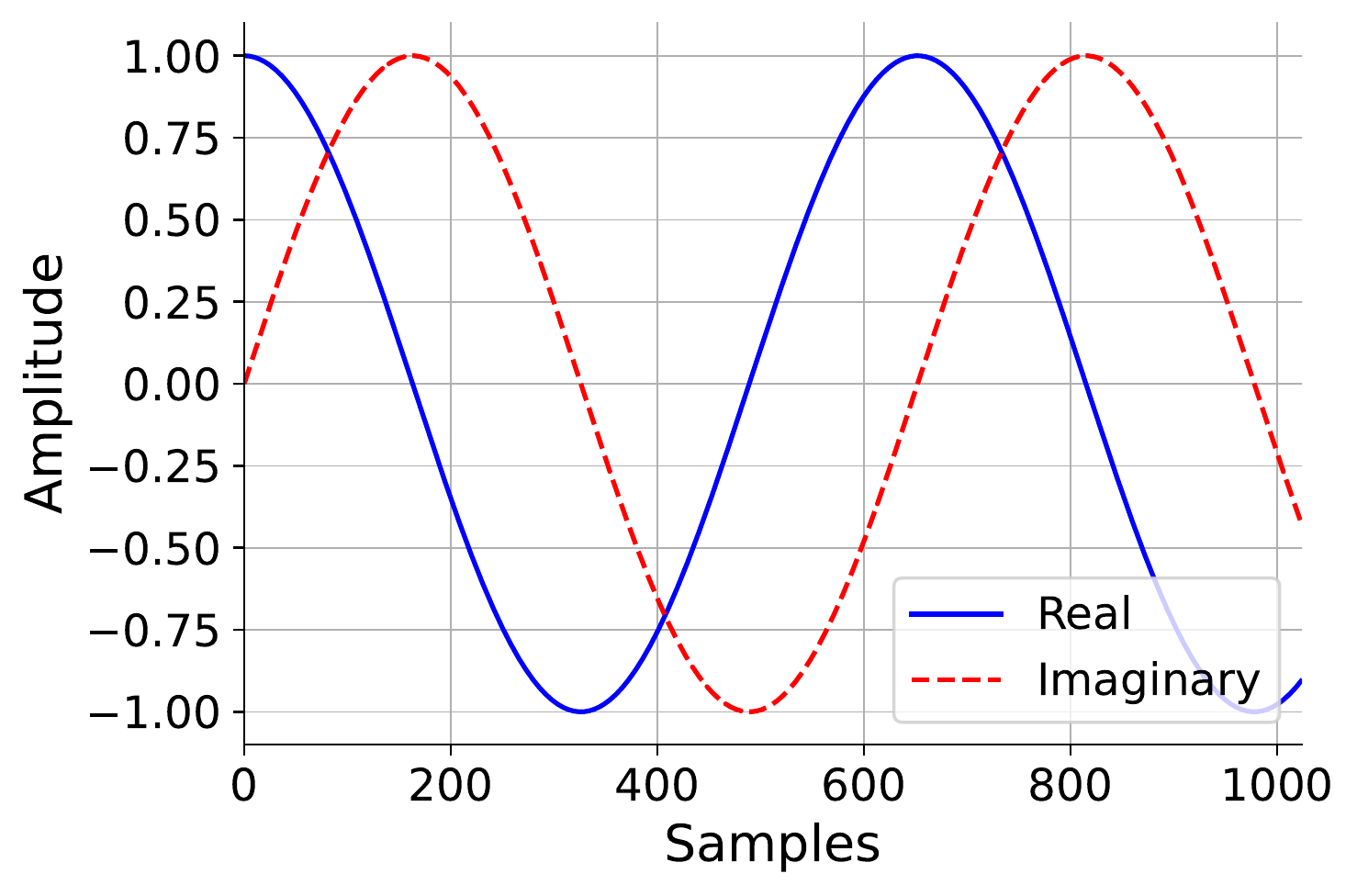}}
  \subfloat[Impaired Data]{\includegraphics[width=0.35\textwidth]{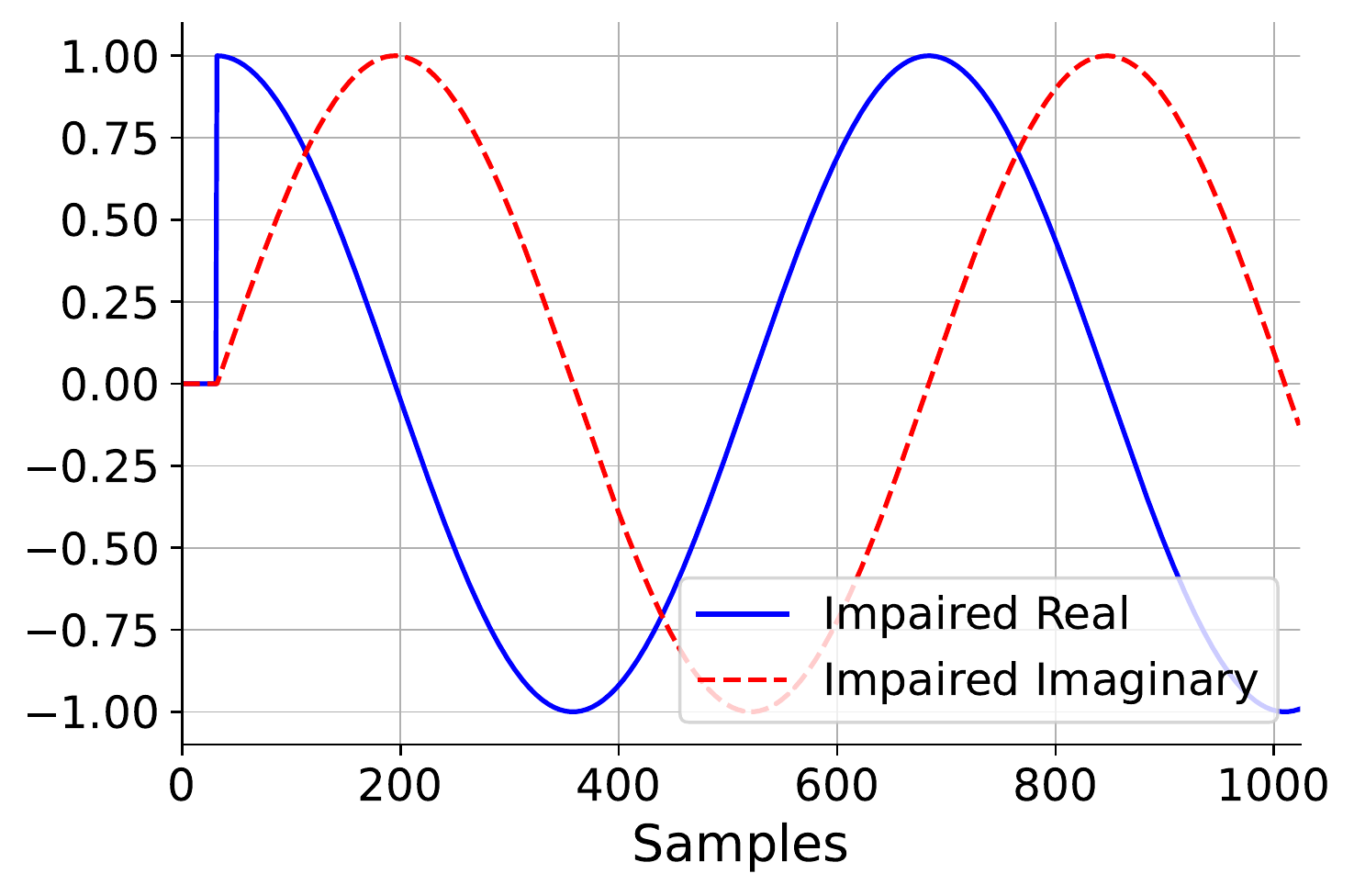}}
  \caption{Time Shift Impairment}
  \label{fig:time_shift}
\end{figure*}

\textbf{Time Shift Impairment.} The impaired Sig53 dataset also modifies data with a random time shift ($t_s$) transform. 
The time shift transform randomly applies a positive or negative shifting of the real and imaginary components, filling the new empty regions on either side with zeros (\cref{fig:time_shift}).

\[s'(t)=s(t-t_s) \]

\begin{figure*}[!h]
  \centering
  \subfloat[Original Data]{\includegraphics[width=0.35\textwidth]{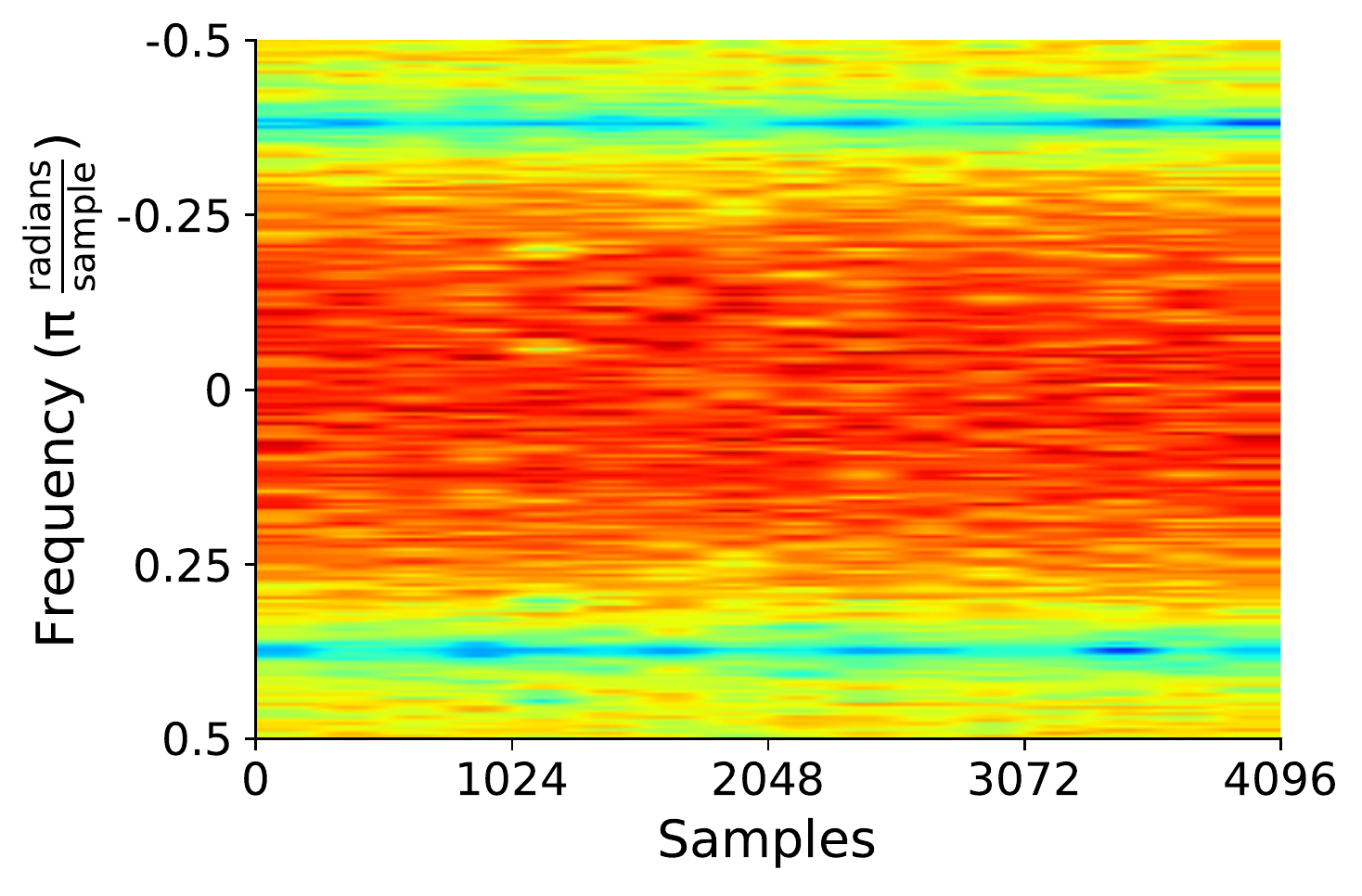}}
  \subfloat[Impaired Data]{\includegraphics[width=0.35\textwidth]{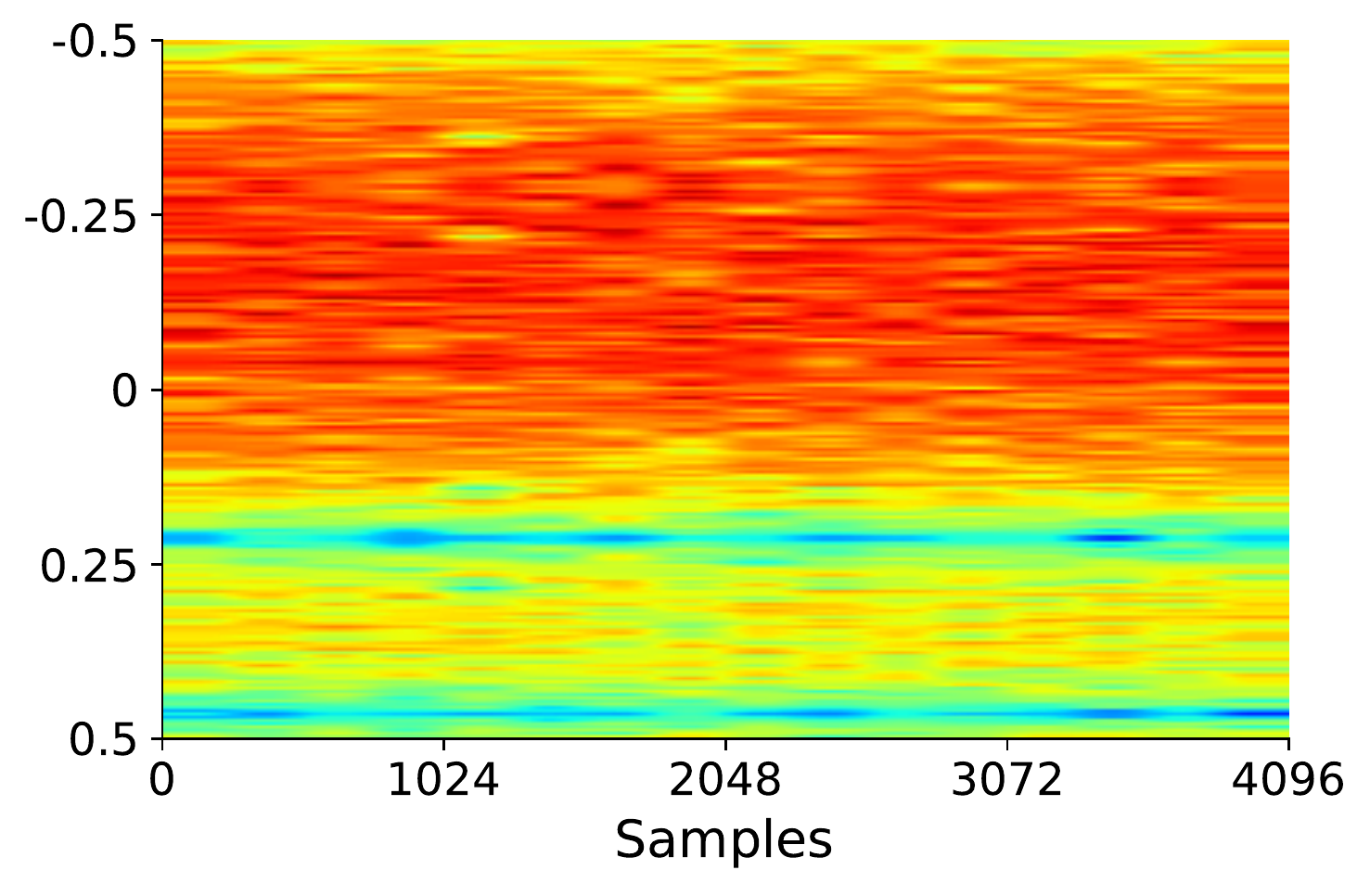}}
  \caption{Frequency Shift Impairment}
  \label{fig:freq_shift}
\end{figure*}

\clearpage
\textbf{Frequency Shift Impairment.} Random frequency shifting is also applied where the input is multiplied with a randomized complex exponential to achieve a positive or negative frequency shift (\cref{fig:freq_shift}).

\[s'(t)=s(t)*e^{2 j \pi t_s t} \]

\begin{figure*}[!h]
  \centering
  \subfloat[Original Data]{\includegraphics[width=0.30\textwidth]{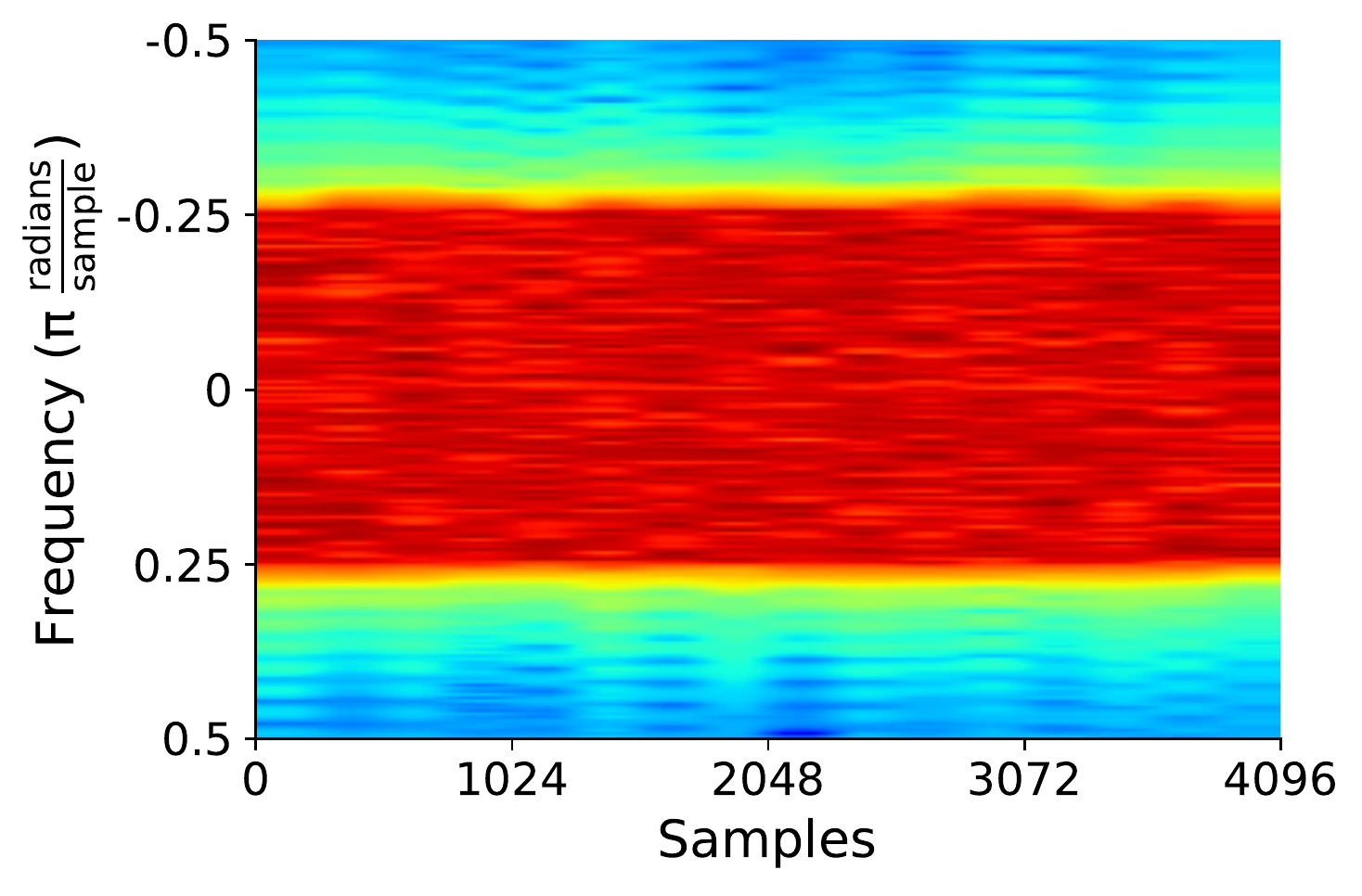}}
  \subfloat[Impaired Data]{\includegraphics[width=0.30\textwidth]{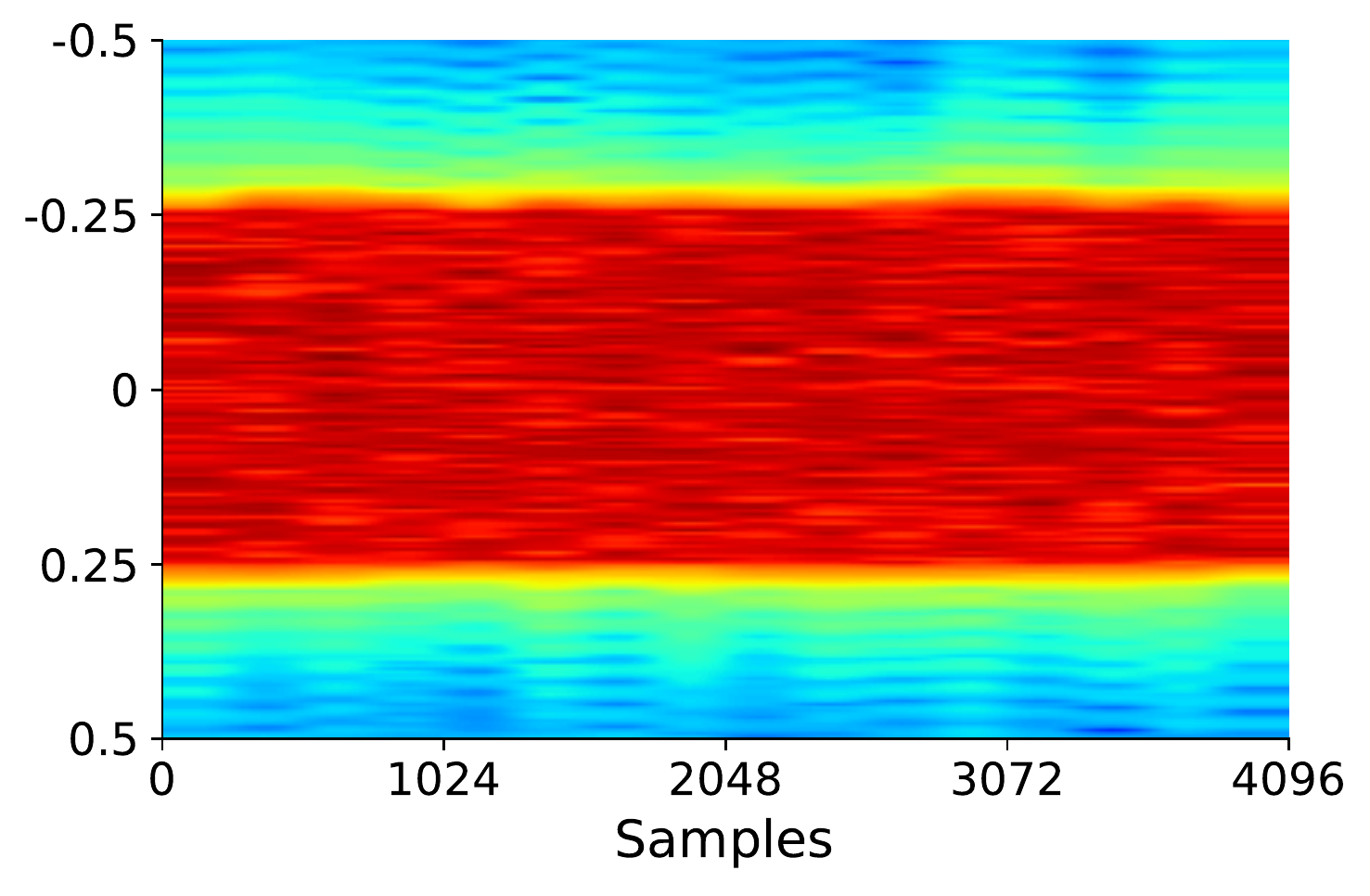}}
  \subfloat[Difference]{\includegraphics[width=0.30\textwidth]{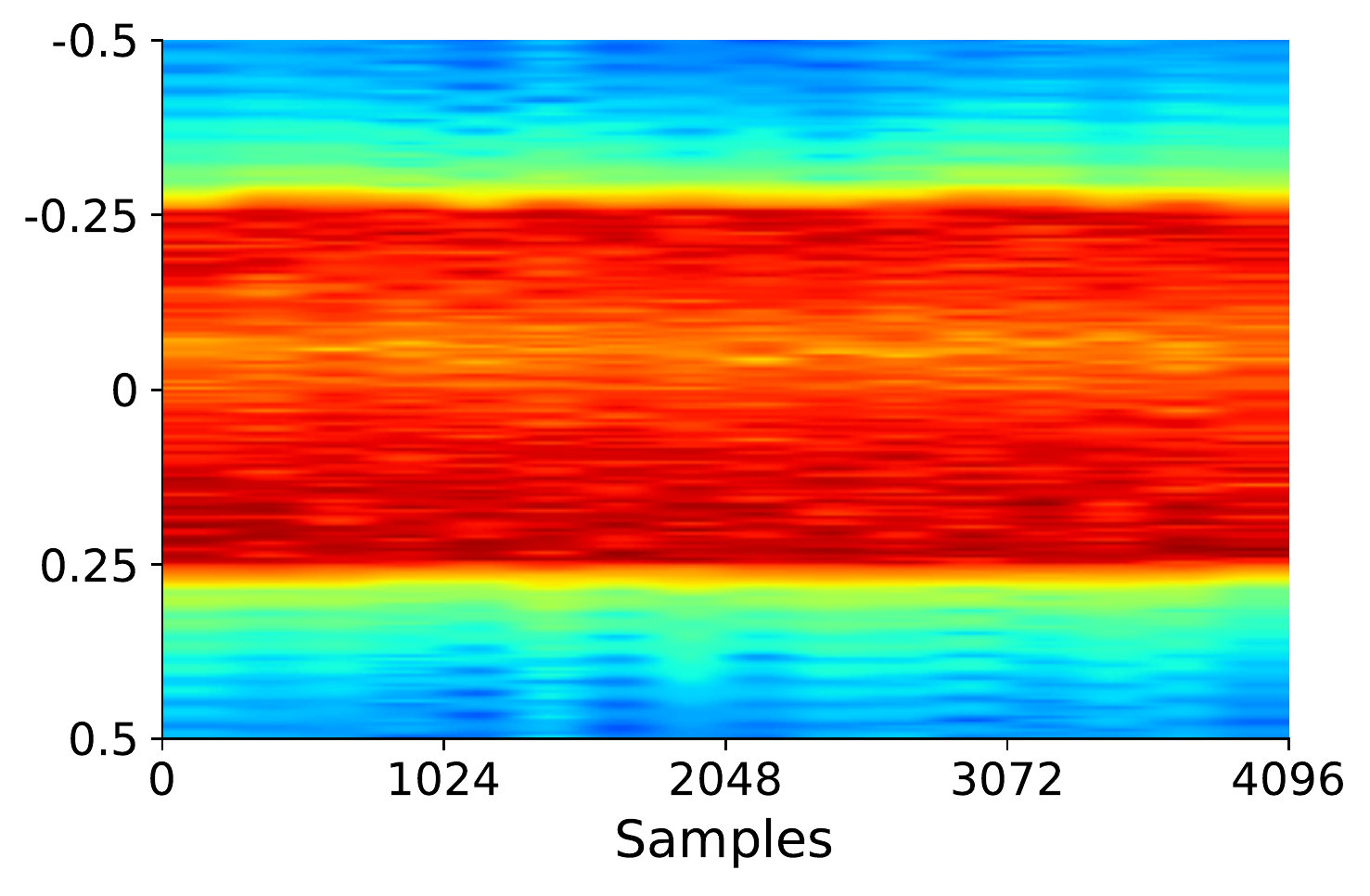}}
  \caption{Rayleigh Fading Impairment}
  \label{fig:rayleigh_fading}
\end{figure*}

\textbf{Rayleigh Fading Channel Impairment.} A Rayleigh fading channel is modeled as a finite impulse response (FIR) filter with Gaussian distributed taps. 
The FIR filter is randomly generated under constraints and then convolved with the input data, simulating the effects of a frequency-selective, time-invariant Rayleigh fading channel. 
\cref{fig:rayleigh_fading} shows an OFDM signal, the OFDM signal with Rayleigh fading applied, and the difference between the original and impaired data to show the frequency-selective nature.

\begin{figure}[!h]
\vskip 0.1in
\begin{center}
\centerline{\includegraphics[width=0.4\textwidth]{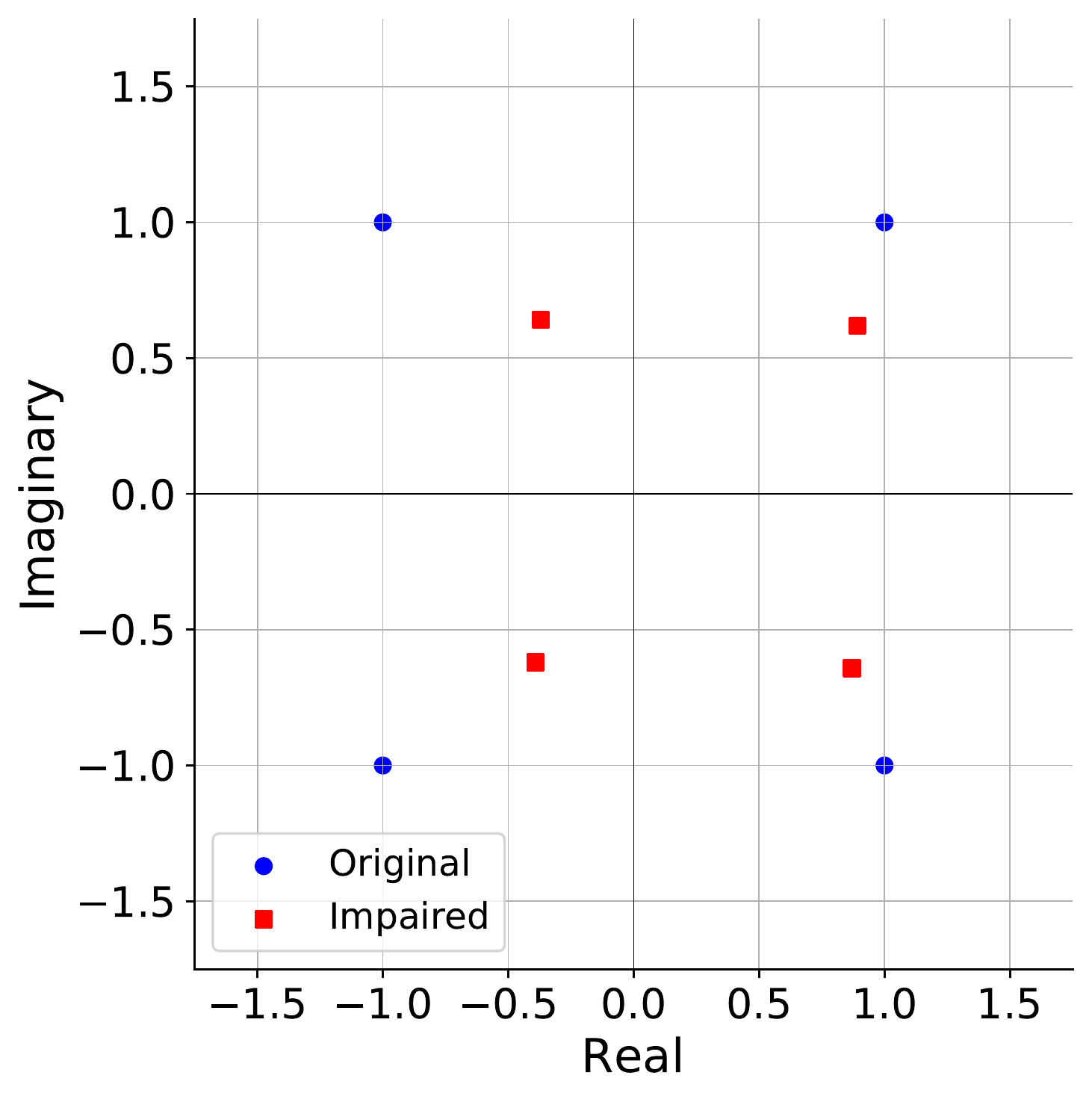}}
\caption{IQ Imbalance Impairment}
\label{fig:iq_imbalance}
\end{center}
\vskip -0.3in
\end{figure}

\clearpage
\textbf{IQ Imbalance Impairment.} IQ imbalance is applied with three randomized parameters: amplitude imbalance, phase imbalance, and DC offset. 
The IQ imbalance impairment is best visualized with a constellation diagram, where the amplitude imbalance grows or shrinks the original constellation points with respect to zero, phase imbalance rotates the original constellations points about the origin, and the DC offset applies a shift along the real axis (\cref{fig:iq_imbalance}).

\begin{figure*}[!h]
  \centering
  \subfloat[Original Data]{\includegraphics[width=0.30\textwidth]{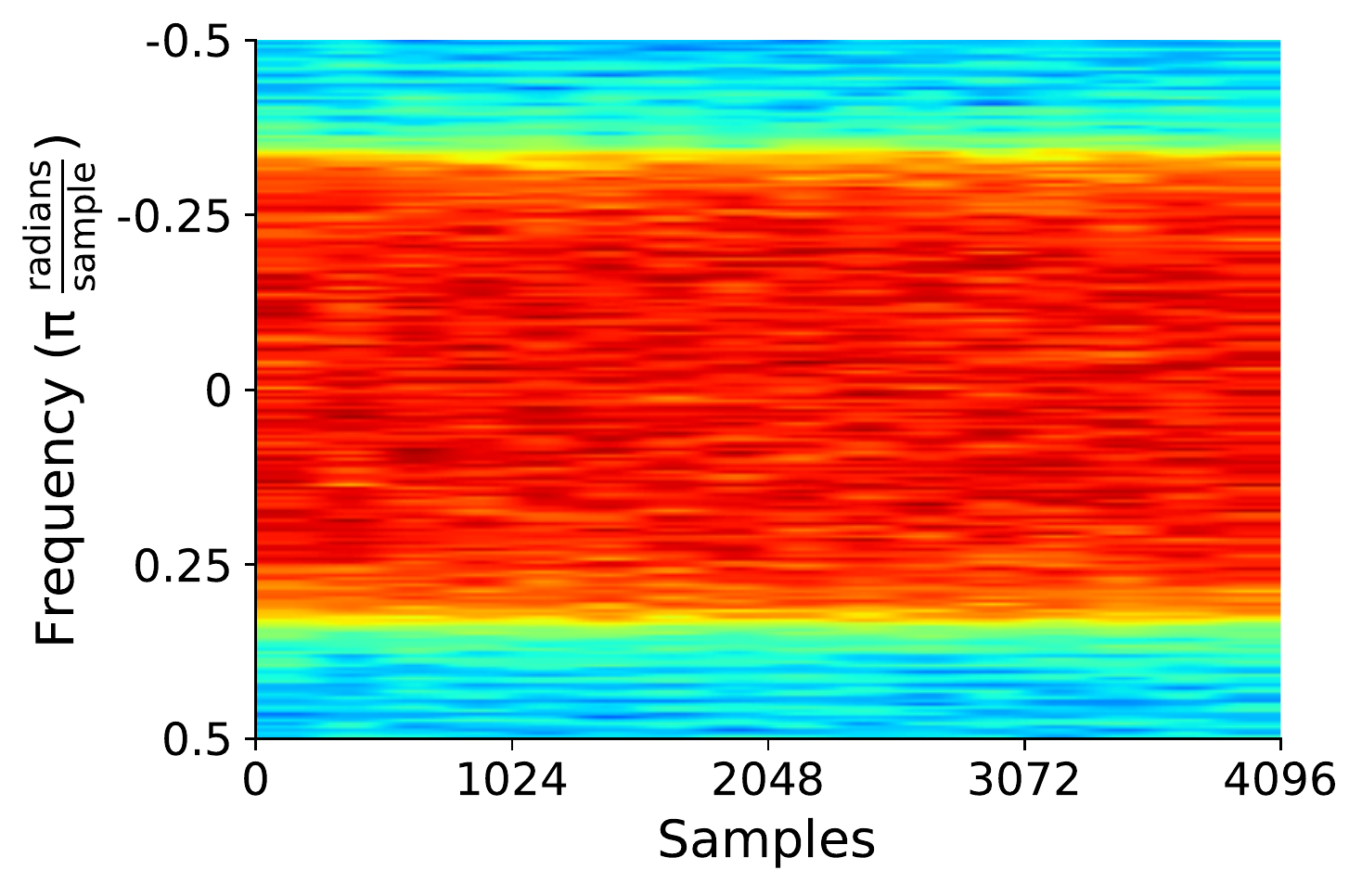}}
  \subfloat[Downsampled Data]{\includegraphics[width=0.30\textwidth]{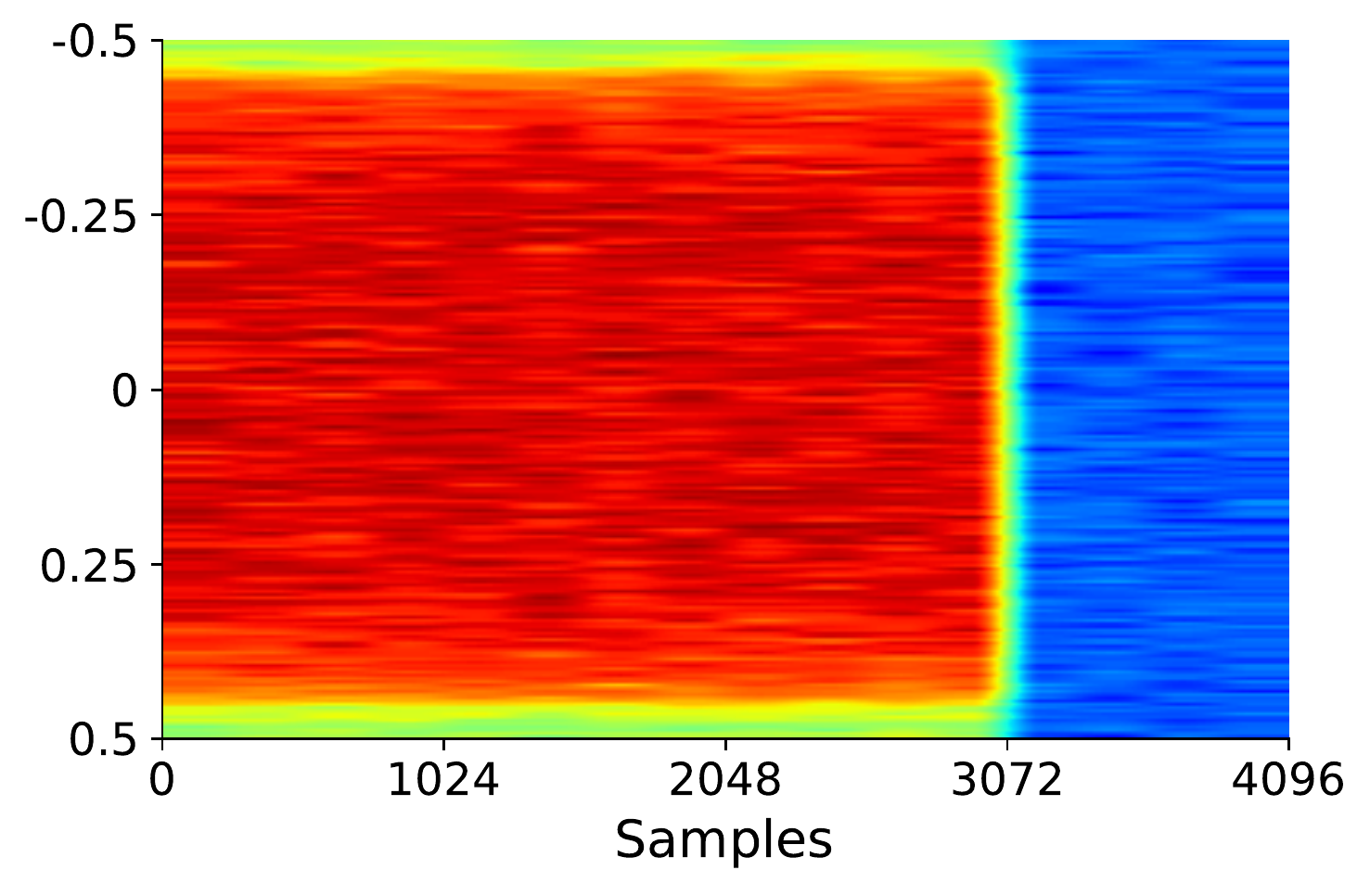}}
  \subfloat[Upsampled Data]{\includegraphics[width=0.30\textwidth]{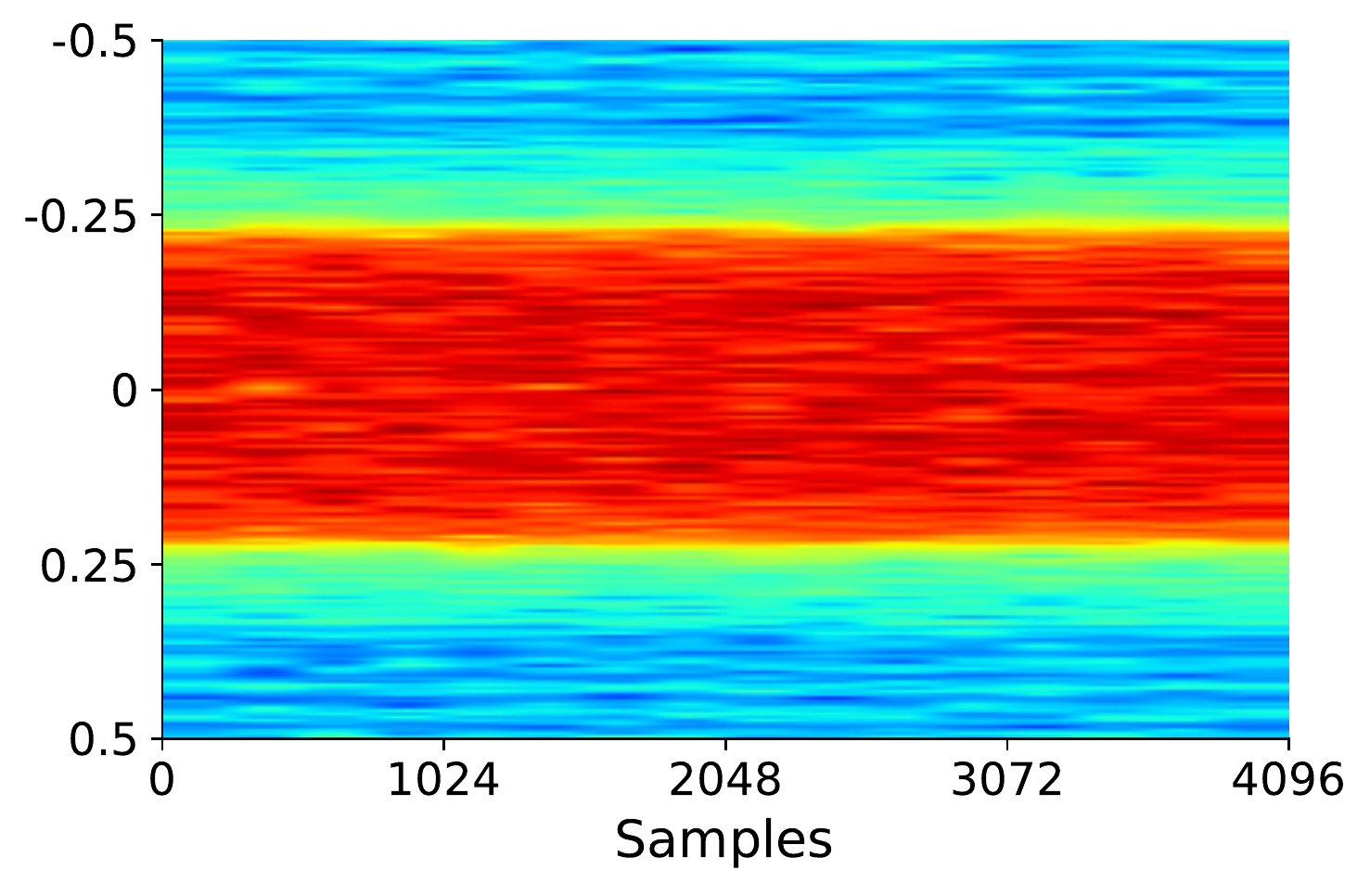}}
  \caption{Random Resample Impairment}
  \label{fig:random_resample}
\end{figure*}

\textbf{Random Resample Impairment.} The random resample impairment randomly selects a new sampling rate and resamples the input data to the new rate. 
\cref{fig:random_resample} shows an example input data capture, the data with a random downsampling applied, and the data with a random upsampling applied. 
Note with downsampling, the example is zero-padded to maintain the original number of IQ samples, and with upsampling, the example is cropped to maintain the original number of IQ samples.

\clearpage
\subsection{Tools Appendix}
\label{sec:appendix_tools}

\subsubsection{Data Augmentations}

In addition to the impairments used in the generation of the Sig53 dataset, 
TorchSig provides support for other signal domain-specific data augmentations we can use during training. 
Since the earlier impairments are used when generating the synthetic dataset, 
it is important to draw augmentations from different distributions in order to avoid emulating an infinite dataset at training time. 
While the infinite dataset approach ultimately provides better results, 
we intentionally separate the data generation impairments from the augmentations when discussing research as applied strictly to the ML techniques involved with the Sig53 static datasets. 
The augmentations below are TorchSig data augmentations discussed in \cref{sec:AugOnline}.

\begin{figure*}[!h]
  \centering
  \subfloat[Original Data]{\includegraphics[width=0.35\textwidth]{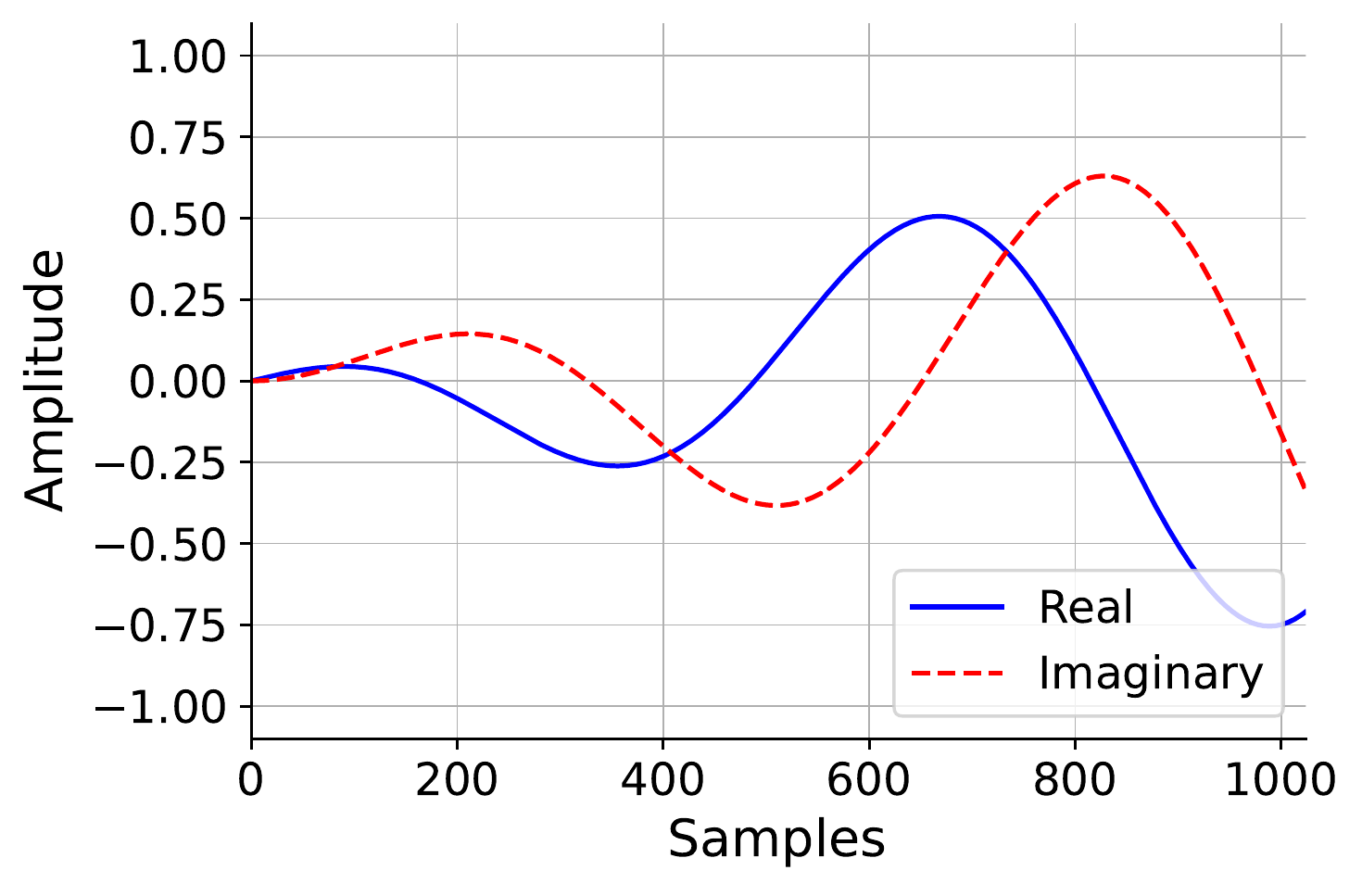}}
  \subfloat[Augmented Data]{\includegraphics[width=0.35\textwidth]{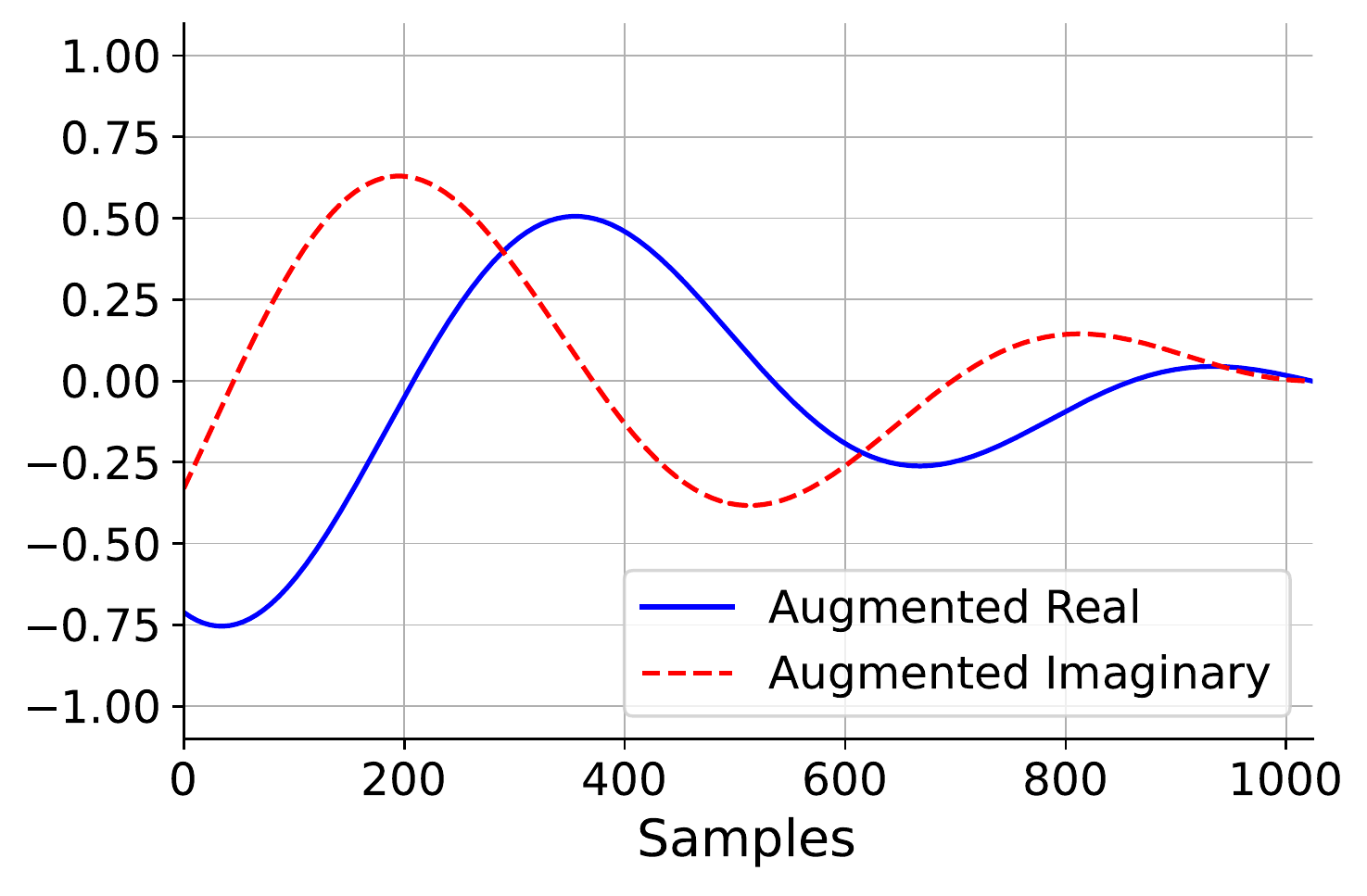}}
  \caption{Time Reversal Augmentation}
  \label{fig:time_reversal}
\end{figure*}

\textbf{Time Reversal.} The time reversal augmentation reverses the order of the IQ samples in the input. 
Since time reversal in the signal domain also results in a spectral inversion, the TorchSig time reversal augmentation additionally has the option to undo the spectral inversion effect if desired (\cref{fig:time_reversal}).

\begin{figure*}[!h]
  \centering
  \subfloat[Original Data]{\includegraphics[width=0.35\textwidth]{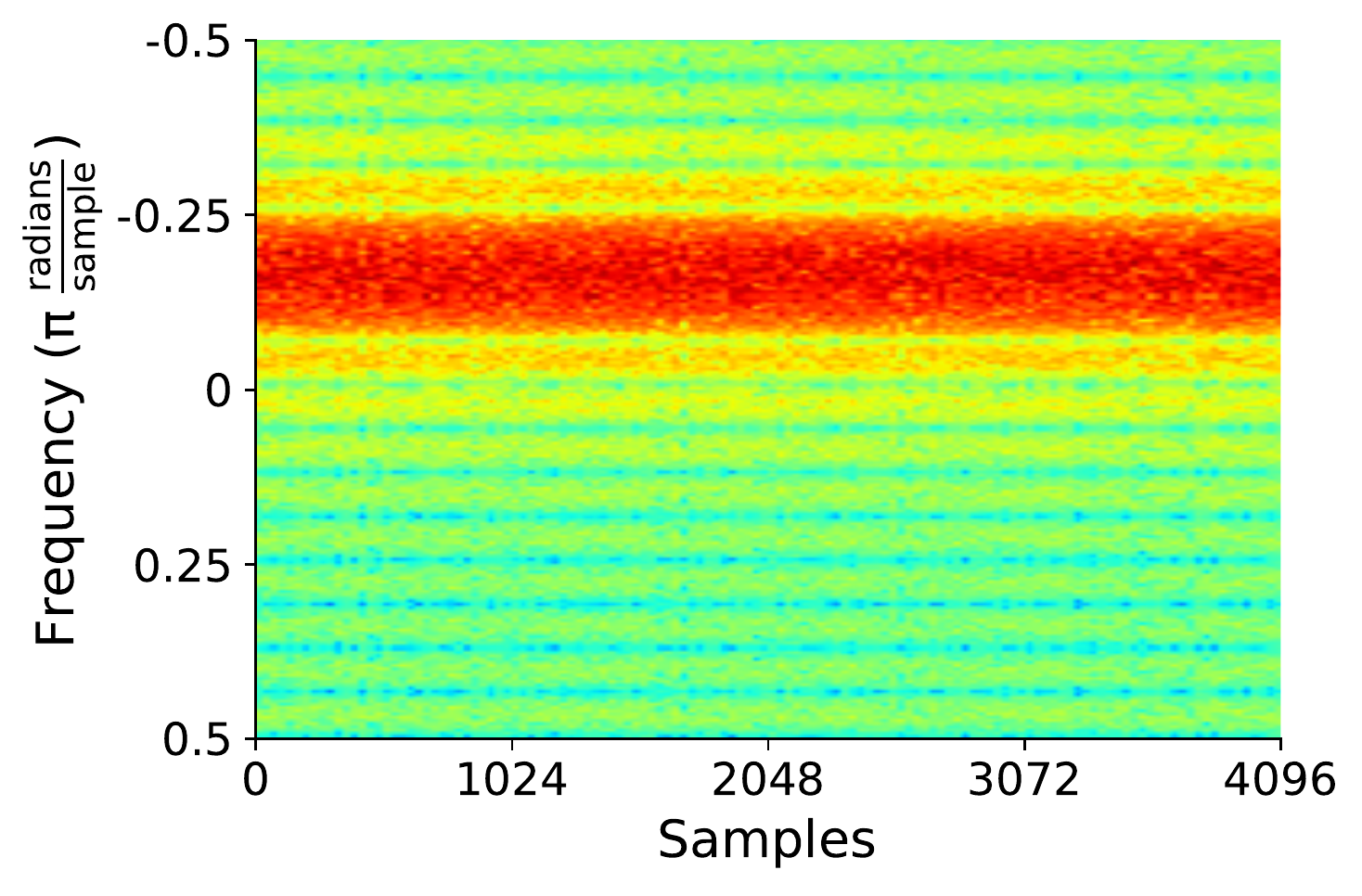}}
  \subfloat[Augmented Data]{\includegraphics[width=0.35\textwidth]{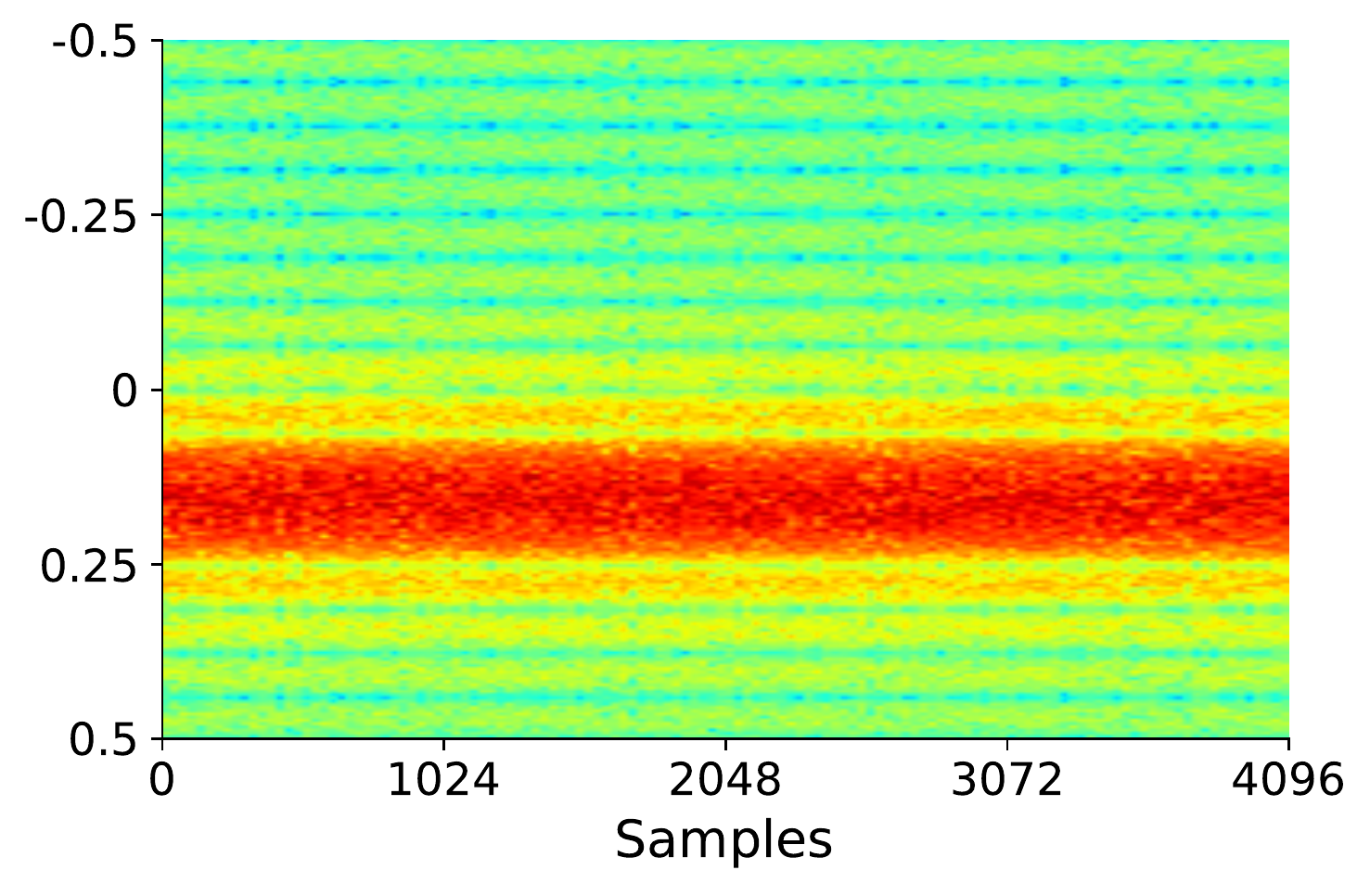}}
  \caption{Spectral Inversion Augmentation}
  \label{fig:spectral_inversion}
\end{figure*}

\textbf{Spectral Inversion.} The spectral inversion augmentation inverts the frequency components of the input data by negating the imaginary components of the input (\cref{fig:spectral_inversion}).

\begin{figure*}[!h]
  \centering
  \subfloat[Original Data]{\includegraphics[width=0.35\textwidth]{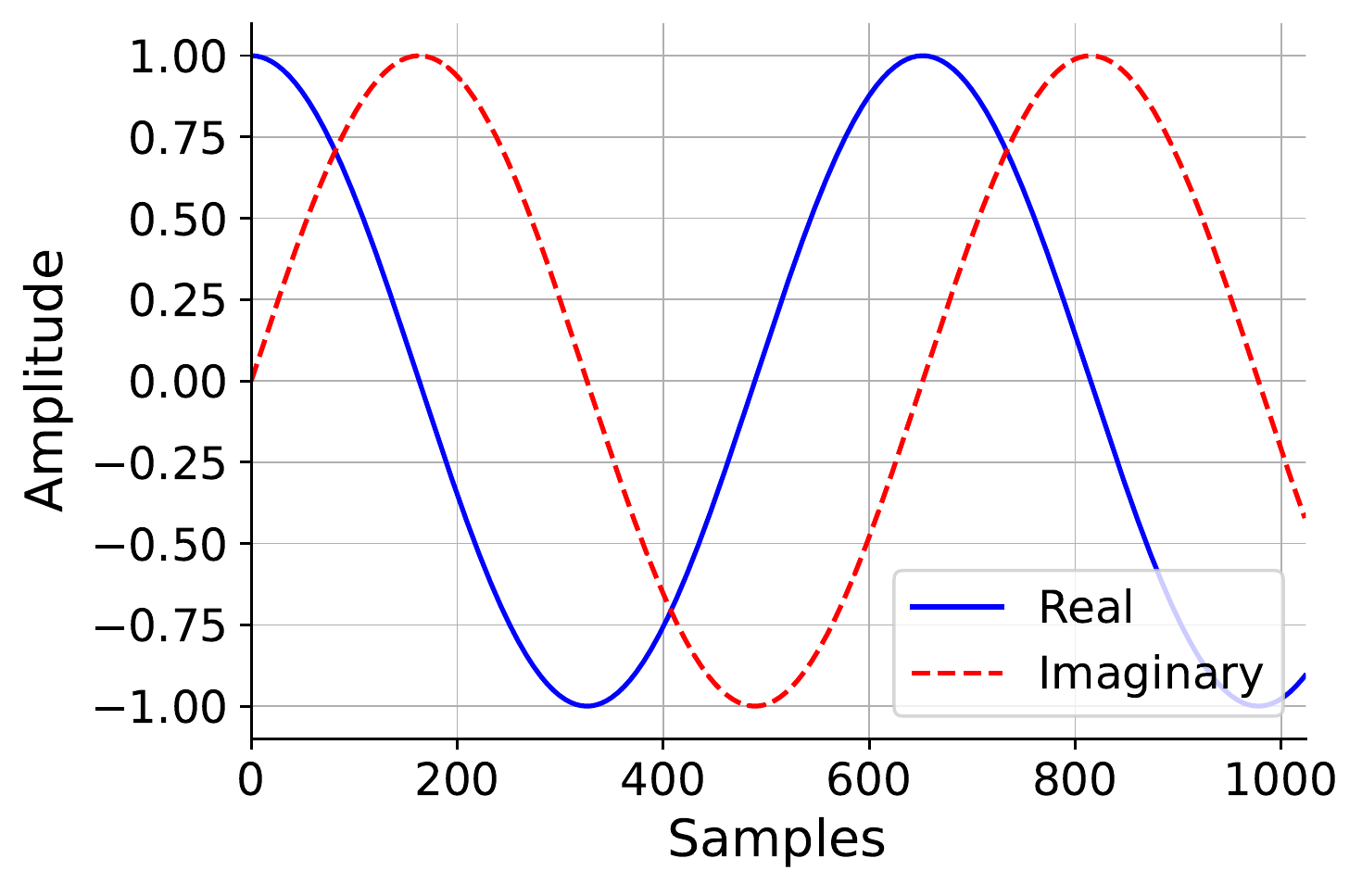}}
  \subfloat[Augmented Data]{\includegraphics[width=0.35\textwidth]{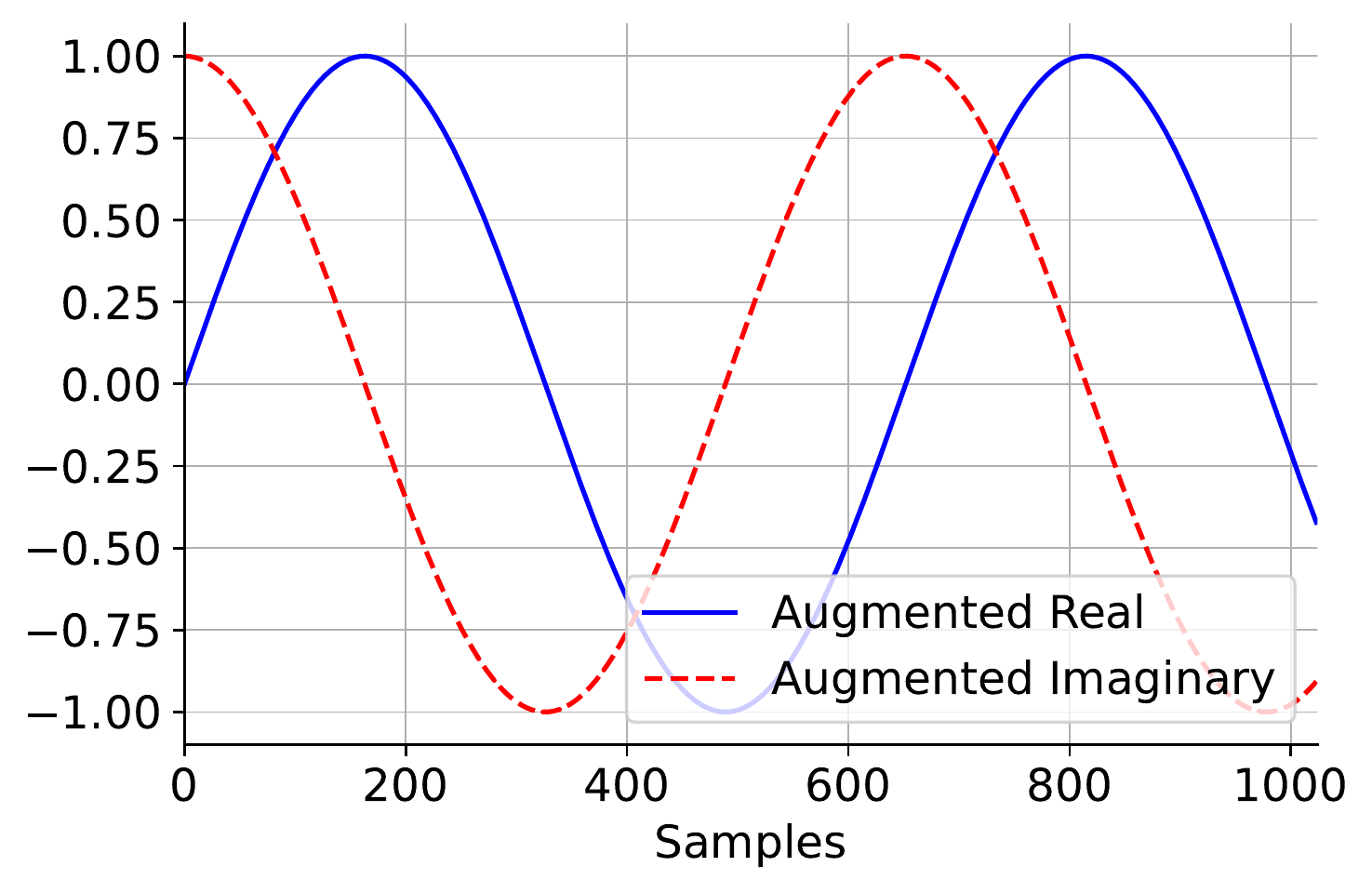}}
  \caption{Channel Swap Augmentation}
  \label{fig:channel_swap}
\end{figure*}

\clearpage
\textbf{Channel Swap.} The channel swap augmentation switches the real and imaginary componets of the input complex data. 
In the signal domain, this has the same effect as a spectral inversion followed by a static $\pi$/2 phase shift (\cref{fig:channel_swap}).

\begin{figure*}[!h]
  \centering
  \subfloat[Original Data]{\includegraphics[width=0.35\textwidth]{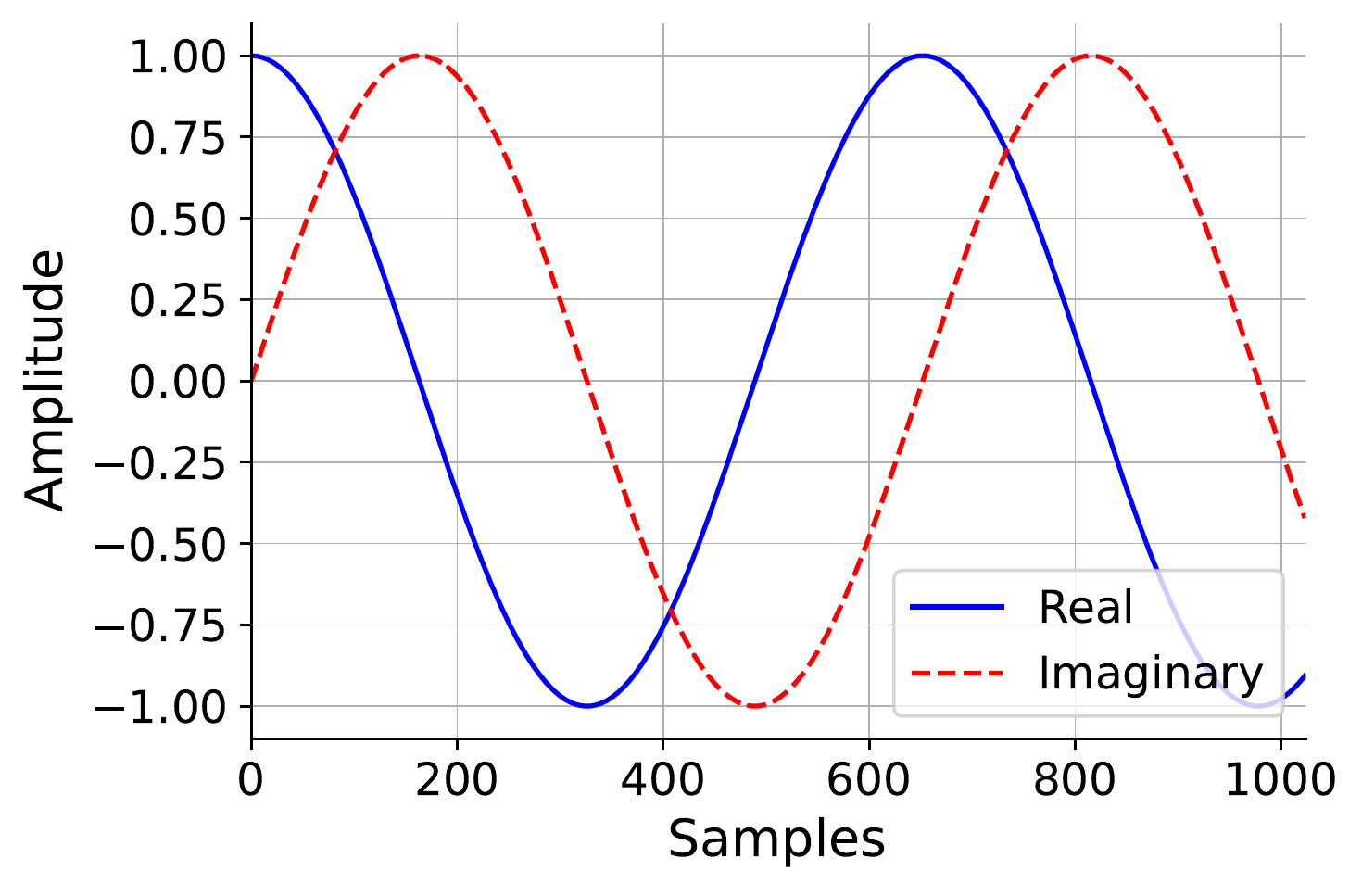}}
  \subfloat[Augmented Data]{\includegraphics[width=0.35\textwidth]{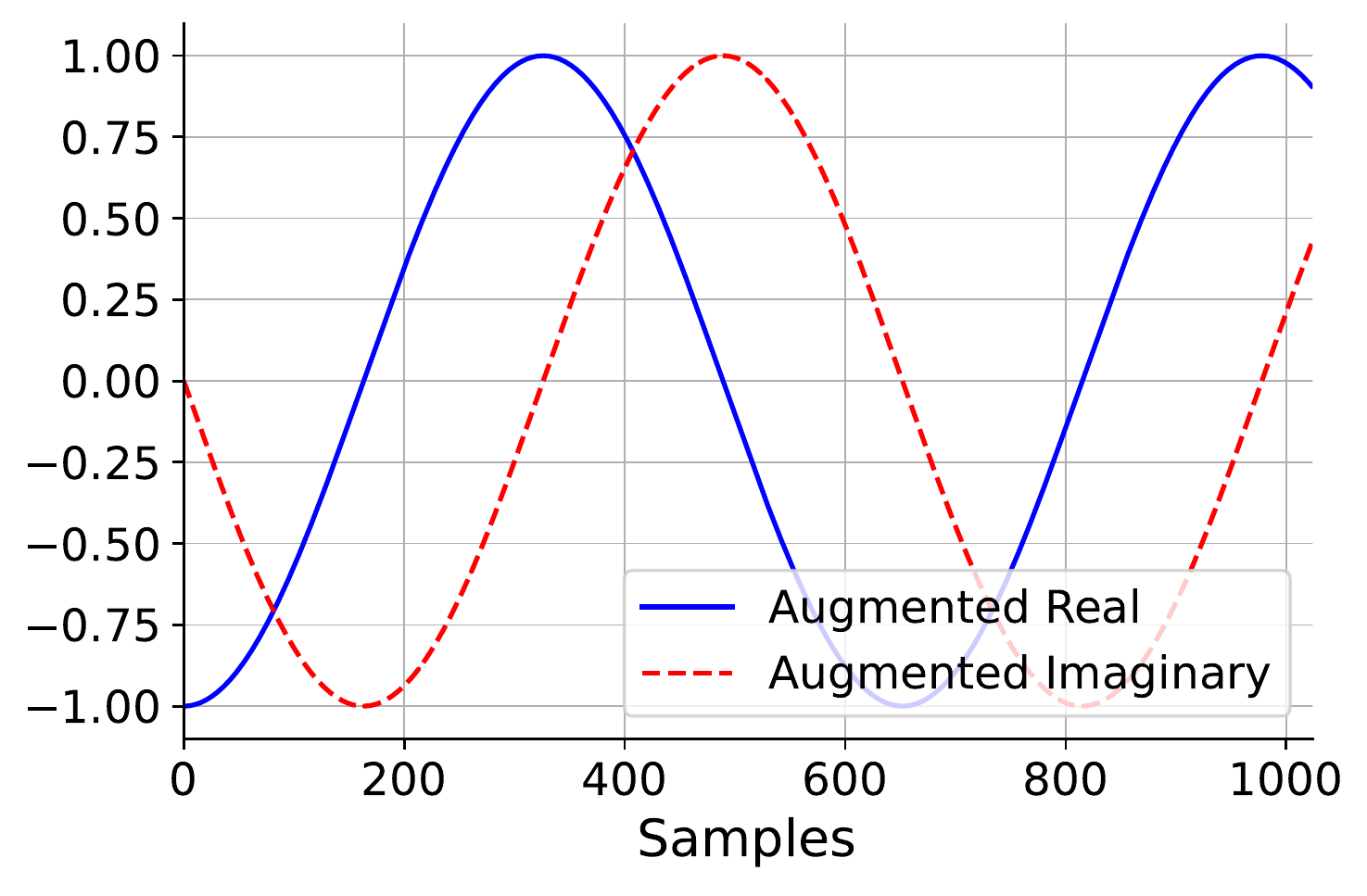}}
  \caption{Amplitude Reversal Augmentation}
  \label{fig:amp_reversal}
\end{figure*}

\textbf{Amplitude Reversal.} Amplitude reversal augments the input data by simply multiplying by -1 (\cref{fig:amp_reversal}).

\begin{figure*}[!h]
  \centering
  \includegraphics[width=0.4\textwidth]{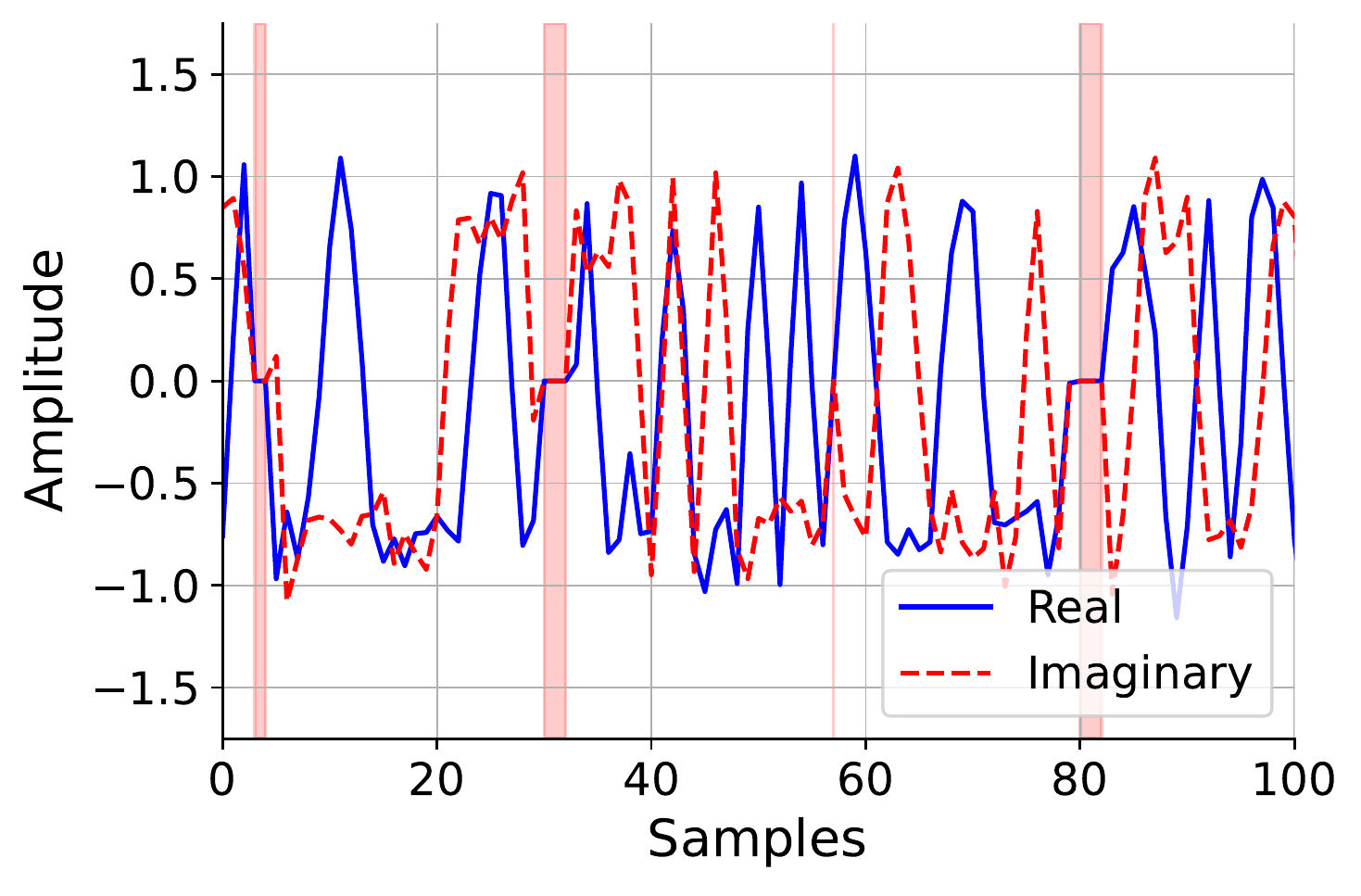}
  \caption{Drop Samples Augmentation}
  \label{fig:drop_samples}
\end{figure*}

\textbf{Drop Samples.} The drop samples augmentation randomly drops IQ samples from the input data using randomized values for the drop rate, the size of each dropped region, and the fill methods for how to replace the regions with dropped samples. 
The fill methods can be statically or randomly set to choose the following methods: front fill, back fill, mean, or zero; 
where front fill replaces each drop regions' samples with the last previous valid value, 
back fill replaces each drop regions' samples with the next valid value, 
mean replaces each drop regions' samples with the mean value of the full data example, 
and zero replaces each drop regions' samples with zeros. 
This transform is loosely based on TSAug's DropOut transform \cite{wen2019tsaug} (\cref{fig:drop_samples}).

\begin{figure*}[!h]
  \centering
  \subfloat[Original Data]{\includegraphics[width=0.35\textwidth]{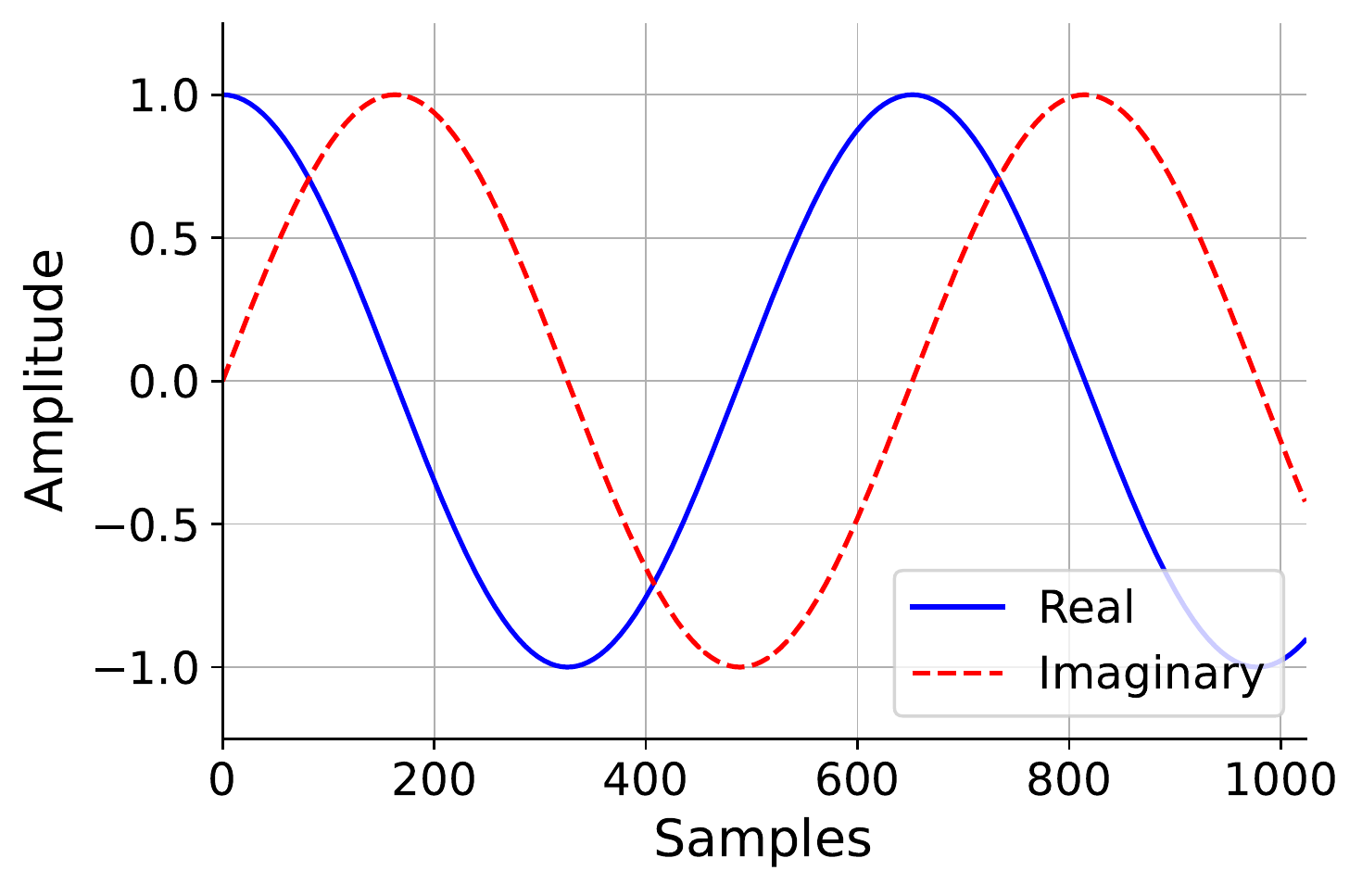}}
  \subfloat[Augmented Data]{\includegraphics[width=0.35\textwidth]{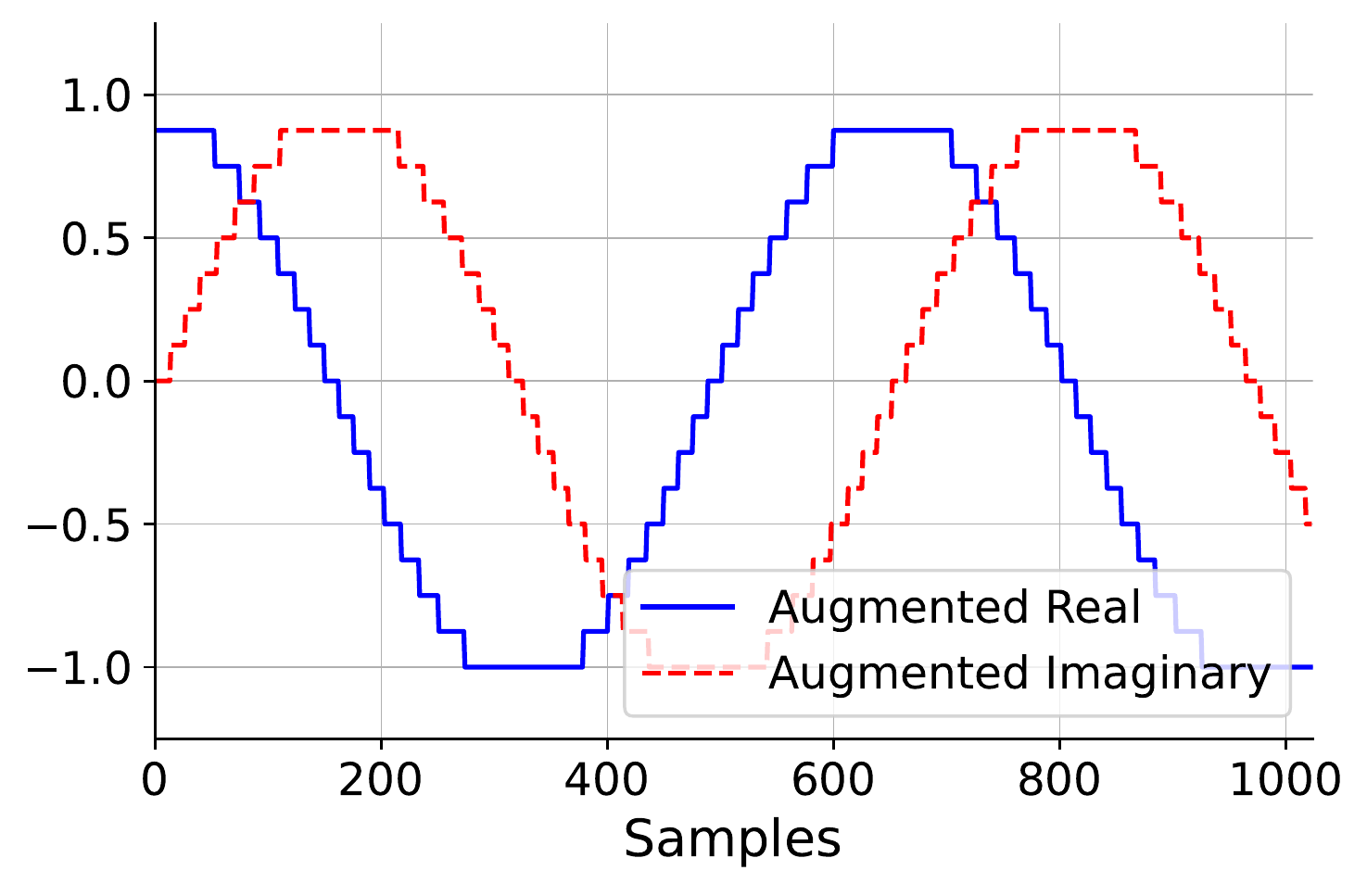}}
  \caption{Quantize Augmentation}
  \label{fig:quantize}
\end{figure*}

\textbf{Quantize.} Data can be quantized to randomly selected numbers of levels with the quantize data transform, loosely emulating the bit-depth in an analog-to-digital converter (ADC) seen in digital RF systems. 
The quantization transform also allows for various rounding types between: 
flooring the observed values to the next-lowest valid quantized value, 
setting every value in a region to the middle value of the region, 
or rounding each value to the next largest valid quantized value (\cref{fig:quantize}).

\begin{figure*}[!h]
  \centering
  \subfloat[Time Series Visualization]{\includegraphics[width=0.35\textwidth]{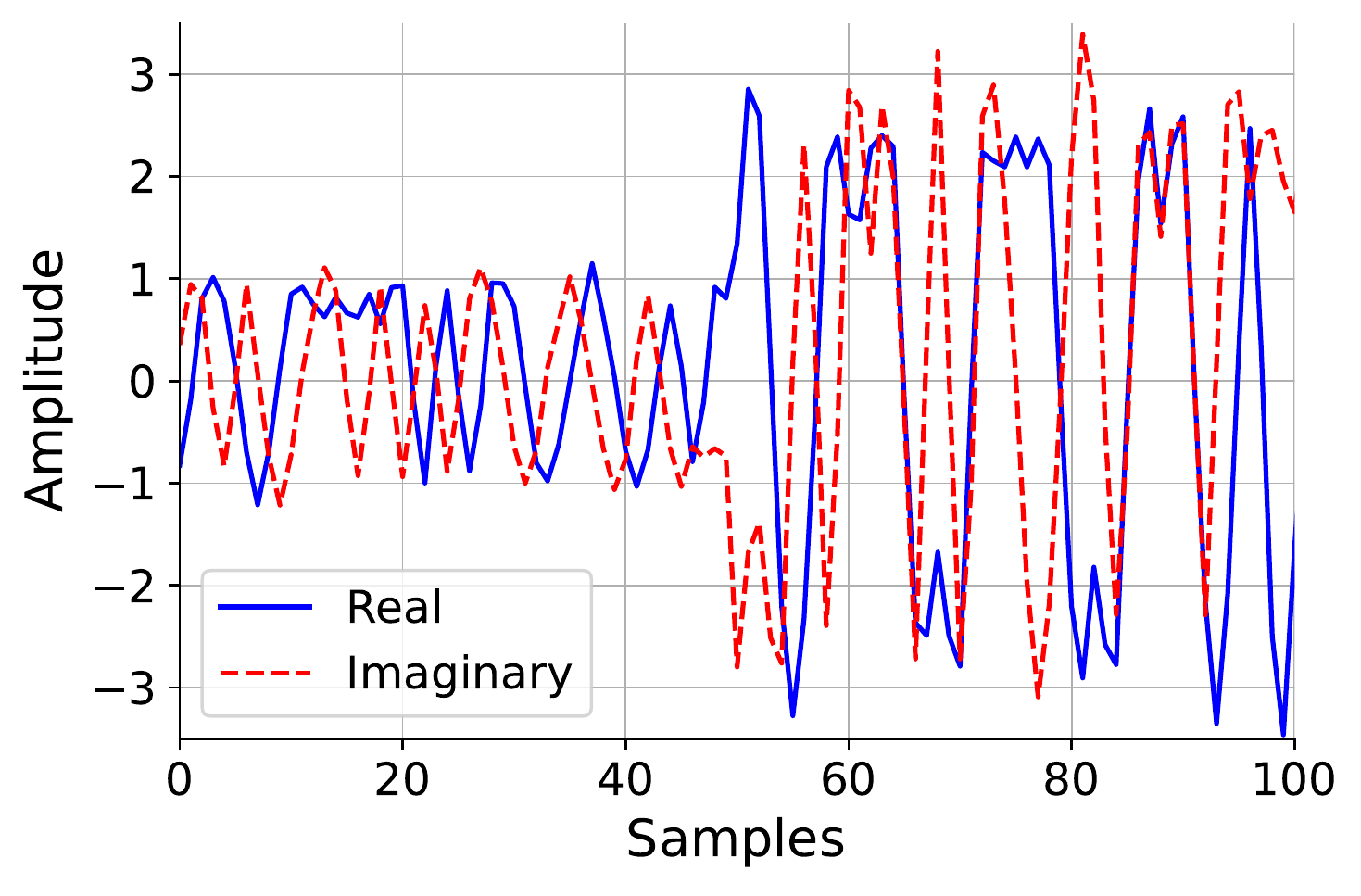}}
  \subfloat[Spectrogram Visualization]{\includegraphics[width=0.35\textwidth]{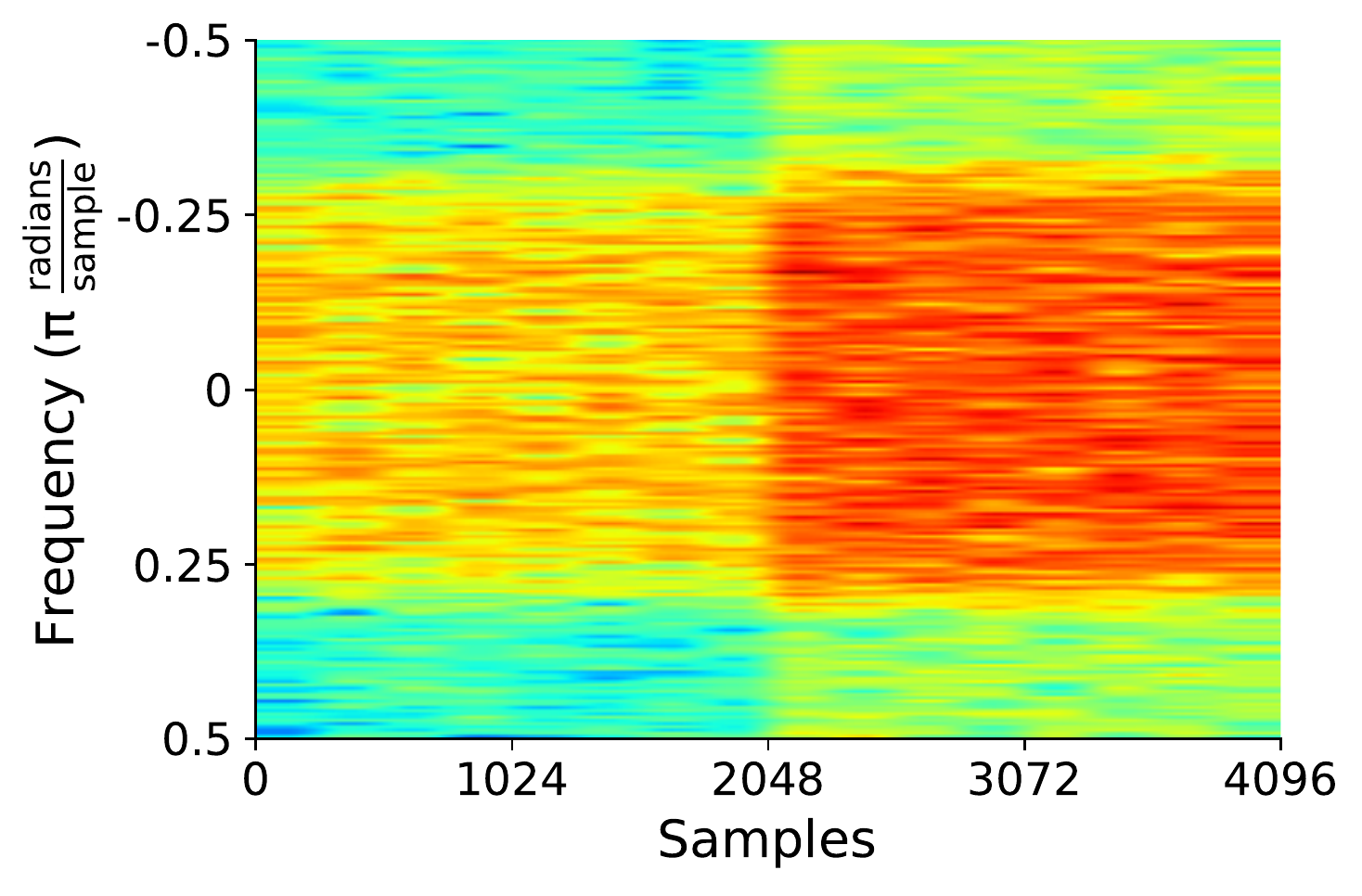}}
  \caption{Magnitude Rescaling Augmentation}
  \label{fig:mag_rescale}
\end{figure*}

\textbf{Magnitude Rescale.} The magnitude rescaling transform randomly selects a starting point of the input example to rescale the magnitude of the data by multiplying by a random constant. 
This behavior emulates an RF front end gain adjustment (\cref{fig:mag_rescale}).

\begin{figure*}[!h]
  \centering
  \subfloat[Original Data]{\includegraphics[width=0.35\textwidth]{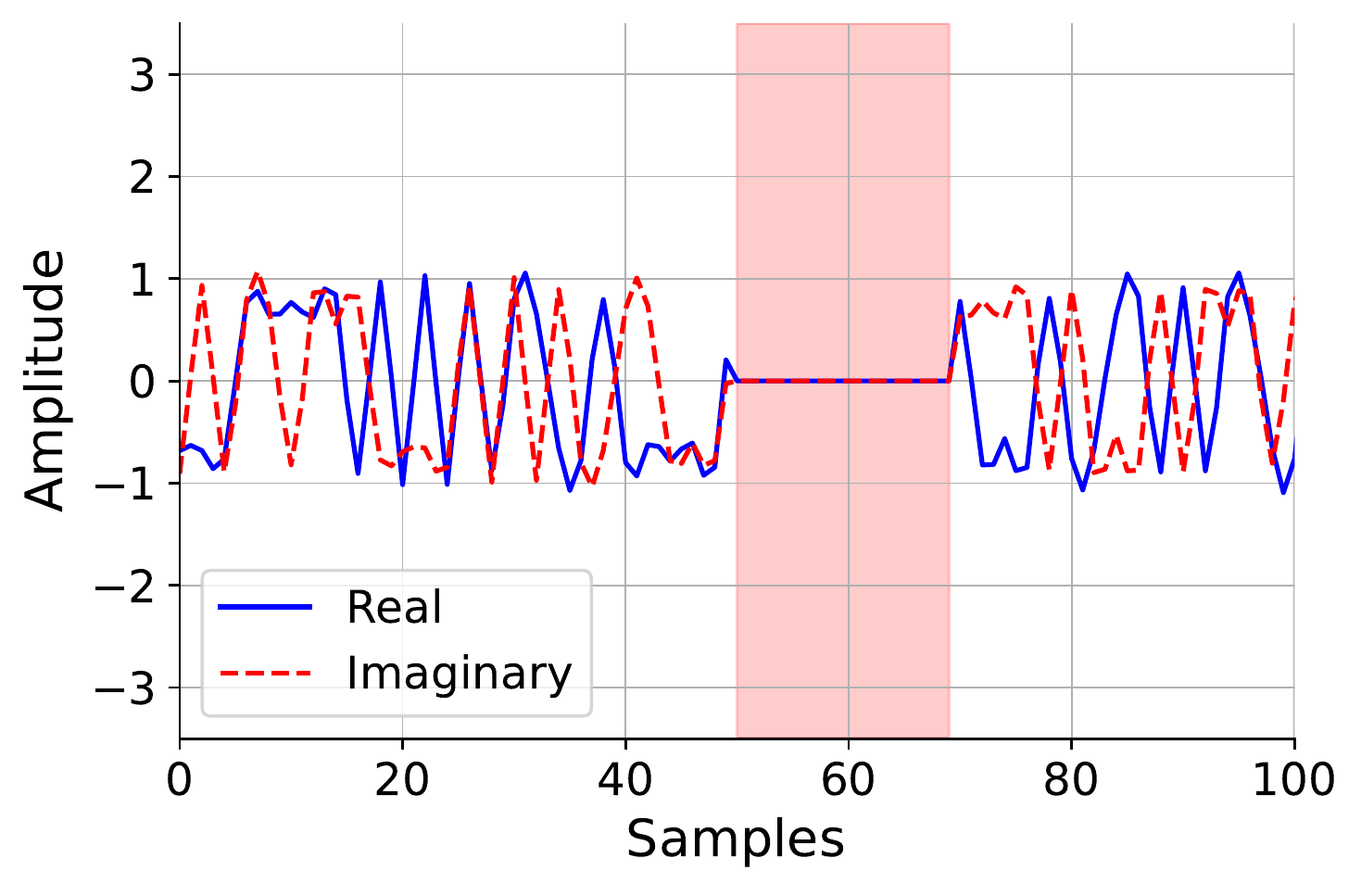}}
  \subfloat[Augmented Data]{\includegraphics[width=0.35\textwidth]{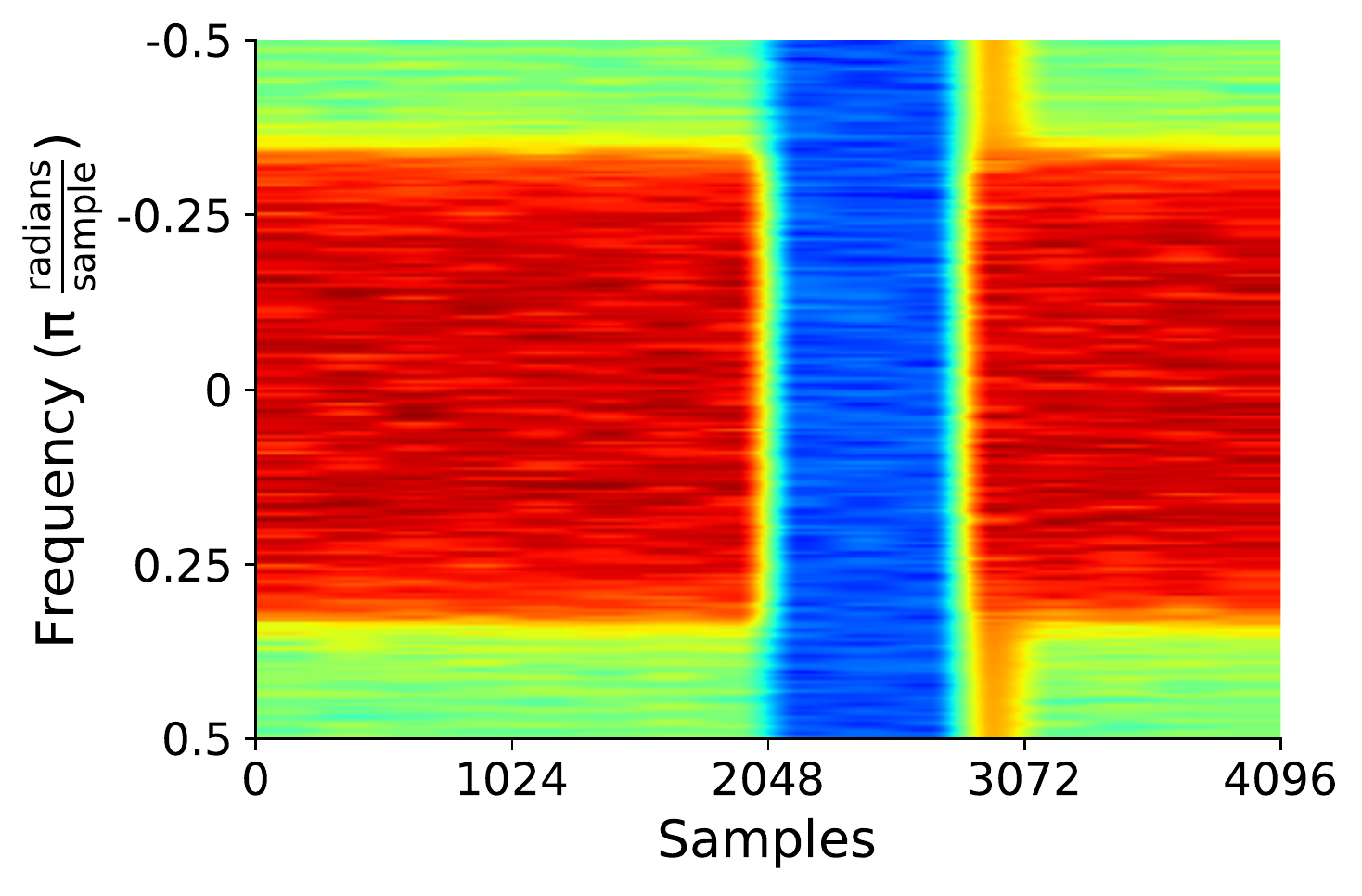}}
  \caption{CutOut Augmentation}
  \label{fig:cutout}
\end{figure*}

\clearpage
\textbf{CutOut.} The CutOut transform is a modified version of the computer vision domain's CutOut as seen in \cite{devries2017improved}.
Our version of CutOut inputs randomized cut durations and cut types to select how large a region in time should be cut out. 
The cut out region is then filled with either zeros, ones, low-SNR noise, average-SNR noise, or high-SNR noise (\cref{fig:cutout}).

\begin{figure*}[!h]
  \centering
  \subfloat[Original Data]{\includegraphics[width=0.35\textwidth]{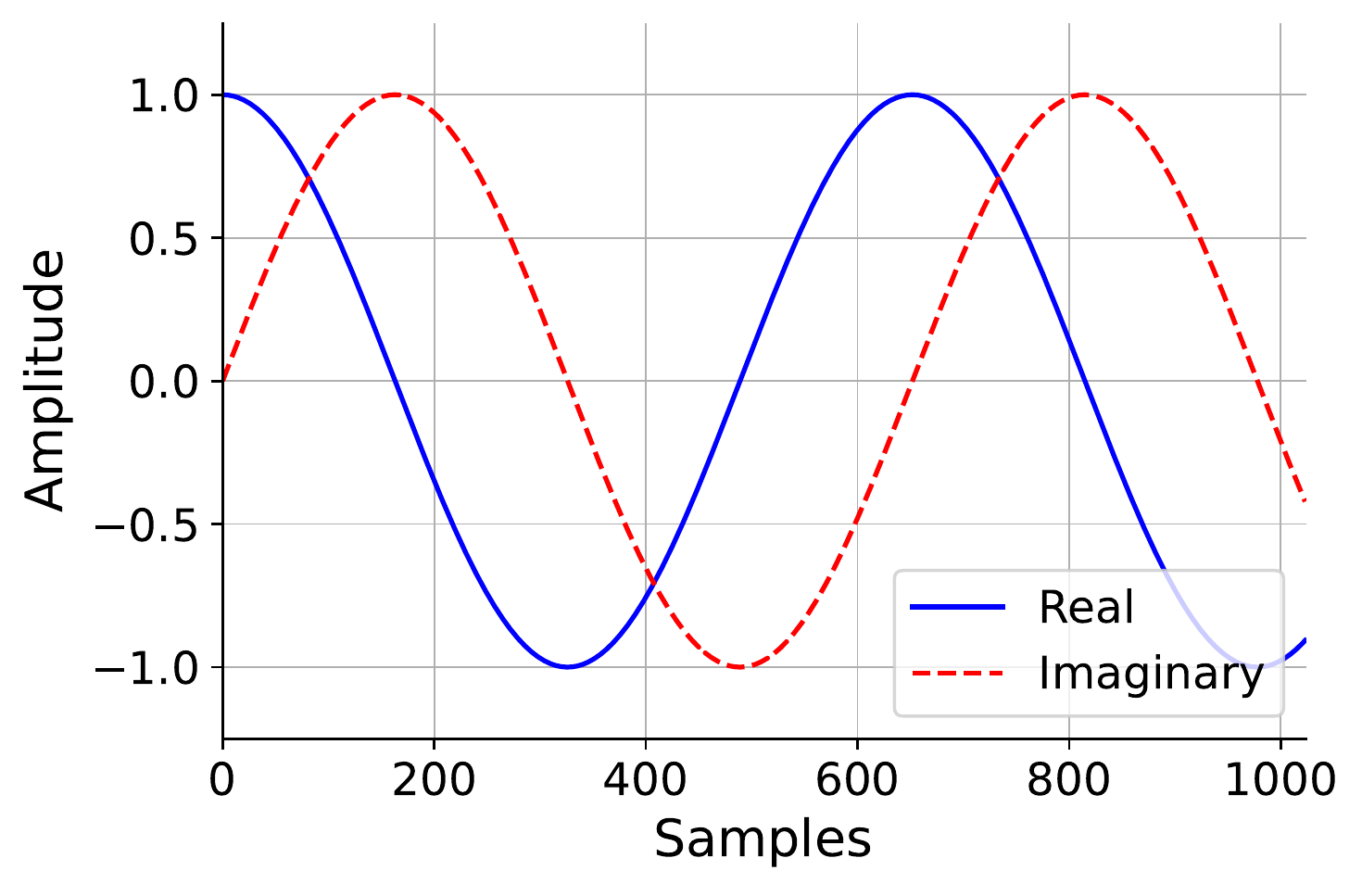}}
  \subfloat[Augmented Data]{\includegraphics[width=0.35\textwidth]{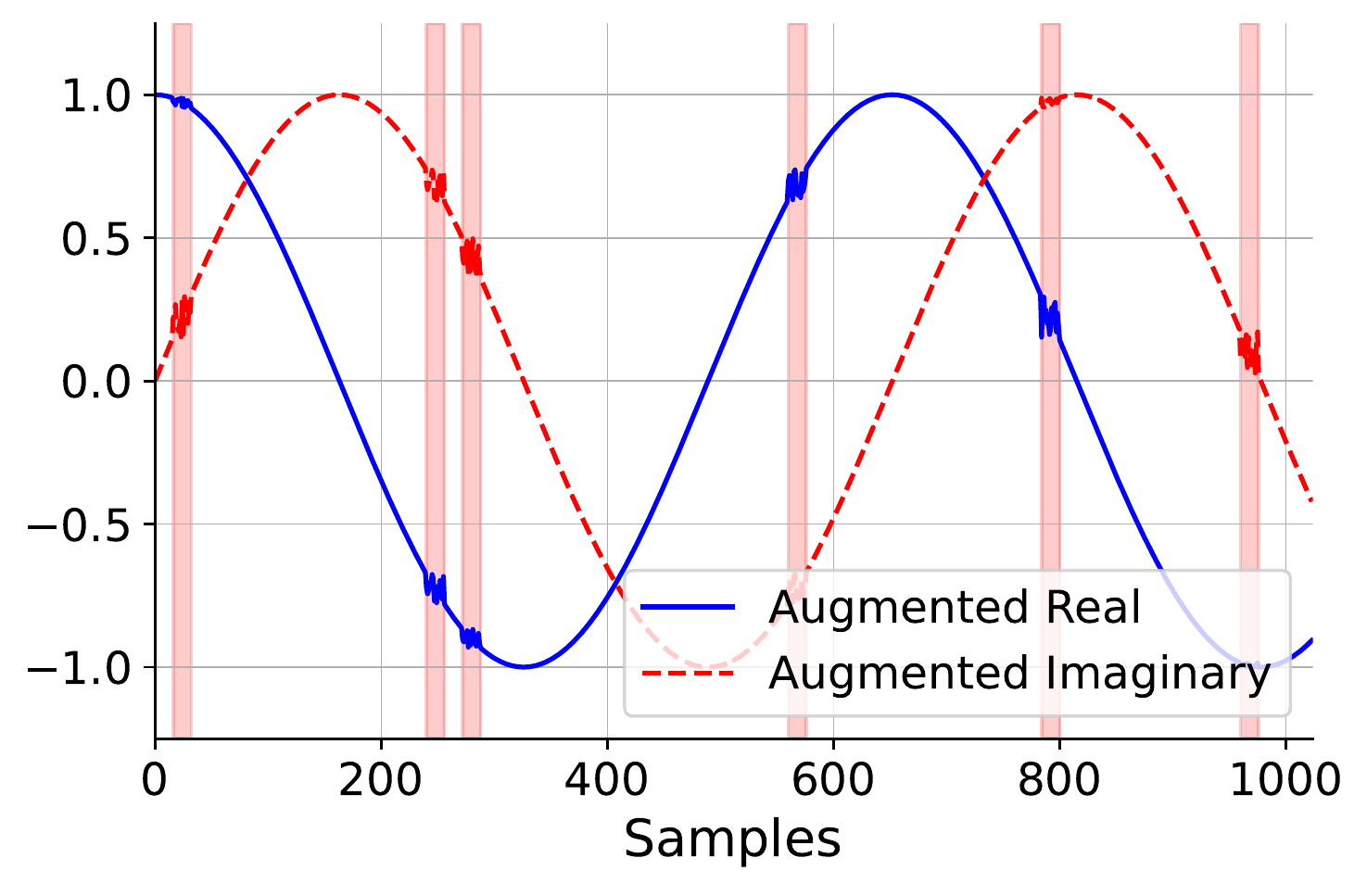}}
  \caption{PatchShuffle Augmentation}
  \label{fig:patch_shuffle}
\end{figure*}

\textbf{PatchShuffle.} The PatchShuffle transform is a modified version of the computer vision domain's PatchShuffle as seen in \cite{kang2017patchshuffle}. 
Our version of PatchShuffle operates solely on in the time domain, randomly shuffling multiple local regions of IQ samples, using a randomized patch size input distribution and a randomized shuffle ratio to discern how many of the patches should undergo random local shuffling (\cref{fig:patch_shuffle}).

\clearpage
\subsubsection{Additional TorchSig Transforms}

The TorchSig toolkit also contains a handful of additional transforms that are not used in the scope of this paper's experimentation. 
These additional transforms are briefly described below, and additional research on the effectiveness of these transforms in addition to or in lieu of the above transformations is an open research question.

\begin{figure*}[!h]
  \centering
  \subfloat[Original Data]{\includegraphics[width=0.35\textwidth]{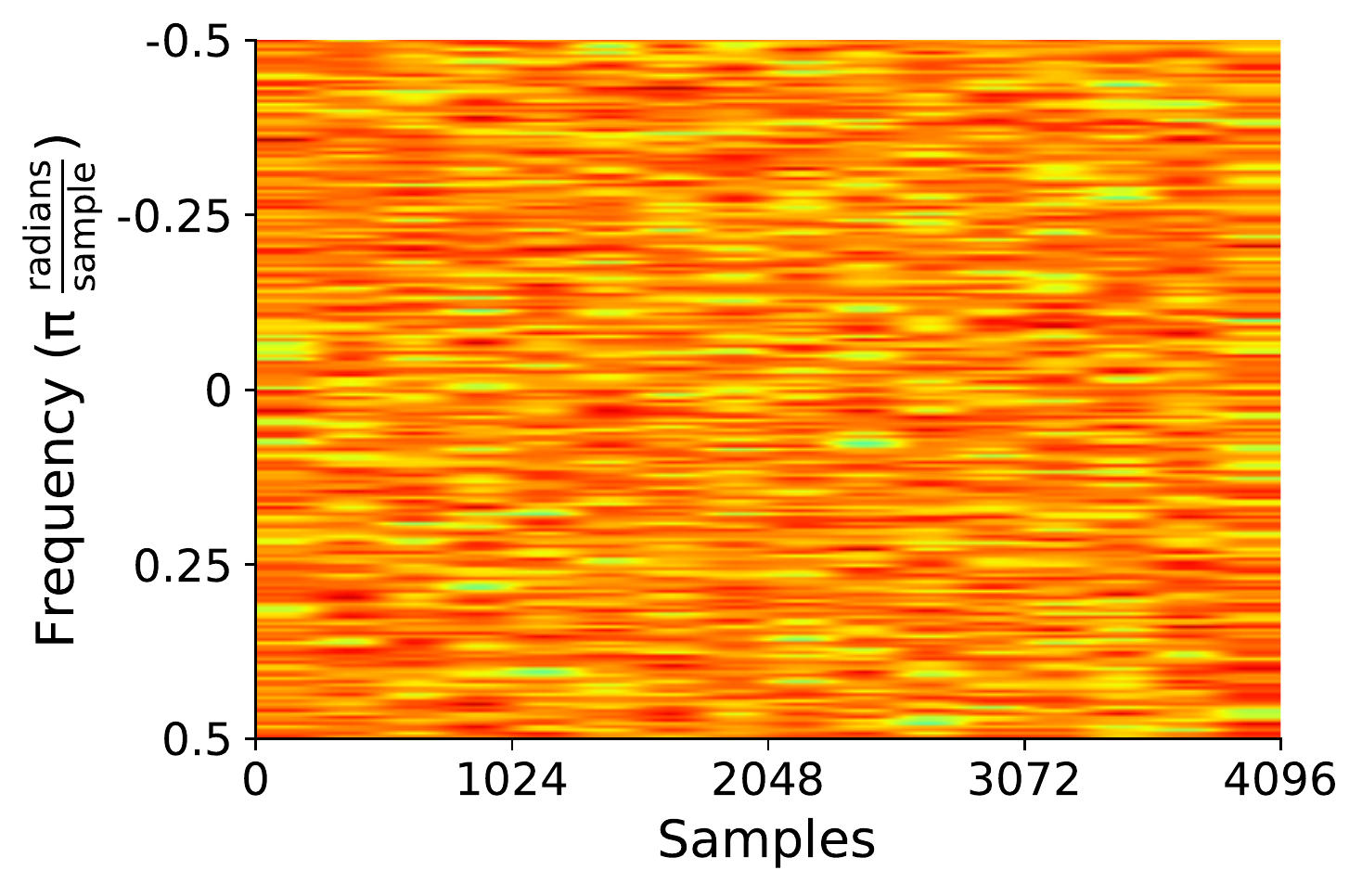}}
  \subfloat[Augmented Data]{\includegraphics[width=0.35\textwidth]{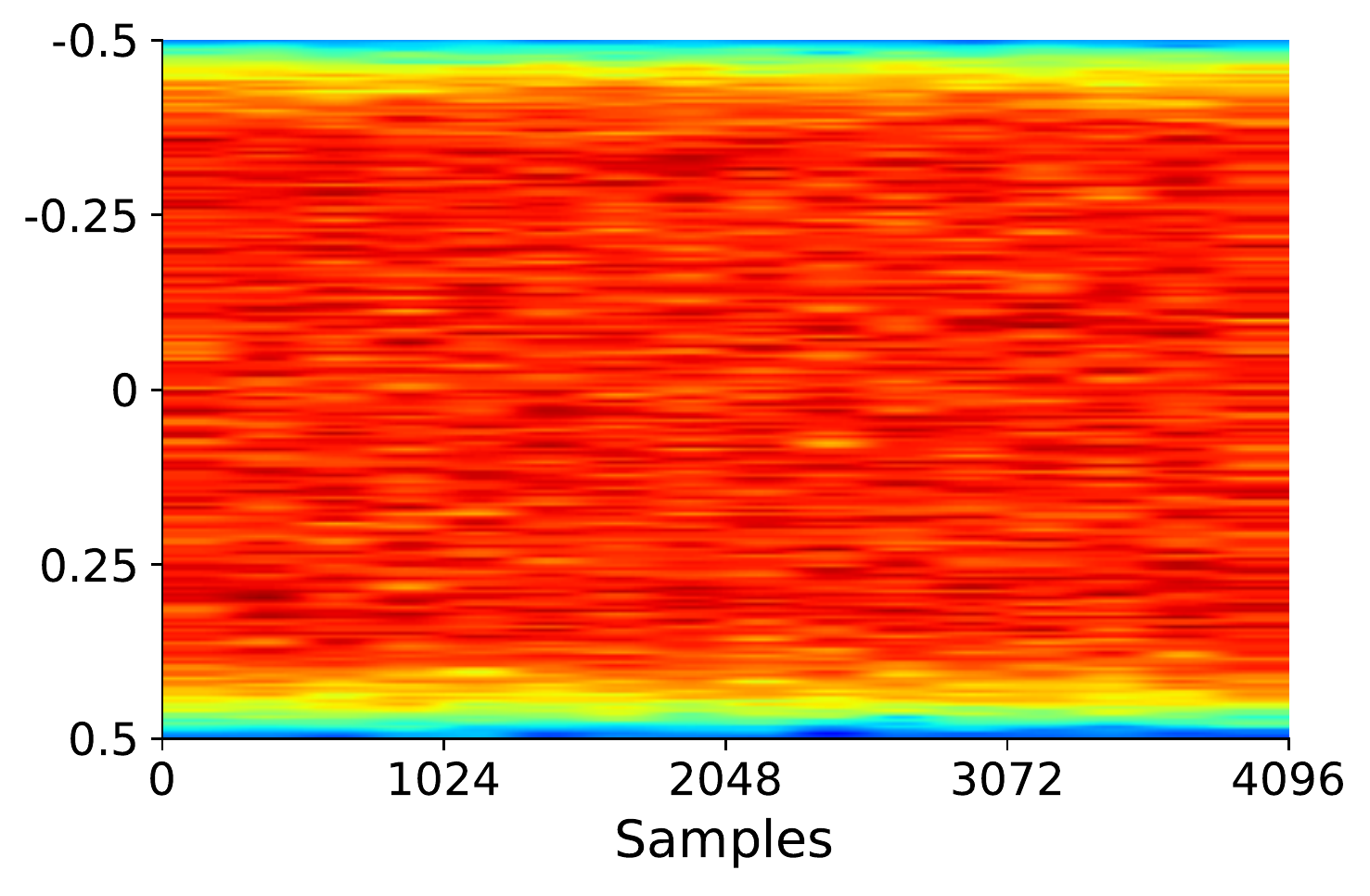}}
  \caption{Roll-Off Augmentation}
  \label{fig:rolloff}
\end{figure*}

\textbf{Signal Roll-Off.} The signal roll-off transform applies a lower, upper, or both-sided band-edge RF roll-off effect, simulating front end filtering (\cref{fig:rolloff}).

\begin{figure*}[!h]
  \centering
  \subfloat[Original Data]{\includegraphics[width=0.35\textwidth]{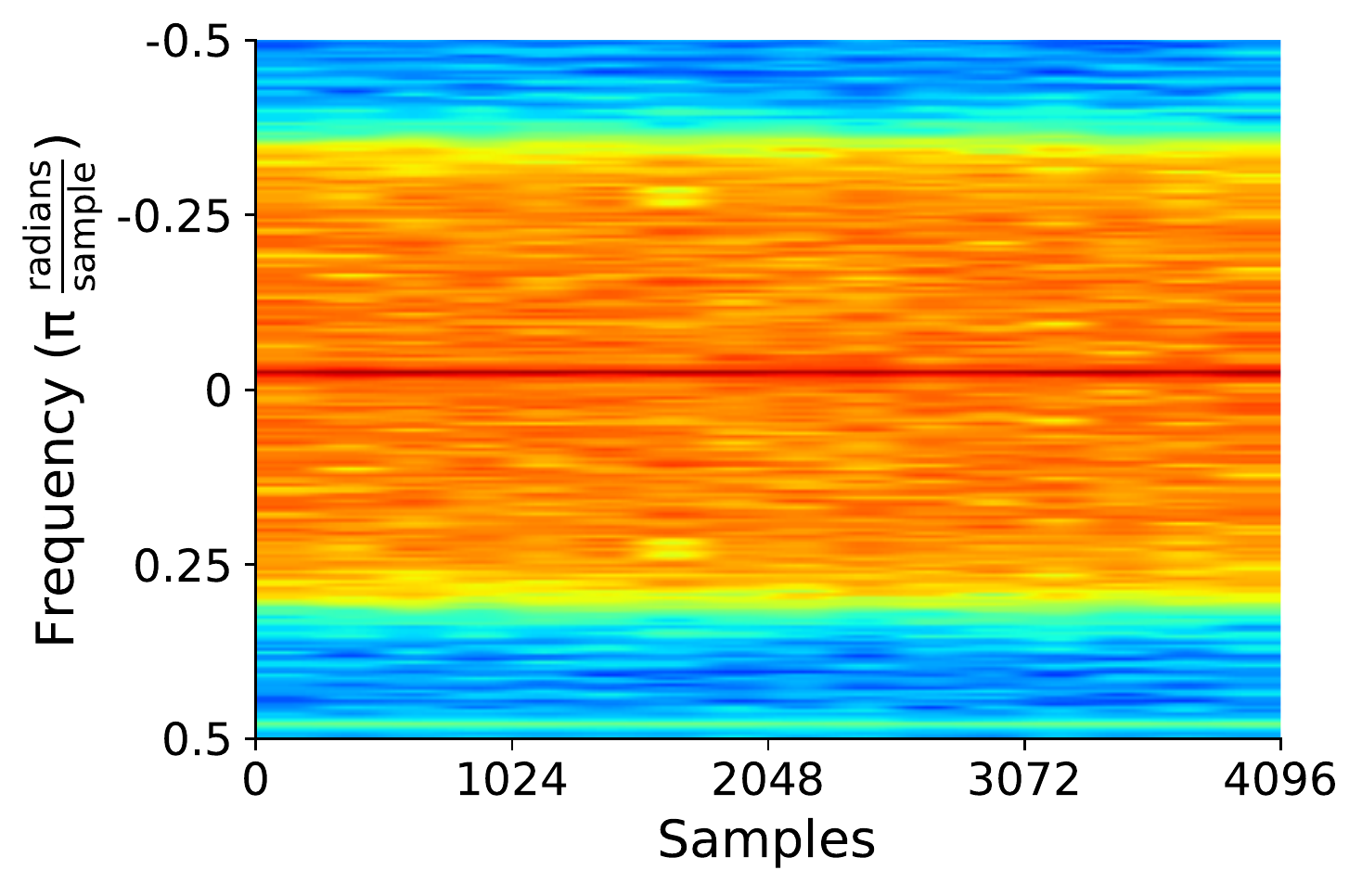}}
  \subfloat[Augmented Data]{\includegraphics[width=0.35\textwidth]{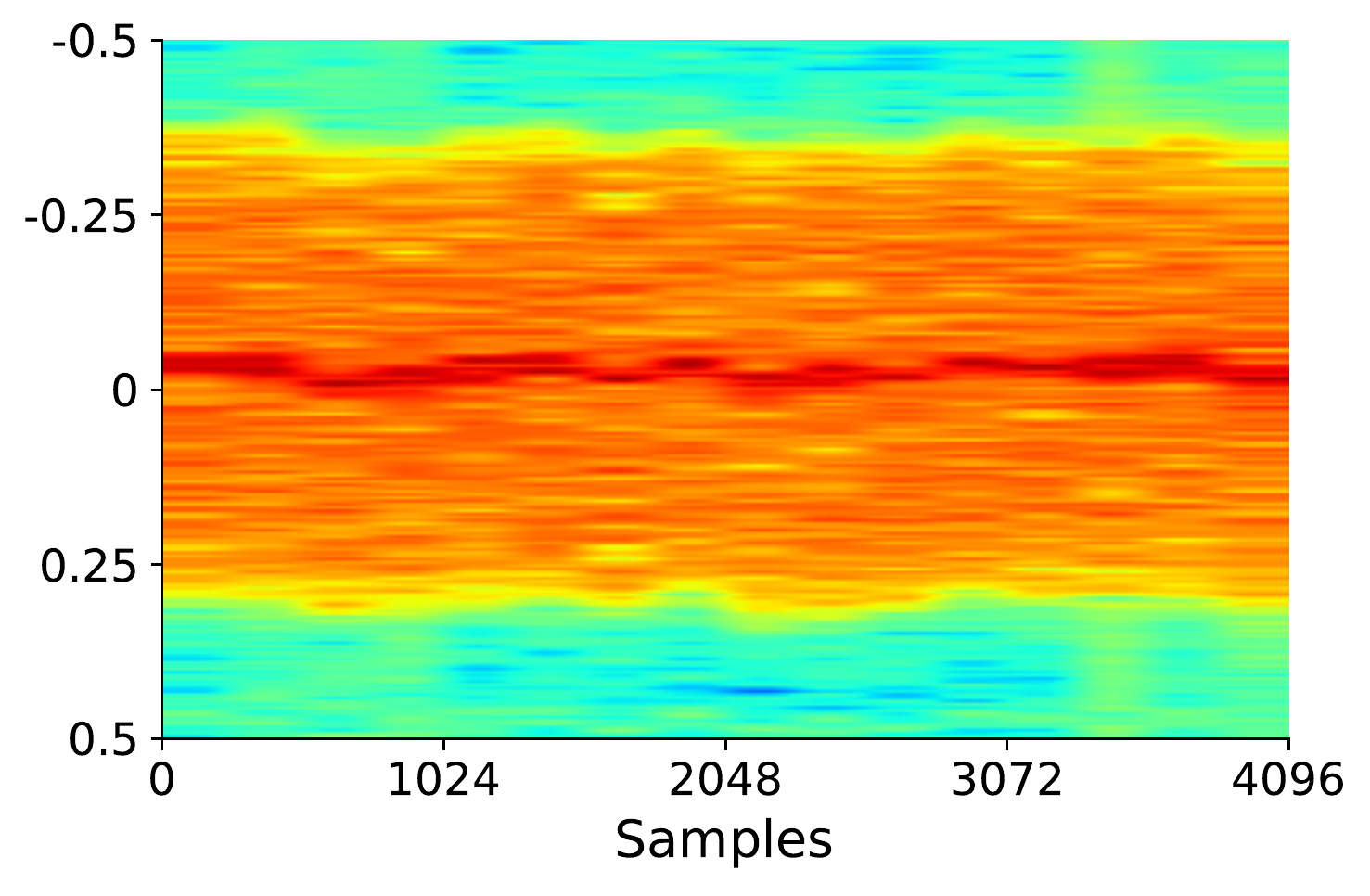}}
  \caption{Local Oscillator Drift Augmentation}
  \label{fig:lo_drift}
\end{figure*}

\textbf{Local Oscillator Drift.} The local oscillator (LO) drift transform emulates the imperfections of a receiver's LO. 
This transform models the LO drift by implementing a random walk in frequency with a drift rate and a max drift set as inputs, where when the max drift is reached, the frequency offset is reset to 0 (\cref{fig:lo_drift}).

\begin{figure*}[!h]
  \centering
  \subfloat[Original Data]{\includegraphics[width=0.35\textwidth]{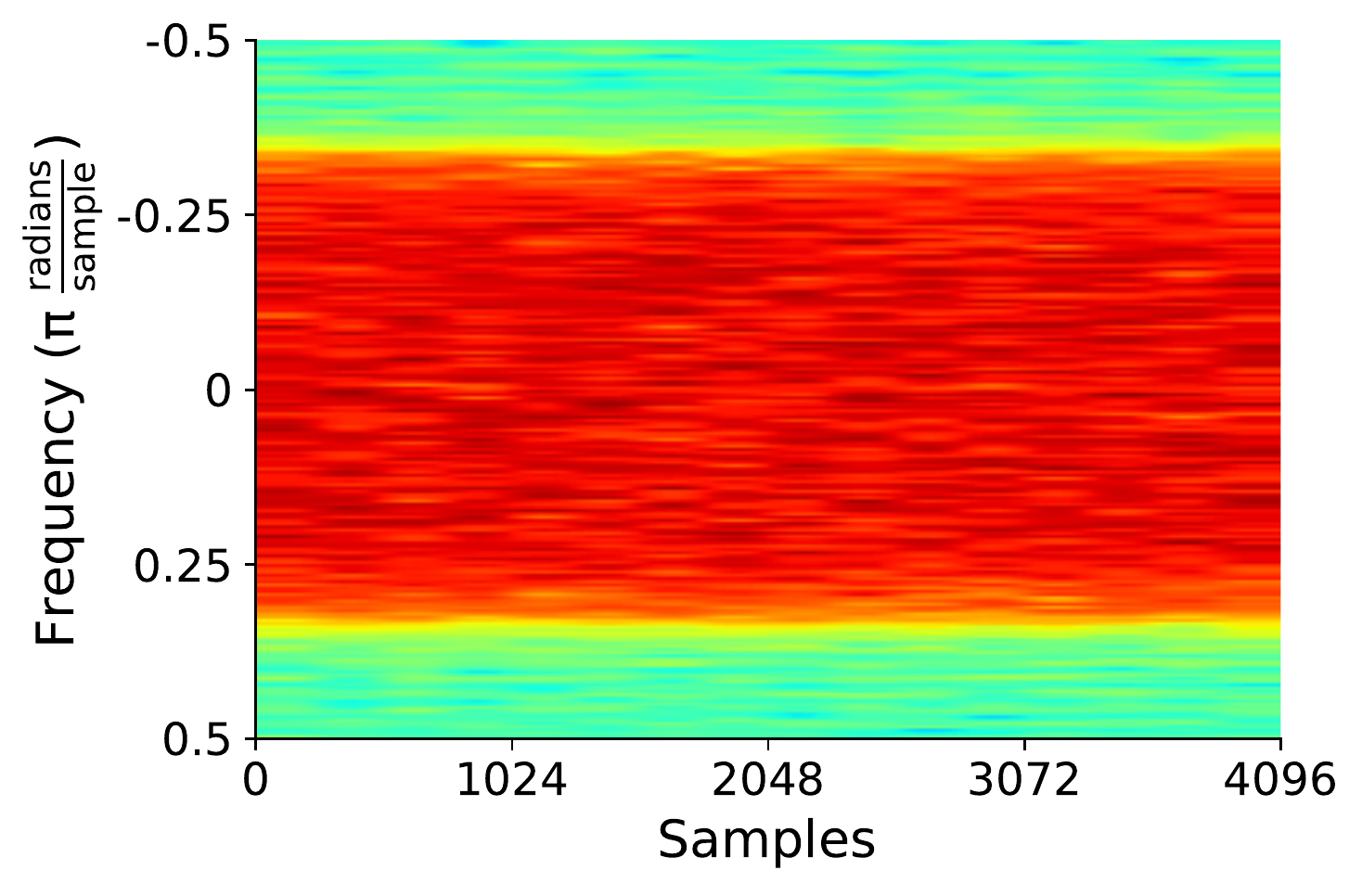}}
  \subfloat[Augmented Data]{\includegraphics[width=0.35\textwidth]{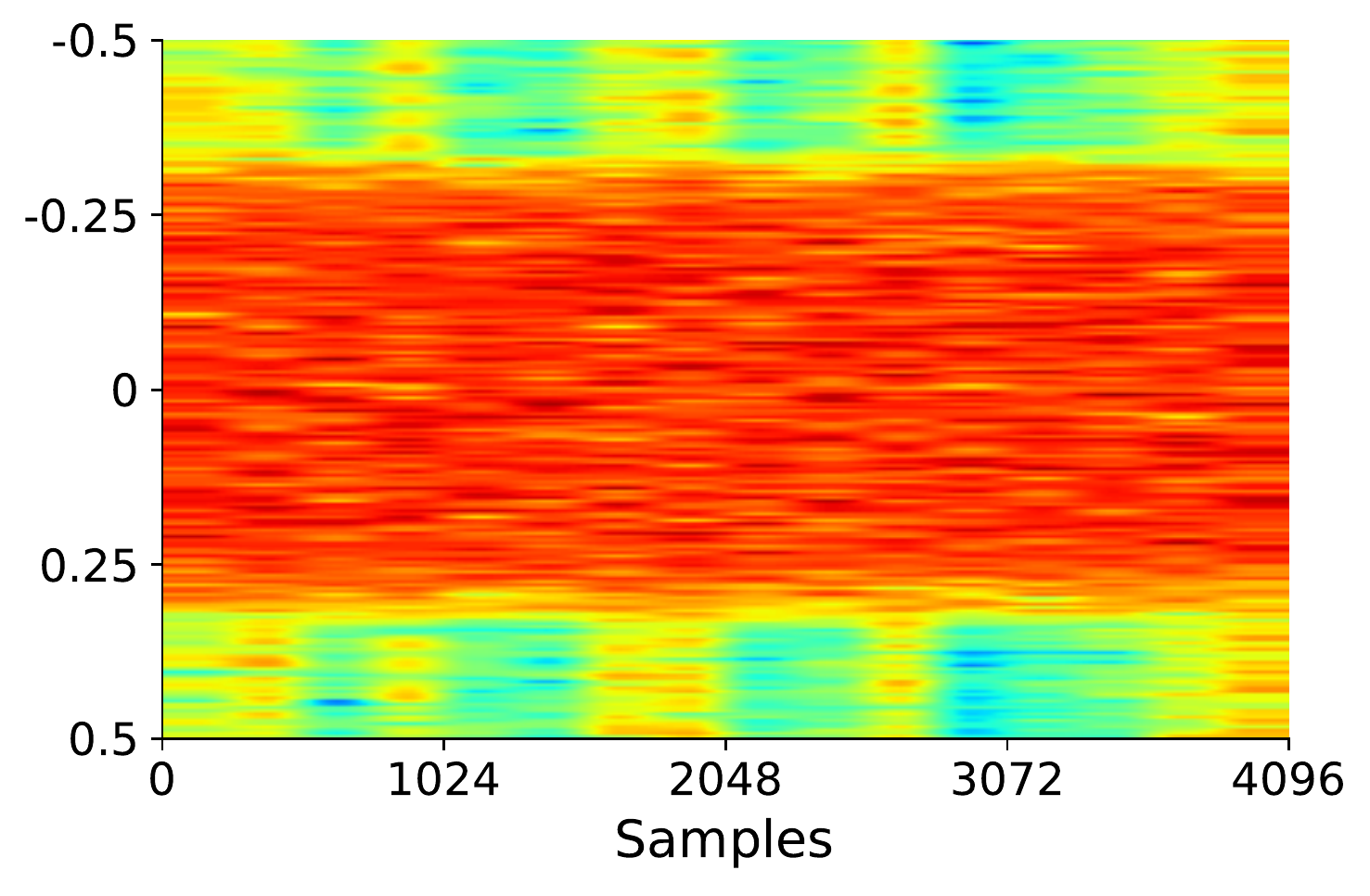}}
  \caption{Time-Varying Noise Augmentation}
  \label{fig:time_varying_noise}
\end{figure*}

\clearpage
\textbf{Time-Varying Noise.} The time-varying transform adds AWGN within a specified low to high SNR range with a specified number of inflection points at which point(s) the slope of the time-varying noise reverses direction (\cref{fig:time_varying_noise}).

\begin{figure*}[!h]
  \centering
  \subfloat[Original Data]{\includegraphics[width=0.35\textwidth]{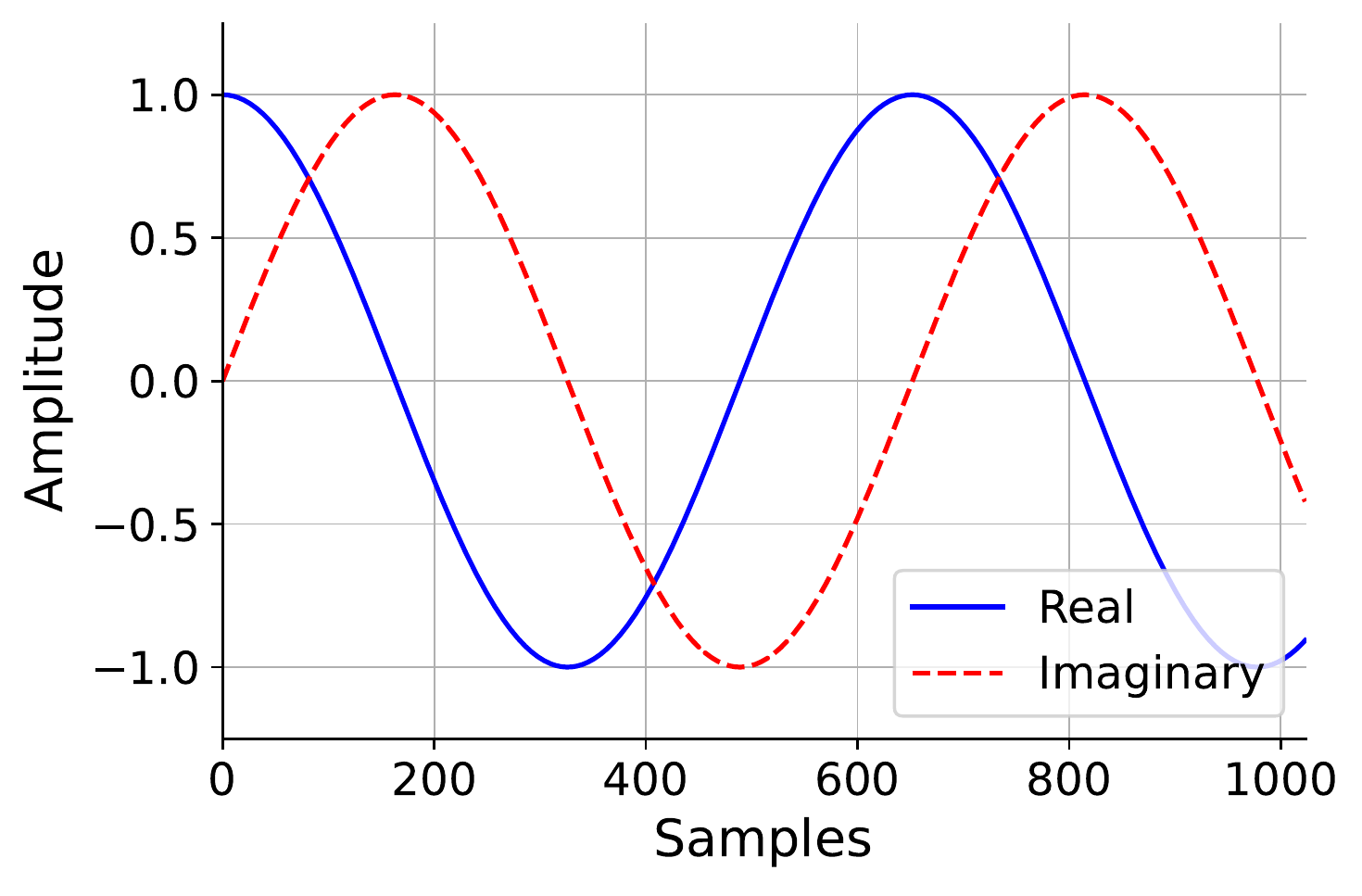}}
  \subfloat[Augmented Data]{\includegraphics[width=0.35\textwidth]{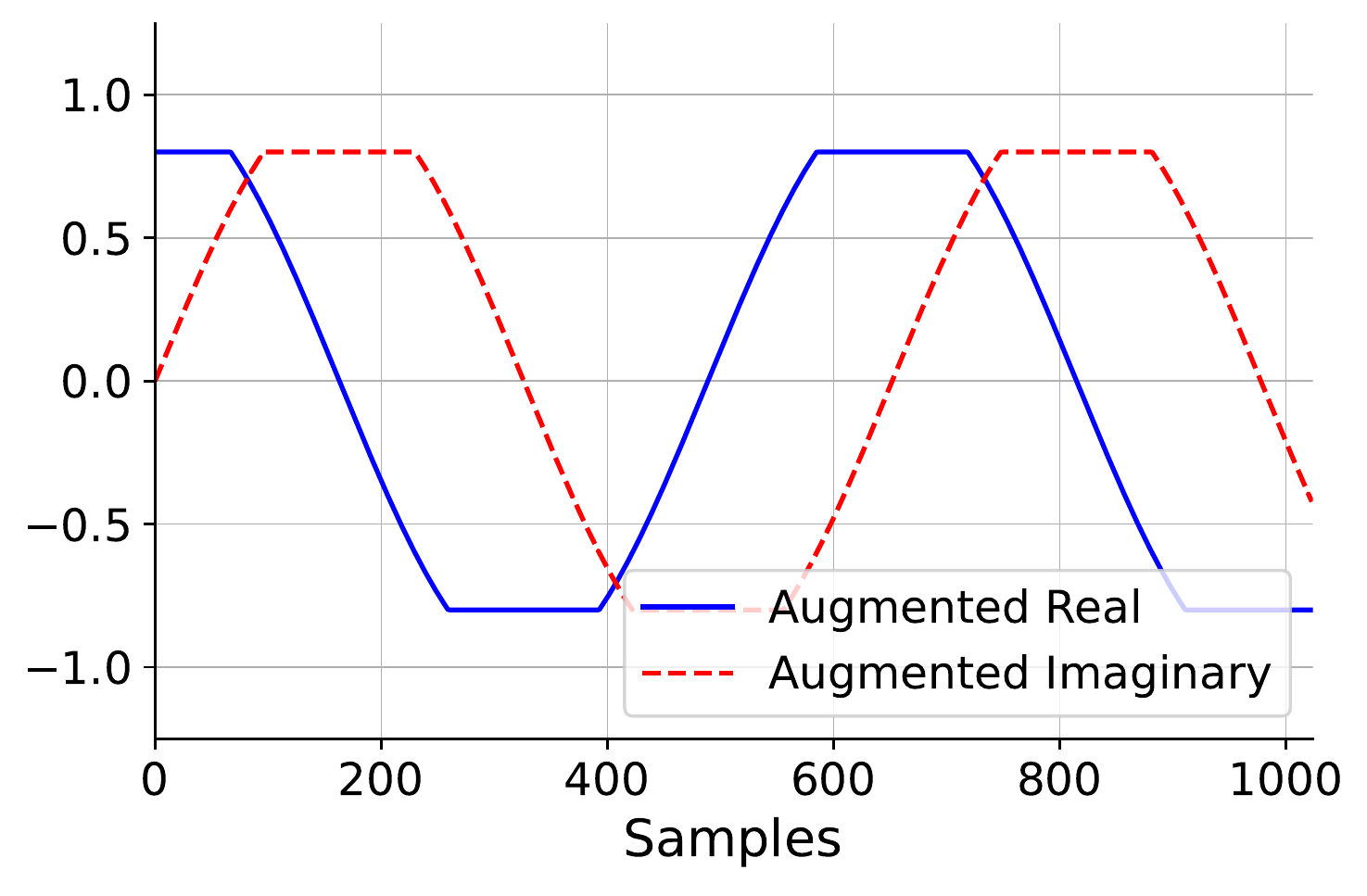}}
  \caption{Clip Augmentation}
  \label{fig:clip}
\end{figure*}

\textbf{Clip.} The clip transform inputs a percentage that it uses to calculate the max and min values allowable through the clipping transform, setting all values above and below these values to the max and min, respectively (\cref{fig:clip}).

\begin{figure*}[!h]
  \centering
  \includegraphics[width=0.45\textwidth]{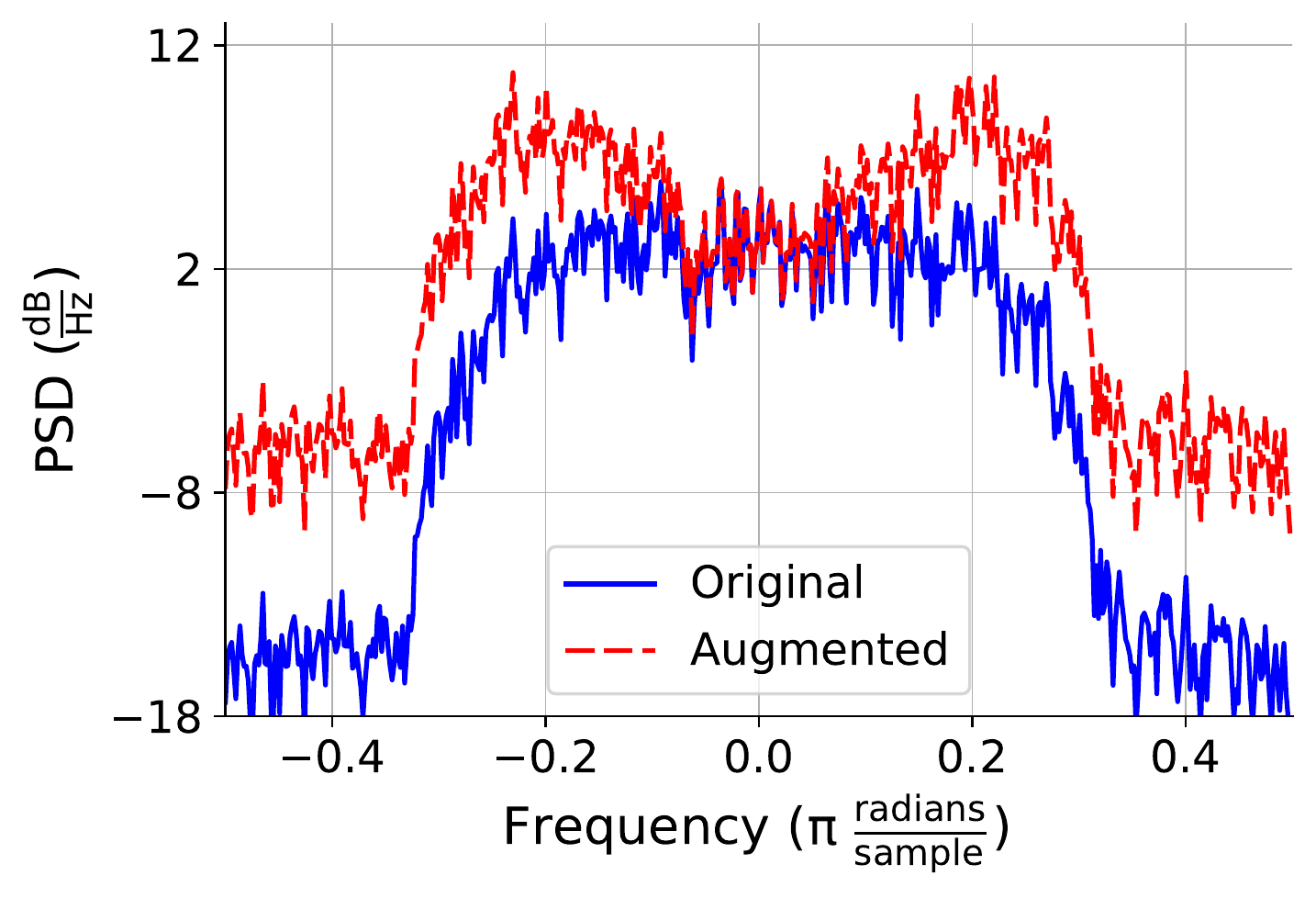}
  \caption{Add Slope Augmentation}
  \label{fig:add_slope}
\end{figure*}

\textbf{Add Slope.} The add slope transform computes the slope between every IQ sample in the input with its preceding sample and adds the slope to its current IQ sample. 
This transform has the effect of amplifying higher frequency components more than the lower frequency components (\cref{fig:add_slope}).

\begin{figure*}[!h]
  \centering
  \includegraphics[width=0.45\textwidth]{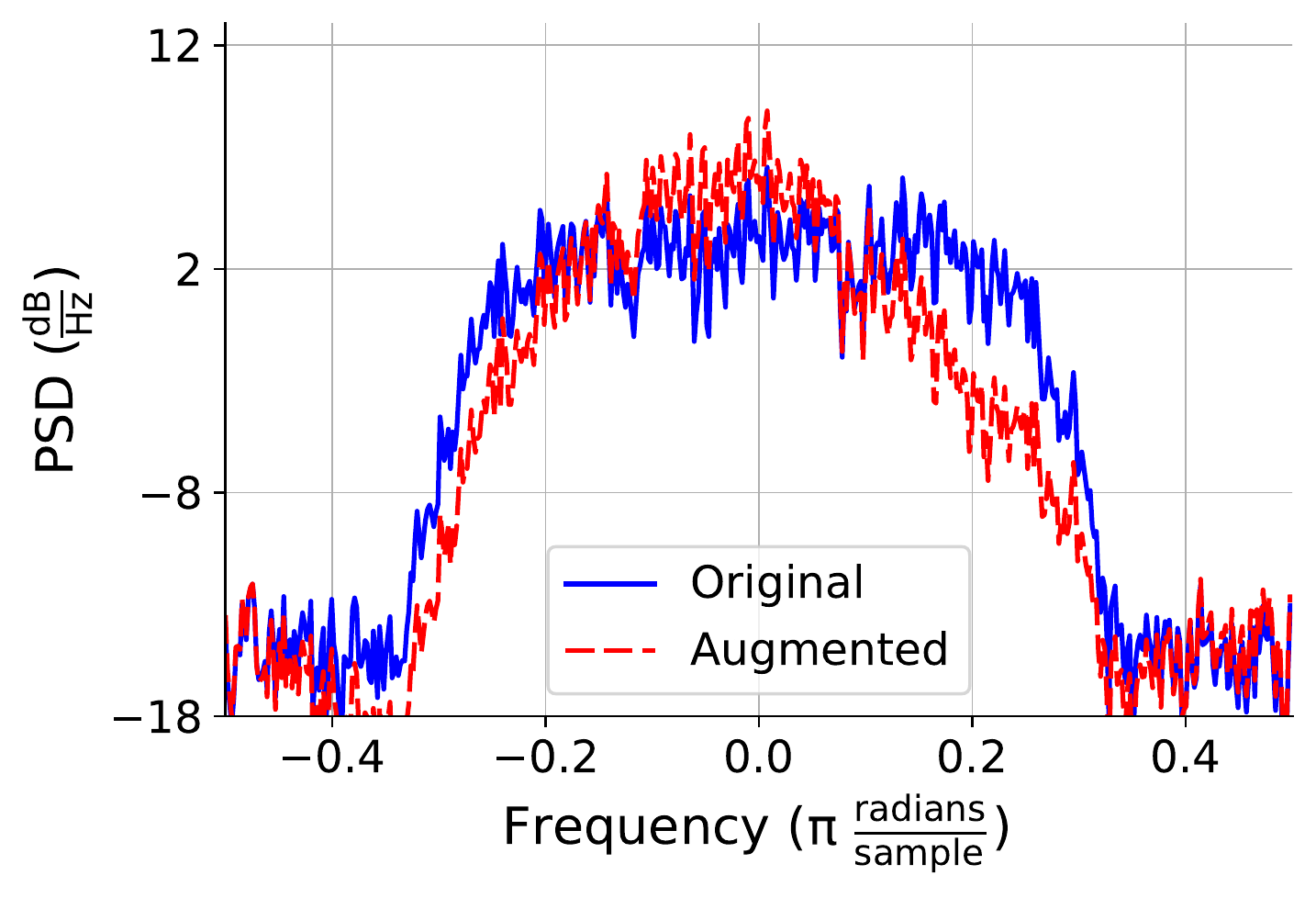}
  \caption{Random Convolve Augmentation}
  \label{fig:random_convolve}
\end{figure*}

\clearpage
\textbf{Random Convolve.} The random convolve transform inputs a random distribution of the number of taps in a filter, where each tap is assigned random values in range 0 to 1.
The randomly generated filter is then convolved with the input IQ data. 
An alpha value is also input to the transform and it is used to dampen the effect of the randomly filtered data by weighting the newly filtered data and inversely weighting the original data and then summing the results. 
This random convolution is a relatively cheap form of applying a frequency-selective fading model (\cref{fig:random_convolve}).

\begin{figure*}[!h]
  \centering
  \subfloat[Original Data]{\includegraphics[width=0.35\textwidth]{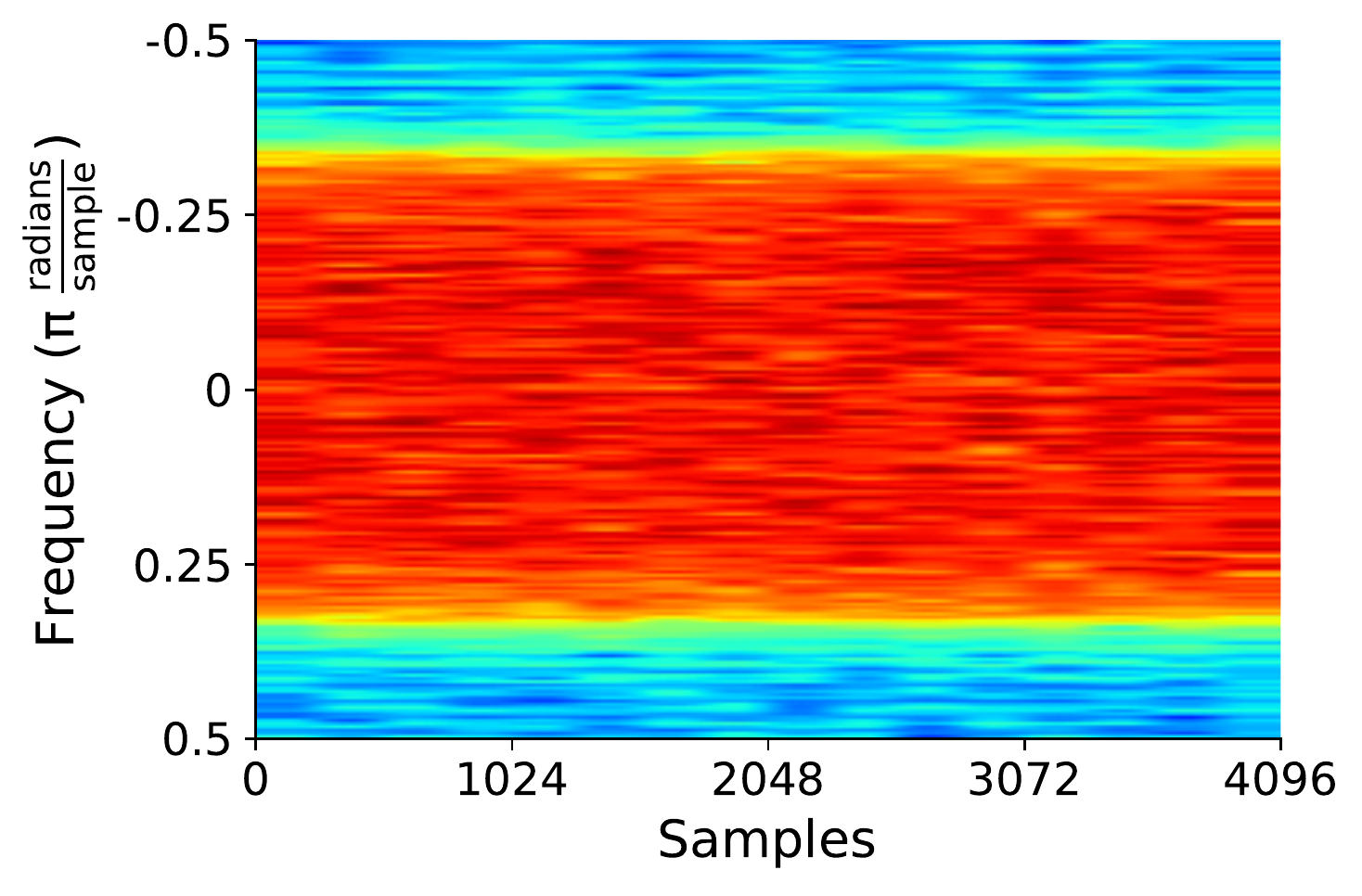}}
  \subfloat[Augmented Data]{\includegraphics[width=0.35\textwidth]{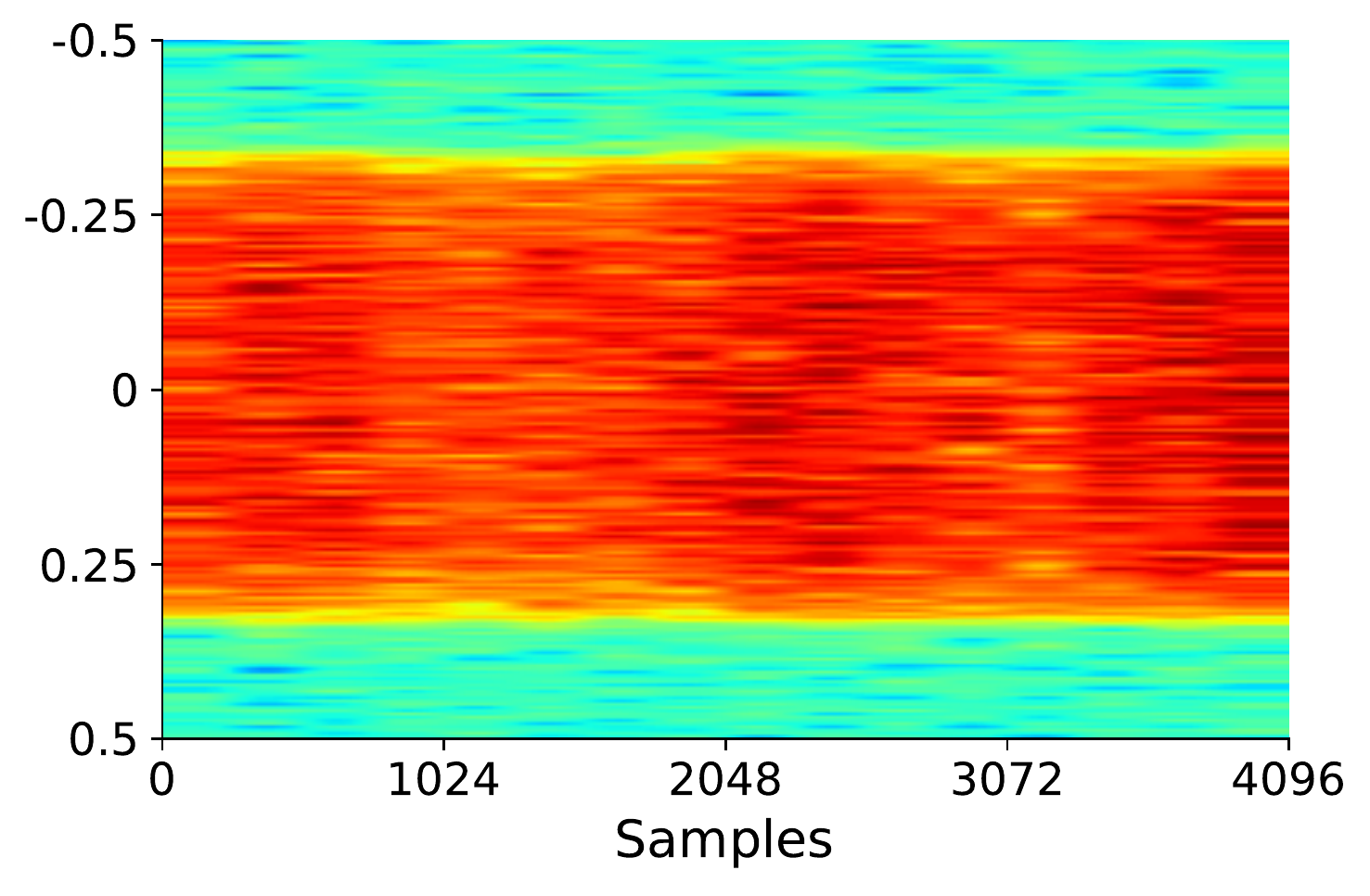}}
  \caption{Gain Drift Augmentation}
  \label{fig:gain_drift}
\end{figure*}

\textbf{Gain Drift.} A gain drift transform is also implemented to complement the LO drift transform's frequency effects with magnitude effects. 
The gain drift transform inputs similar max/min drift values and a drift rate, which are used in a random walk of adjusting the magnitudes of the input data IQ samples over time (\cref{fig:gain_drift}).

\begin{figure*}[!h]
  \centering
  \subfloat[Original Data]{\includegraphics[width=0.35\textwidth]{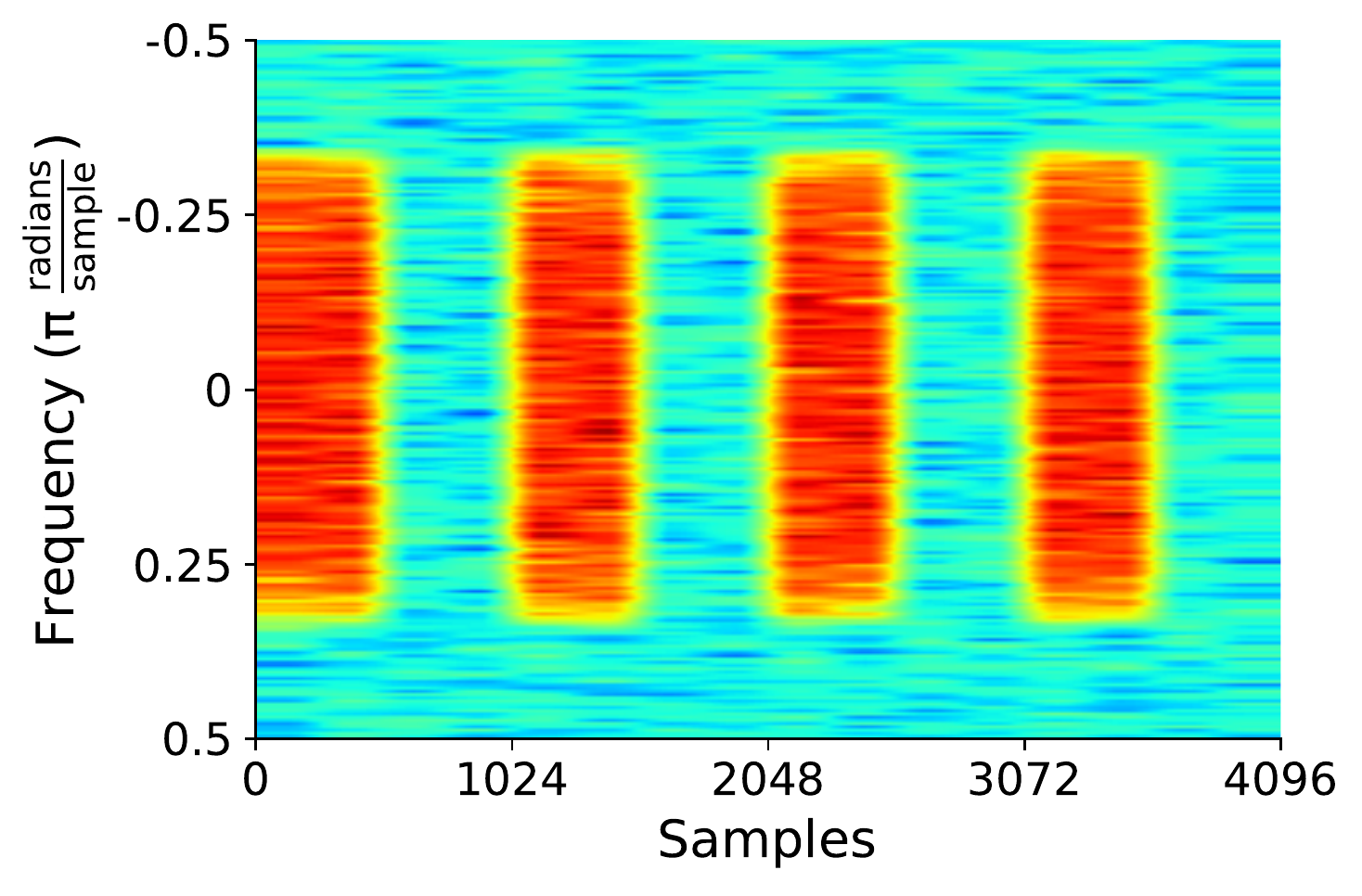}}
  \subfloat[Augmented Data]{\includegraphics[width=0.35\textwidth]{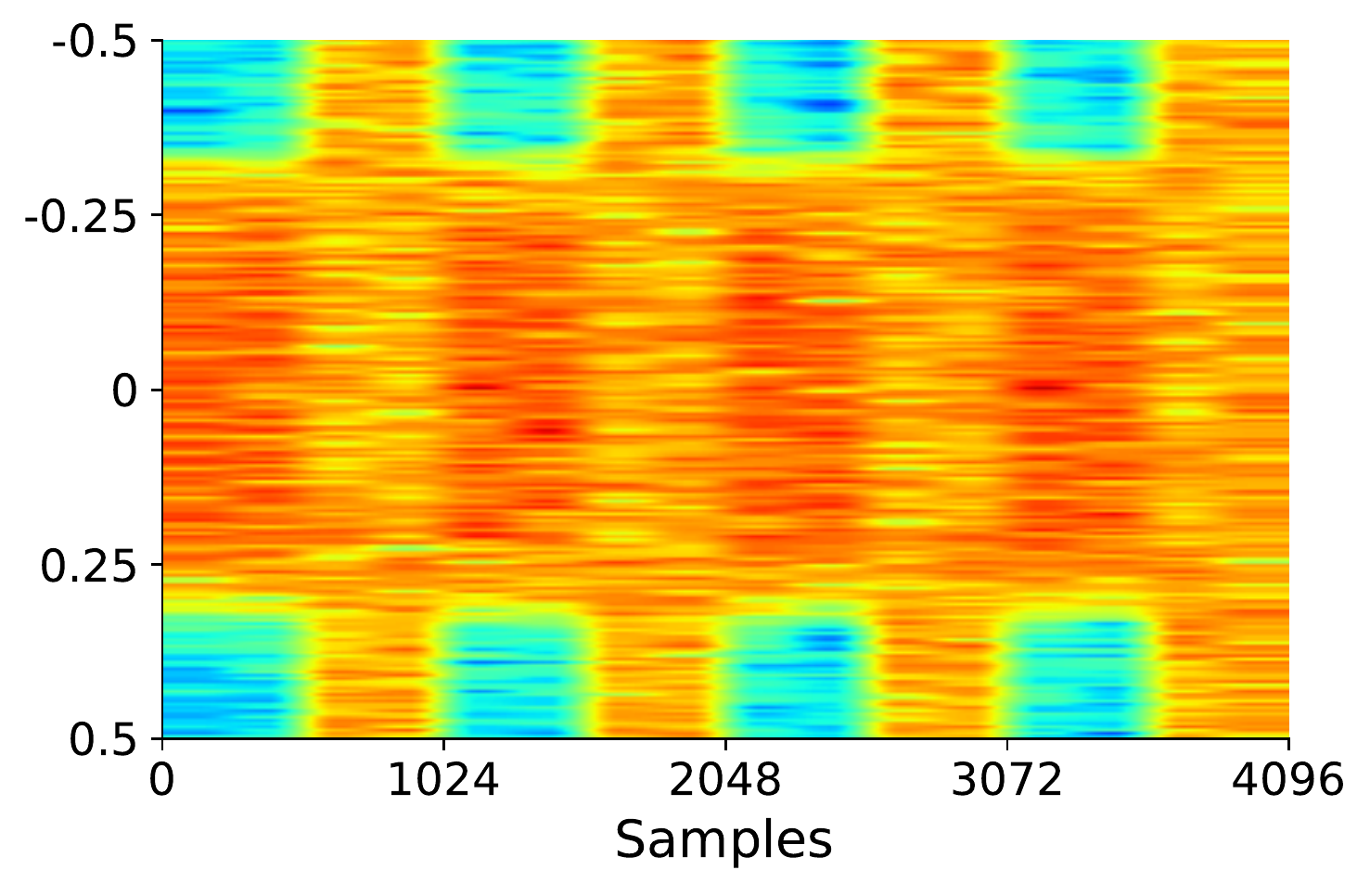}}
  \caption{Automatic Gain Control Augmentation}
  \label{fig:agc}
\end{figure*}

\clearpage
\textbf{Automatic Gain Control.} An automatic gain control (AGC) implementation is included in TorchSig as a data transform with an input scaling of default values for randomization across augmentation calls. 
The default values under random scaling can also be updated with arguments including: 
an initial gain value, 
an alpha for averaging the measure signal level, 
an alpha amount by which to adjust gain when in the tracking state, 
an alpha value by which to adjust gain when in the overflow state, 
an alpha value by which to adjust gain when in the acquire state, 
a reference level specifying the level of intended gain adjustment, 
a tracking range of allowable deviation before going into the acquire state, 
a low level which specifies when the AGC is disabled, and 
a high level which specifies when the AGC enters the overflow state (\cref{fig:agc}).

\begin{figure*}[!h]
  \centering
  \subfloat[OFDM Data]{\includegraphics[width=0.30\textwidth]{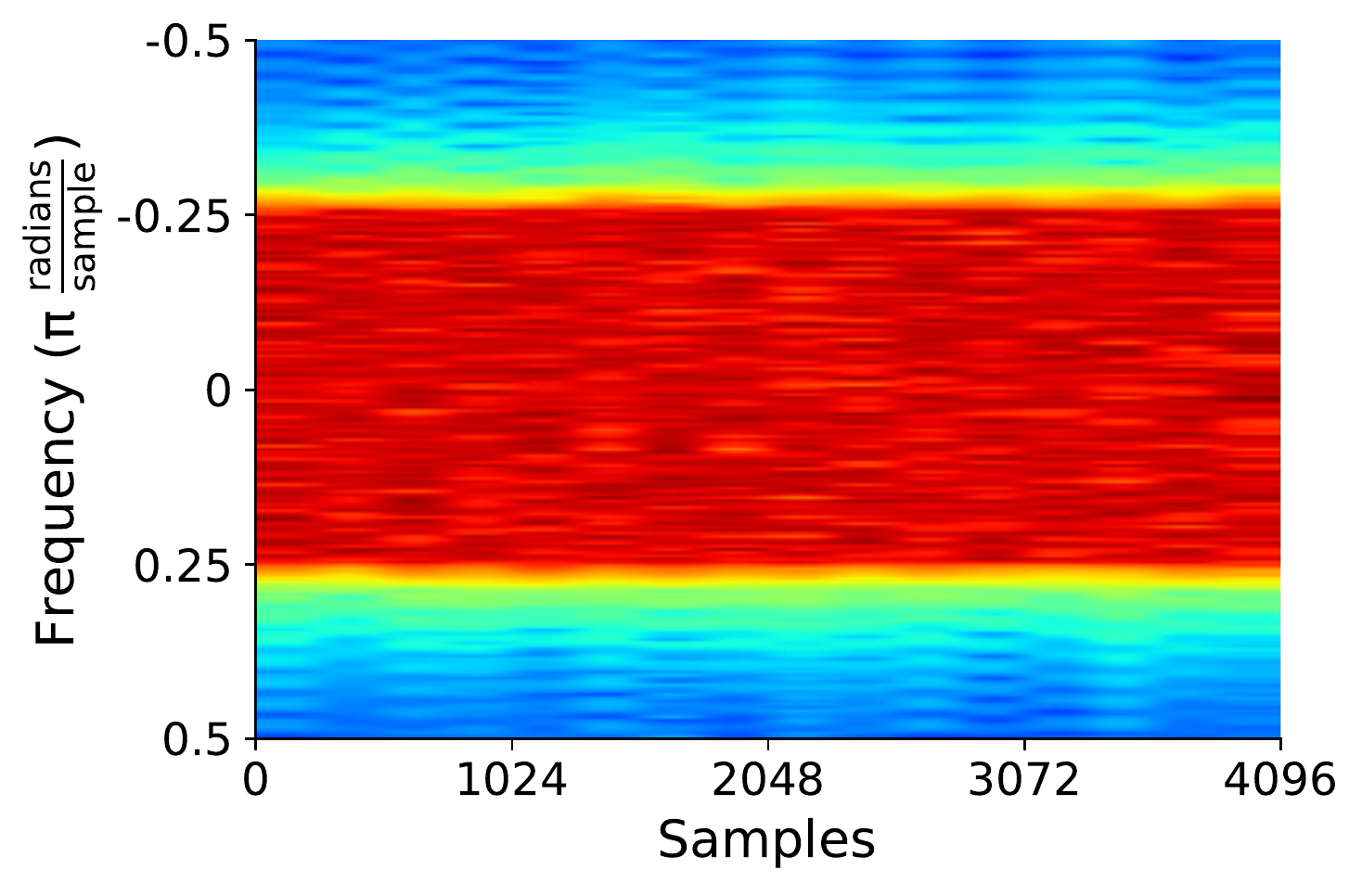}}
  \subfloat[FSK Data]{\includegraphics[width=0.30\textwidth]{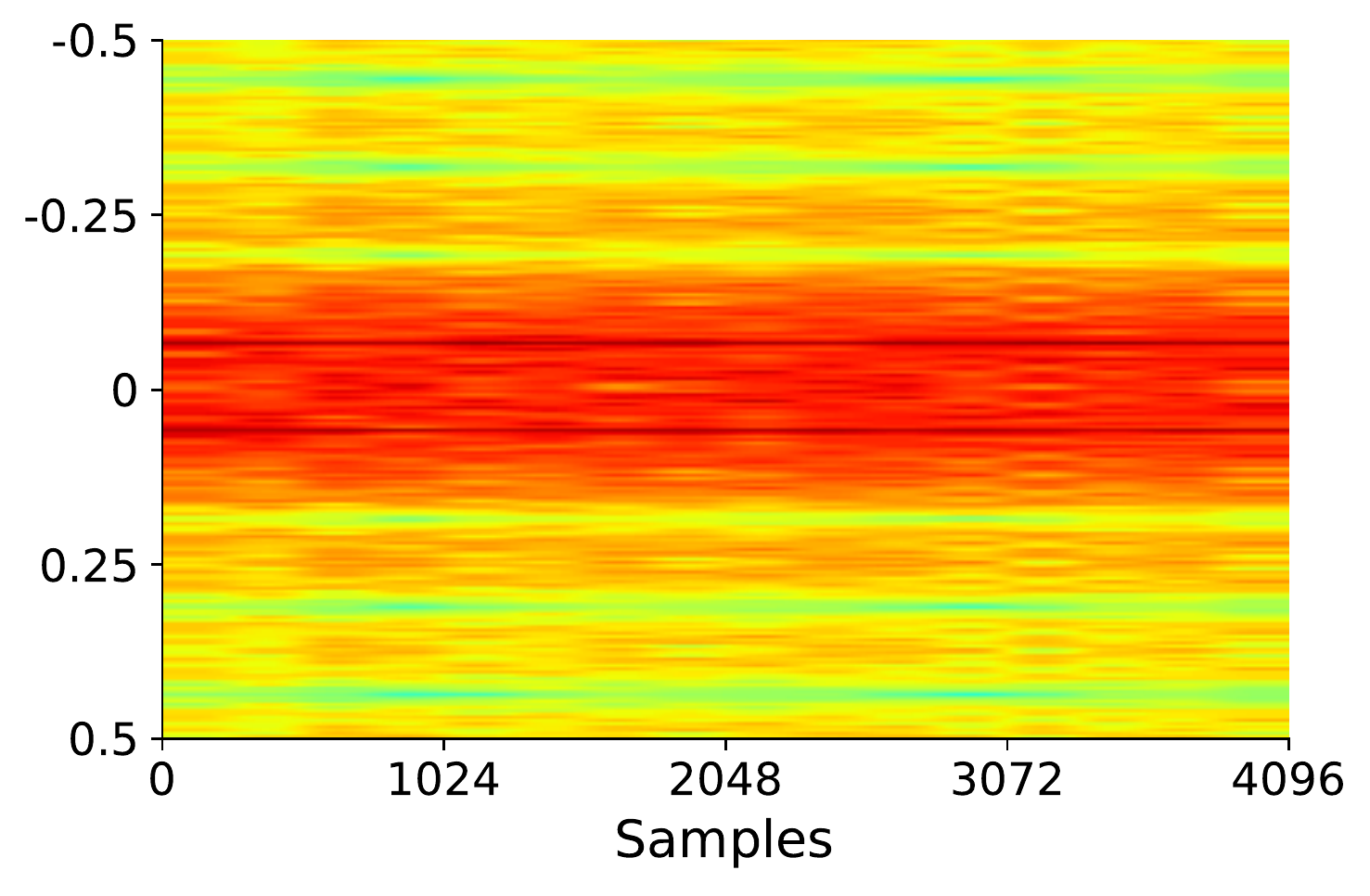}}
  \subfloat[MixUp of OFDM \& FSK Data]{\includegraphics[width=0.30\textwidth]{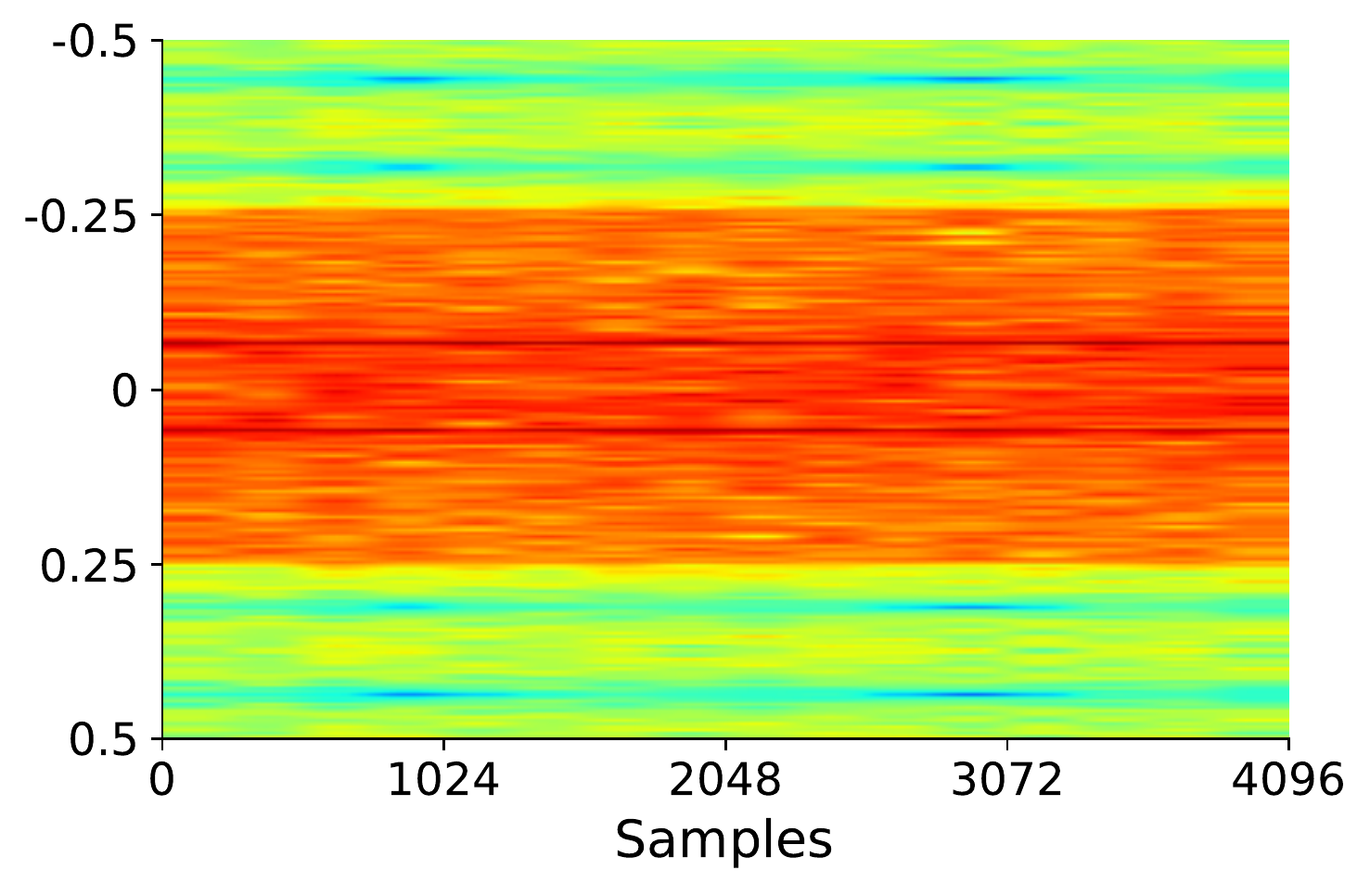}}
  \caption{MixUp Augmentation}
  \label{fig:mixup}
\end{figure*}

\textbf{MixUp.} The MixUp transform implements a modified version of the computer vision domain's MixUp \cite{zhang2018mixup}. 
Our MixUp transform inputs a dataset from which to randomly sample the secondary signal to be mixed with the original signal. 
The transform also inputs an alpha value specifying the logarithmic difference in SNR levels between the two signals. 
Additionally, since the secondary signal may not be of the same class as the original signal, 
the class label information is updated to include the added signal's metadata (\cref{fig:mixup}).

\begin{figure*}[!h]
  \centering
  \subfloat[OFDM Data]{\includegraphics[width=0.30\textwidth]{images/ofdm_data.pdf}}
  \subfloat[FSK Data]{\includegraphics[width=0.30\textwidth]{images/fsk_data.pdf}}
  \subfloat[CutMix of OFDM \& FSK Data]{\includegraphics[width=0.30\textwidth]{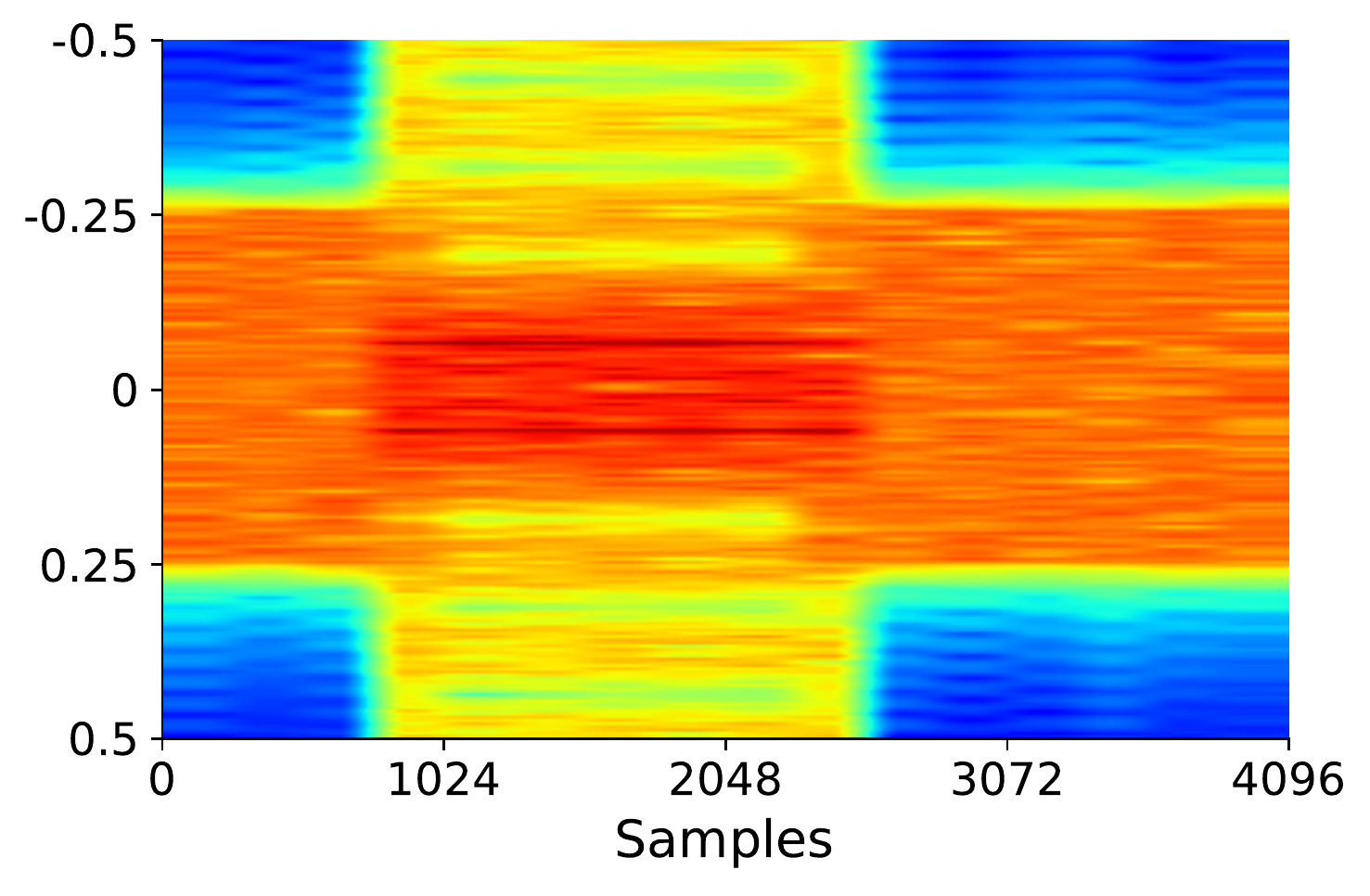}}
  \caption{CutMix Augmentation}
  \label{fig:cutmix}
\end{figure*}

\clearpage
\textbf{CutMix.} The CutMix transform implements a modified version of the computer vision domain's CutMix \cite{yun2019cutmix}. 
Like the MixUp transform, our CutMix transform inputs a dataset to randomly sample the secondary signal from and insert into the original signal, 
replacing a random region of the signal with the new signal. 
An additional alpha value is input specifying the relative durations in time to occupy between the original and the newly inserted signal (\cref{fig:cutmix}).

\textbf{Expert Feature Transforms.} In addition to the data impairment transforms above, TorchSig also includes multiple expert feature transforms for representing the complex IQ signal data in multiple different representations, 
such as through interleaving the real and imaginary floating point values, 
converting the complex IQ samples to two channels of real and imaginary parts, 
computing the complex magnitude, 
computing the wrapped phase, 
transforming to the discrete Fourier representation, 
transforming to variations of a spectrogram, 
and wavelet transforms. 
For the research contained within this paper, the complex IQ samples are converted to 2 channels of real and imaginary parts.
Exploring the other supported transforms' effects while training on the Sig53 dataset remains an open research area.

\clearpage
\subsection{Network Backgrounds}
\label{sec:appendix_model}

\subsubsection{EfficientNet} \label{sec:effnet_background}

\begin{figure*}[!h]
  \centering
  \includegraphics[width=0.60\textwidth]{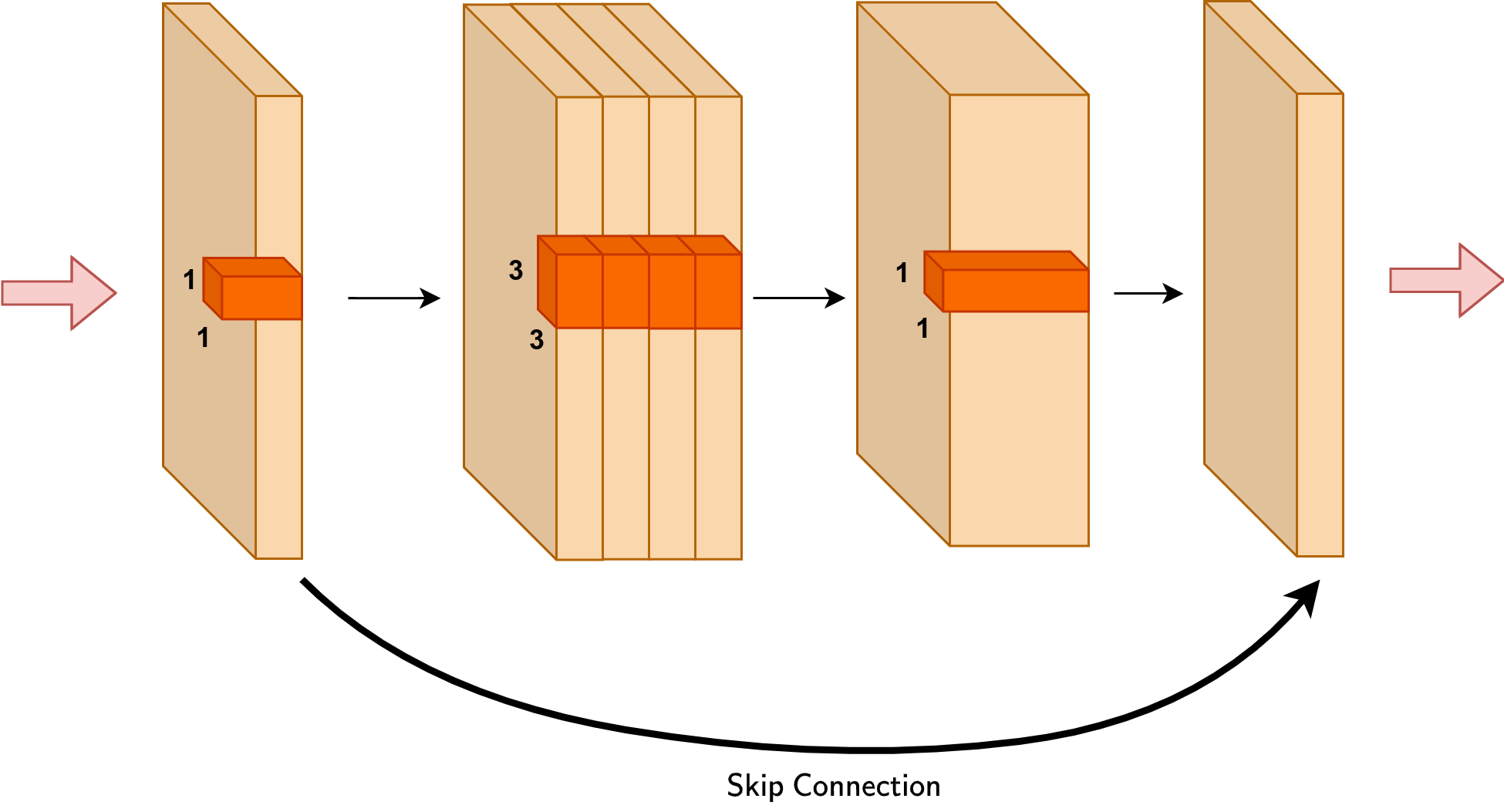}
  \caption{Inverted Residual Block}
  \label{fig:inverted_residual_block}
\end{figure*}

\cite{tan2019efficientnet} provided a family of convolutional Deep Neural Networks (DNN)s, where the model scaling has been optimized. 
EfficientNets were chosen as a good model baseline for the ease of training and configuration. 
This is primarily due to the Squeeze and Excite structures found in the Inverted Residual Blocks used throughout the network. 
This structure is shown in \cref{fig:inverted_residual_block}. 
Note the narrow-wide-narrow characteristic (number of channels are represented as a width in the x-axis) of the network. 
This creates a bottleneck, where the network forces channel outputs from the previous layer to be projected to a lower space. 
EfficientNets are composed of multiple layers of Inverted Residual Blocks.

The power of this structure is that it provides three key advantages:
\begin{itemize}
  \setlength\itemsep{0em}
  \item It imposes a bottleneck in the channel space which helps with generalization.
  \item Depthwise convolutional layers reduce computational load.
  \item Skip connections mitigate vanishing gradients through the network.
\end{itemize}

Note that the sum between the skip connection and the bottleneck layers creates a form of attention. 
The channels are dynamically weighted and added with the original.  
This biases which channels are contributing most the follow-on activation functions.  
However, EfficientNets do not have the same receptive field as the attention mechanism found in transformer networks.

\subsubsection{XCiT} \label{sec:xcit_background}

\begin{figure*}[!h]
  \centering
  \includegraphics[width=0.50\textwidth]{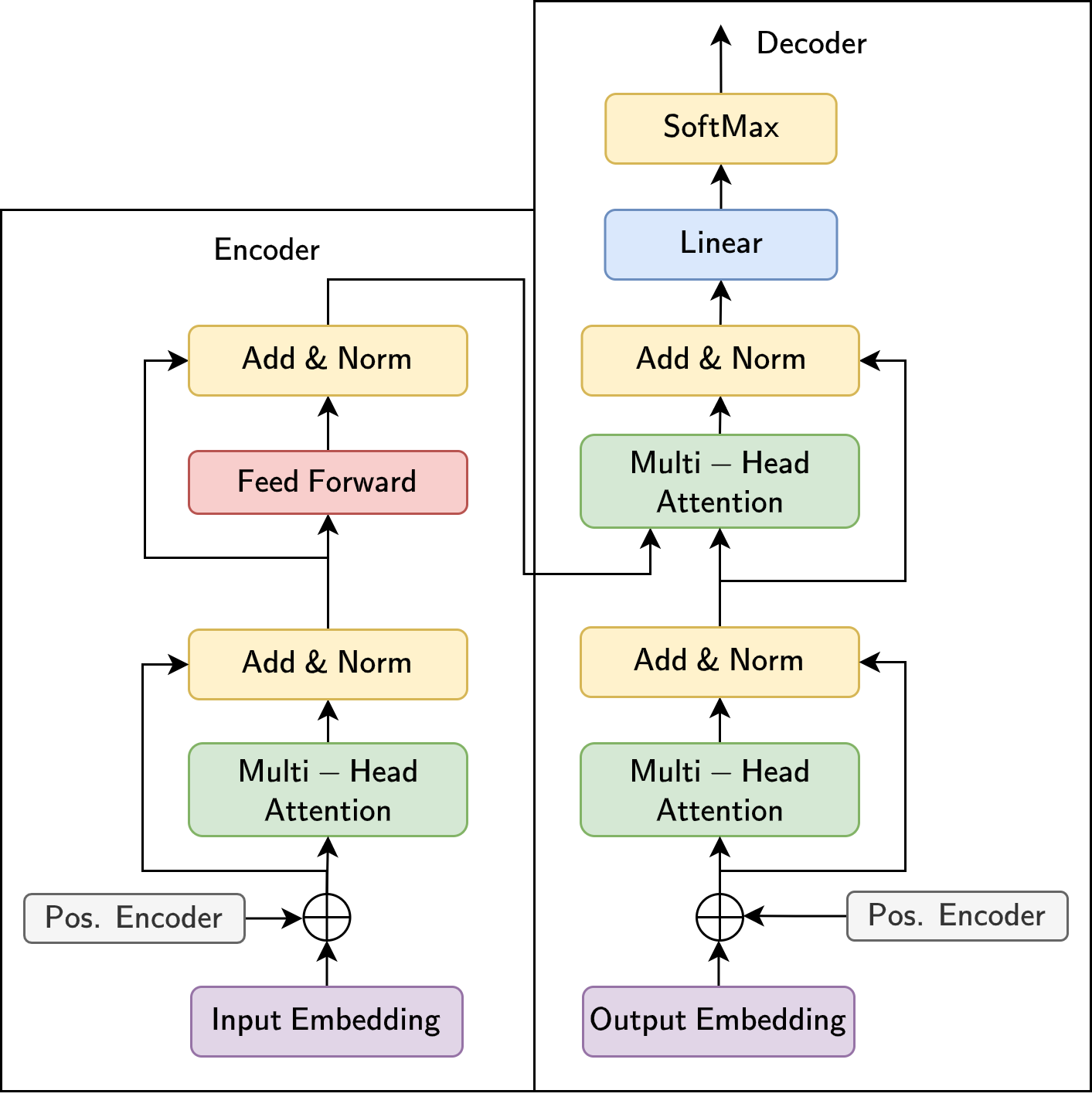}
  \caption{Basic Transformer}
  \label{fig:transformer}
\end{figure*}

The Cross-Covariance Image Transformer (XCiT) is described in \cite{el2021xcit}. 
The base transformer block diagram is shown in \cref{fig:transformer}. 
The tranformer was first used for Natural Language Processing (NLP) tasks and has the common encoder/decoder structure. 
The difference between the input and output embedding is that the output embedding is one time step delayed from the embedding. 
This forces the decoder to predict the next time step based on the learned embedding (Key-value paired description) of the encoder. 
Modern transformer architectures stack the encoder/decoder layers to form a deep network. 

\begin{figure*}[!h]
  \centering
  \includegraphics[width=0.40\textwidth]{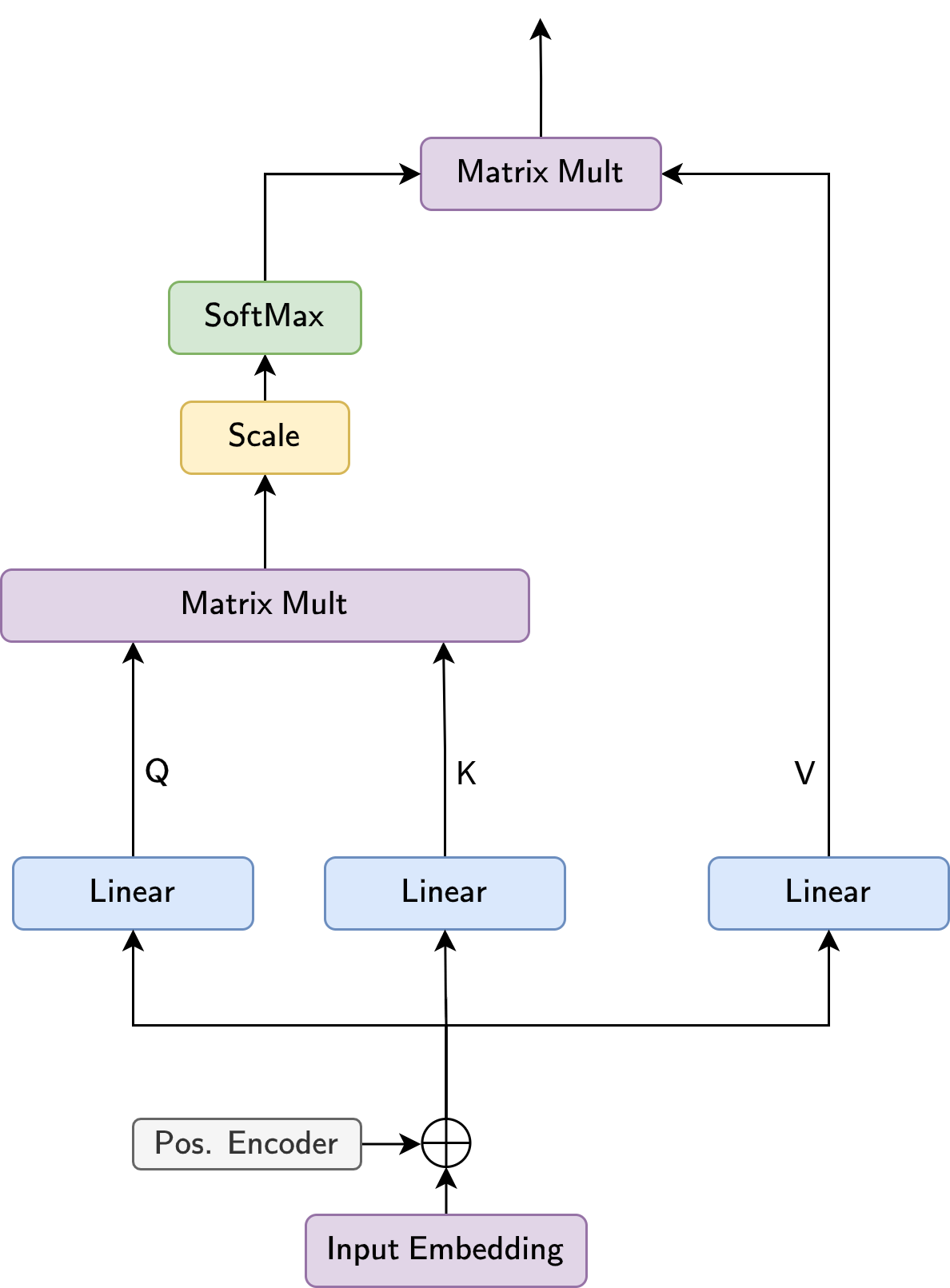}
  \caption{Multi-Head Attention}
  \label{fig:multi_head_attention}
\end{figure*}

The distinguishing feature of transformers are the Multi-Head Attention mechanism shown in \cref{fig:multi_head_attention}. 
The input embedding is projecting to Query ($Q$), Key ($K$), and Value ($V$) pairs through linear layers.  
The dot-product of the query, $Q$, is then computed against all keys, $K$ and the Softmax function is applied to the output as 

\begin{equation} \label{eq:dot_product}
  \mathcal{A}(K,Q) = Softmax\left(\frac{Q K^T}{\sqrt{d_k}} \right)
\end{equation}

Where $d_k$ is the number dimensions in $Q$ and $K$. Generally, this is the token length for NLP applications. Finally, the values $V$ are then weighted by~\eqref{eq:dot_product} resulting in the final attention mechanism given as

\begin{equation} \label{eq:attention_mech}
    Attention(K,Q,V) = Softmax\left(\frac{Q K^T}{\sqrt{d_k}} \right)V
\end{equation}

The main advantage of transformers over convolutional networks is the theoretically infinite receptive field due to the full cross-product of $K$ and $Q$. 
Multi-Head Attention is achieved by applying this architecture per output channel (determined by hyperparameters).
Note that \cref{eq:dot_product} results in a quadratic increase in computational complexity as the sequence length grows linearly. 
The key modification that the XCiT transformer makes is to apply the attention mechanism in channel space as 

\begin{equation} \label{eq:xcit_dprod}
    \mathcal{A}(K,Q) = Softmax\left(\left(\frac{\hat{K}^T}{\tau}\right)\hat{Q}^T \right)
\end{equation}

Where $\hat{K}$ and $\hat{Q}$ are $l_2$ normalized Key, Query pairs in feature space. 
$\tau$ is a temperature parameter used to scale the inner product before the Softmax operation. 
It is a learned parameter. 
XCiT now achieves linear growth in complexity since the channel dimension, $d$, remains fixed as token size increases.  
The linear layers are the only modules that grow in complexity and the input shape is increased.
  
\end{document}